\documentclass{article}

\usepackage{microtype}
\usepackage{graphicx}
\usepackage{subfigure}
\usepackage{booktabs} 

\usepackage{hyperref}

\usepackage[accepted]{icml2024}

\usepackage{amsmath}
\usepackage{amssymb}
\usepackage{mathtools}
\usepackage{amsthm}

\usepackage{natbib}
\usepackage{pifont}
\usepackage{graphicx}
\usepackage{subfigure}
\usepackage{threeparttable}
\usepackage[figuresright]{rotating}
\usepackage{xcolor}
\usepackage{colortbl}
\usepackage{multirow}
\usepackage{wrapfig}
\usepackage{algorithm}
\usepackage{algorithmic}
\usepackage[most]{tcolorbox}
\usepackage{pythonhighlight}

\usepackage[capitalize,noabbrev]{cleveref}

\theoremstyle{plain}
\newtheorem{theorem}{Theorem}[section]

\newtheorem{lemma}[theorem]{Lemma}

\newtheorem{definition}[theorem]{Definition}
\newtheorem{assumption}[theorem]{Assumption}
\newtheorem{remark}[theorem]{Remark}

\usepackage[textsize=tiny]{todonotes}

\icmltitlerunning{MOKD: Cross-domain Finetuning for Few-shot Classification via Maximizing Optimized Kernel Dependence}

\begin{document}

\twocolumn[
\icmltitle{MOKD: Cross-domain Finetuning for Few-shot Classification via\\ Maximizing Optimized Kernel Dependence}




\begin{icmlauthorlist}
\icmlauthor{Hongduan Tian}{hkbu-tmlr,hkbu}
\icmlauthor{Feng Liu}{melb}
\icmlauthor{Tongliang Liu}{sydney}
\icmlauthor{Bo Du}{whu}
\icmlauthor{Yiu-ming Cheung}{hkbu}
\icmlauthor{Bo Han}{hkbu-tmlr,hkbu}
\end{icmlauthorlist}

\icmlaffiliation{hkbu-tmlr}{TMLR Group, Hong Kong Baptist University}
\icmlaffiliation{hkbu}{Department of Computer Science, Hong Kong Baptist University}
\icmlaffiliation{melb}{TMLR Group, University of Melbourne}
\icmlaffiliation{sydney}{Sydney AI Centre, The University of Sydney}
\icmlaffiliation{whu}{National Engineering Research Center for Multimedia Software, Institute of Artificial Intelligence, School of Computer Science, Wuhan Univeristy}

\icmlcorrespondingauthor{Bo Han}{bhanml@comp.hkbu.edu.hk}

\icmlkeywords{Machine Learning, ICML}

\vskip 0.3in
]



\printAffiliationsAndNotice{} 

\begin{abstract}
 In cross-domain few-shot classification, \emph{nearest centroid classifier} (NCC) aims to learn representations to construct a metric space where few-shot classification can be performed by measuring the similarities between samples and the prototype of each class.
 An intuition behind NCC is that each sample is pulled closer to the class centroid it belongs to while pushed away from those of other classes. 
 However, in this paper, we find that there exist high similarities between NCC-learned representations of two samples from different classes. 
 In order to address this problem, we propose a bi-level optimization framework, \emph{maximizing optimized kernel dependence} (MOKD) to learn a set of class-specific representations that match the cluster structures indicated by labeled data of the given task. 
 Specifically, MOKD first optimizes the kernel adopted in \emph{Hilbert-Schmidt independence criterion} (HSIC) to obtain the optimized kernel HSIC (opt-HSIC) that can capture the dependence more precisely. Then, an optimization problem regarding the opt-HSIC is addressed to simultaneously maximize the dependence between representations and labels and minimize the dependence among all samples. 
 Extensive experiments on Meta-Dataset demonstrate that MOKD can not only achieve better generalization performance on unseen domains in most cases but also learn better data representation clusters. The project repository of MOKD is available at: \href{https://github.com/tmlr-group/MOKD}{https://github.com/tmlr-group/MOKD}.
\end{abstract}

\addtocontents{toc}{\protect\setcounter{tocdepth}{-1}}
\section{Introduction}
Cross-domain few-shot classification~\citep{SUR,url,urt,metadatasets}, also known as CFC, is a learning paradigm which aims at learning to perform classification on tasks sampled from previously unseen data or domains with only a few labeled data available. Compared with conventional few-shot classification~\citep{maml,ravi2016optimization,prototypical,MatchingNet} which learns to adapt to new tasks sampled from unseen data with the same distribution as seen data, cross-domain few-shot classification is a much more challenging learning task since there exist discrepancies between the distributions of source and target domains~\citep{chi2021tohan,kuzborskij2013stability}.

Due to its simplicity and scalability, \emph{nearest centroid classifier} (NCC)~\citep{prototypical} has been widely applied in recent works~\citep{crosstransformer,url,urt,metadatasets} regarding cross-domain few-shot classification. The goal of NCC is to learn representations to construct a metric space where few-shot classification can be performed by measuring the similarities between samples and the prototype of each class. Intuitively, the learning process via NCC is pulling each sample closer to the class centroid it belongs to while pushing it away from other class centroids. Thus, the learned representations are expected to be specific enough to be distinguished from other classes while identified by the class they belong to.

\begin{figure*}[t]
    \vspace{-0.5em}
	\vskip 0.0in
	\begin{center}
	\centering
    \subfigure[NCC-based Loss (ImageNet with 13 classes)\label{Fig:intro_imagenet_pa}]{
			\begin{minipage}[t]{0.48\linewidth}
				\centering
				\includegraphics[width=1.0\linewidth]{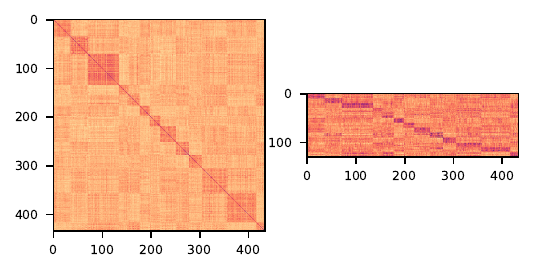}
		\end{minipage}}\vspace{-0.1cm}
    \subfigure[MOKD (ImageNet with 13 classes)\label{Fig:intro_imagenet_mokd}]{
			\begin{minipage}[t]{0.48\linewidth}
				\centering
				\includegraphics[width=1.0\linewidth]{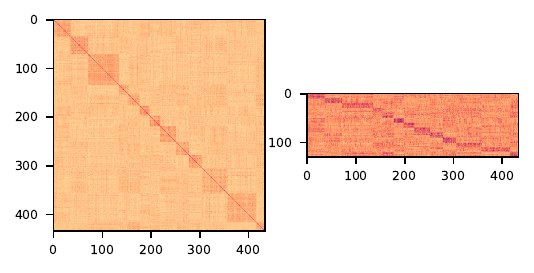}
		\end{minipage}}\vspace{-0.1cm}	
    \caption{\textbf{Heatmaps of similarity matrices of representations respectively learned with NCC-based loss and MOKD.} The left of each figure describes the similarities among all support data representations and the right side describes the similarities between query data and support data representations. As shown in (a), NCC-learned representations of samples are not only similar to samples belonging to their own class but also similar to samples from other classes. (b) shows that the undesirable high similarities existing between samples from different classes are significantly alleviated and the cluster structures of the given set of data are well explored by applying MOKD.}
    \label{Fig:intro_fig}
    \end{center}
    \vskip -0.2in
    \vspace{-0.2em}
\end{figure*}

However, in this paper, we find that there exist high similarities between NCC-learned representations of two samples coming from different classes. For example, as shown in Fig.~\ref{Fig:intro_imagenet_pa}, the heatmap of the similarity matrix, which depicts the similarities among support data representations, reveals that the NCC-learned representations of each sample not only resemble the samples that belong to the same class but also have high similarities with samples from other classes. 
Such undesirable high similarities among samples may induce uncertainty and further result in misclassification of samples. Thus, learning a set of \emph{class-specific} representations, where similarities among samples within the same class are maximized while similarities between samples from different classes are minimized, is crucial for CFC. 


To this end, we propose an efficient and effective approach, \emph{\underline{m}aximizing \underline{o}ptimized \underline{k}ernel \underline{d}ependence} (MOKD), to learn a set of class-specific representations, where the similarities among samples belonging to the same class are maximized while the similarities between samples from different classes are minimized, with optimized kernel HSIC measures, where test power is maximized. In general, MOKD is formulated as a bi-level optimization problem for dependence optimization based on optimized kernel measures. Specifically, MOKD first optimizes the kernel used in Hilbert-Schmidt Independence Criterion (HSIC) by maximizing its test power to obtain a powerful kernel dependence measure, the optimized kernel HSIC (opt-HSIC). The goal of this step is to increase the sensitivity of the kernel HSIC to dependence. 
Then, an optimization objective with respect to the opt-HSIC is optimized to simultaneously maximize the dependence between the kernelized representations and labels and minimize the dependence among all sample representations. In this way, a set of class-specific representations is learned. As shown in Fig ~\ref{Fig:intro_imagenet_mokd}, MOKD does help alleviate the undesirable high similarities between samples from different classes and learn better data clusters.

Extensive experiments on Meta-Dataset~\cite{metadatasets} benchmark under several task settings and further analysis results demonstrate that MOKD is an efficient and effective algorithm. On the one side, numerical results indicate that MOKD can achieve better generalization performance than previous baselines on unseen domains. On the other side, analysis results further reveal that test power maximization is essential for achieving good performance and our proposed MOKD method can mitigate the undesirable high similarities between data representations from different classes and in turn learn better sample clusters.

\textbf{Our Contribution.} In this paper, we find that there exist high similarities between NCC-learned representations of data from different classes, which may further induce uncertainty and in turn result in misclassification. To address this problem, we first provide a new perspective of the NCC-based loss from dependence measure and reveal that NCC-based loss can be expressed in the format of Hilbert-Schmidt Independence Measure (HSIC). Based on this, we further propose a new method, \emph{maximizing optimized kernel dependence} (MOKD), to learn class-specific representations, where the similarities among samples within the same class are maximized while the similarities between samples from different classes are minimized. Extensive experimental results demonstrate that our proposed MOKD not only achieves better generalization performance without increasing running time but also learns better data clusters.

\section{Preliminary}

\textbf{Dataset Structure.} Let $\mathcal{S}$ denote a meta dataset that contains $n$ sub-datasets with different distributions, i.e. $\mathcal{S}=\{\mathcal{S}_1, \mathcal{S}_2, ..., \mathcal{S}_{|\mathcal{S}|}\}$. For each sub-dataset $\mathcal{S}_i$, three \emph{disjoint} subsets, which are training set $\mathcal{D}^{\rm tr}_{\mathcal{S}_i}$, validation set $\mathcal{D}^{\rm val}_{\mathcal{S}_i}$ and test set $\mathcal{D}^{\rm test}_{\mathcal{S}_i}$, are included, i.e. $\mathcal{S}_i=\{\mathcal{D}^{\rm tr}_{\mathcal{S}_i}, \mathcal{D}^{\rm val}_{\mathcal{S}_i}, \mathcal{D}^{\rm test}_{\mathcal{S}_i}\}$. 
In the context of cross-domain few-shot classification, a feature encoder is trained on the training sets of a portion of sub-datasets of $\mathcal{S}$. Thus, the sub-datasets whose training sets are observed by the feature encoder during the pre-training phase are called \emph{seen domains} and denoted as $\mathcal{S}^{\rm seen}$ while the remaining datasets are called \emph{unseen domains} and denoted as $\mathcal{S}^{\rm unseen}$. Note that $\mathcal{S}^{\rm seen}$ and $\mathcal{S}^{\rm unseen}$ are disjoint.

\textbf{Task Generation.} A task $\mathcal{T}=\{\mathcal{D}_{\mathcal{T}}, \mathcal{Q}_{\mathcal{T}}\}$, where $\mathcal{D}_{\mathcal{T}}=\{\boldsymbol{X}^{\rm s}, Y^{\rm s}\}=\{(\boldsymbol{x}^{\rm s}_i, y^{\rm s}_i)\}_{i=1}^{|\mathcal{D}_{\mathcal{T}}|}$ denotes the support data pairs and $\mathcal{Q}_{\mathcal{T}}=\{\boldsymbol{X}^{\rm q}, Y^{\rm q}\}=\{(\boldsymbol{x}^{\rm q}_j, y^{\rm q}_j)\}_{j=1}^{|\mathcal{Q}_{\mathcal{T}}|}$ denotes query data pairs, is randomly sampled from the specific dataset at the beginning of each episode. 
The sampling of \emph{vary-way vary-shot} cross-domain few-shot classification tasks follows the rules proposed in \citet{metadatasets}. The sampling process can be roughly divided into two steps. Firstly, the number of classes $N$ is randomly sampled from the interval $[5, N_{\max}]$, where $N_{\max}$ denotes the maximum of the number of classes and is either 50 or as many classes as available. Then, the number of shots of each class, $K_n$, for $n=\{1, 2, ..., N\}$, in support set is determined with specific rules~(refer to~\citet{metadatasets} or Appendix~\ref{Appendix:vary_settings} for details). Thus, in the support set of a task, each datapoint belonging to class $n$ can be treated as being independently sampled from the given dataset with the probability $\frac{1}{NK_n}$. 

\textbf{Pre-training Few-shot Classification Pipeline.} Consider a pre-trained backbone $f_{\phi^*}$ parameterized with the optimal pre-trained parameters $\phi^*$ and a linear transformation head $h_{\theta}$ parameterized with $\theta$. Given a support set $\mathcal{D}_{\mathcal{T}}$, the corresponding representations can be obtained by applying the pre-trained backbone $f_{\phi^*}$ and the linear transformation head $h_{\theta}$: $\mathcal{Z}=\{\boldsymbol{z}^{\rm s}_i\}_{i=1}^{|\mathcal{D}_{\mathcal{T}}|} = \{h_{\theta}\circ f_{\phi^*}(\boldsymbol{x}_i^{\rm s})\}_{i=1}^{|\mathcal{D}_{\mathcal{T}}|}$.
Then, according to URL~\citep{url}, the learning problem can be solved by minimizing the following NCC-based loss~\citep{prototypical} on the given support set $\mathcal{D}_{\mathcal{T}}$:
\begin{equation}\label{Eq:prototypical_loss}
    \mathcal{L}_{\rm NCC}(\theta, \mathcal{D}_{\mathcal{T}}) = -\frac{1}{|\mathcal{D}_{\mathcal{T}}|}\sum_{i=1}^{|\mathcal{D}_{\mathcal{T}}|}\log\left(p(y=y^{\rm s}_i|\boldsymbol{z}^{\rm s}_i, \theta)\right),
\end{equation}
where $p(y=c|\boldsymbol{z}, \theta) = \frac{e^{k(\boldsymbol{z}, \boldsymbol{c}_c)}}{\sum_{i=1}^{N_C}e^{k(\boldsymbol{z}, \boldsymbol{c}_i)}}$ denotes the likelihood of a given sample $\boldsymbol{z}$ belonging to class $c$, $k(\cdot, \cdot)$ denotes a kernel function which is formulated as a cosine similarity function in URL, $\boldsymbol{c}_c$ denotes the prototype of class $c$ and is calculated as $\boldsymbol{c}_c = \frac{1}{|\mathcal{C}_c|}\sum_{\boldsymbol{z}\in \mathcal{C}_c}\boldsymbol{z}, \mathcal{C}_c=\{\boldsymbol{z}_j|y_j=c\}$.

\textbf{Hilbert-Schmidt Independence Criterion.} Given separable \emph{reproducing kernel Hilbert spaces}~(RKHSs) $\mathcal{F}$, $\mathcal{G}$ and two feature spaces $\mathcal{X}$, $\mathcal{Y}$, HSIC~\citep{hsic} measures the dependence between two random variables $X\in\mathcal{X}$ and $Y\in\mathcal{Y}$ by evaluating the norm of the cross-variance between the features that are respectively transformed by non-linear kernels: $\varphi: \mathcal{X}\rightarrow \mathcal{F}$ and $\psi: \mathcal{Y}\rightarrow \mathcal{G}$:
\begin{equation}\label{Eq:population_hsic}
\begin{small}
    {\rm HSIC}(X, Y)=||\mathbb{E}[\varphi(X)\psi(Y)^{\top}]-\mathbb{E}[\varphi(X)]\mathbb{E}[\psi(Y)]^{\top}||_{HS}^2,
\end{small}
\end{equation}
where $||\cdot||_{HS}$ is the Hilbert-Schmidt norm which is the Frobenius norm in finite dimensions.
Further, let $(X^{'}, Y^{'})$ and $(X^{''}, Y^{''})$ be the independent copies of $(X, Y)$, the HSIC can be formulated as:
\begin{equation}\label{Eq:empirical_hsic}
\begin{small}
\begin{aligned}
    {\rm HSIC}(X, Y)=&\mathbb{E}_{XX^{'}YY^{'}}[k(X, X^{'})l(Y, Y^{'})] \\ &+\mathbb{E}_{XX^{'}}[k(X, X^{'})]\mathbb{E}_{YY^{'}}[l(Y, Y^{'})] \\ &-2\mathbb{E}_{XY}[\mathbb{E}_{X^{'}}[k(X, X^{'})]\mathbb{E}_{Y^{'}}[l(Y, Y^{''})]],
\end{aligned}
\end{small}
\end{equation}
where $k(x, x^{'})=\langle\varphi(x), \varphi(x^{'})\rangle_{\mathcal{F}}$ and $l(y, y^{'})=\langle\psi(y), \psi(y^{'})\rangle_{\mathcal{G}}$ are kernel functions which are defined as inner product operations in reproducing kernel Hilbert space. Note that HSIC takes zero if and only if $X$ and $Y$ are mutually independent~\citep{hsic,hsic_estimation}.

\textbf{Test power of HSIC.} In this paper, test power is used to measure the probability that, for particular two dependent distributions and the number of samples $m$, the null hypothesis that the two distributions are independent is correctly rejected. Consider a $\widehat{\rm HSIC}_{\rm u}$ as an unbiased HSIC estimator (e.g., U-statistic estimator), under the hypothesis that the two distributions are dependent, the central limit theorem~\citep{serfling2009approximation} holds:
\[
    \sqrt{m}(\widehat{{\rm HSIC}}_{\rm u}-{\rm HSIC})\stackrel{d}{\longrightarrow}\mathcal{N}(0, v^2),
\]
where $v^2$ denotes the variance, $\stackrel{d}{\longrightarrow}$ denotes convergence in distribution. The CLT implies that test power can be formulated as:
\[
    {\rm Pr}\left(m\widehat{{\rm HSIC}}_{\rm u} >  r\right)\rightarrow\Phi\left(\frac{\sqrt{m}{\rm HSIC}}{v}-\frac{r}{\sqrt{m}v}\right),
\]
where $r$ denotes a rejection threshold and $\Phi$ denotes the standard normal CDF. Since the rejection threshold $r$ will converge to a constant, and ${\rm HSIC}$, $v$ are constants, for reasonably large $m$, the test power is dominated by the first term. Thus, a feasible way to maximize the test power is to find a kernel function to maximize ${{\rm HSIC}}/{v}$. The intuition of test power maximization is increasing the sensitivity of the estimated kernel to the dependence among data samples. 

\section{Motivation: Theoretically Understand NCC via the Kernel HSIC Measure}
In this section, we provide an understanding of NCC-based loss from the perspective of HSIC. Specifically, we first reveal two insights behind NCC-based loss. Then, inspired by \citet{sslhsic}, we bridge a connection between NCC-based loss and kernel HSIC and find that the upper bound of NCC-based loss can be treated as the surrogate loss under mild assumptions. All proofs are available in Appendix~\ref{Appendix:Proof_results}. 

\subsection{A Lower Bound of NCC-based Loss}
\begin{assumption}[\citet{sslhsic}]\label{Assumption:proximity}
    Given a kernel function $k(\cdot, \cdot)$, for arbitrary two data representations $\boldsymbol{z}$ and $\boldsymbol{z}^{'}$ from support data set $\mathcal{Z}=\{\boldsymbol{z}_i\}_{i=1}^{|\mathcal{D}_{\mathcal{T}}|}$, we assume that $k(\boldsymbol{z}, \boldsymbol{z}^{'})$ does not deviate much from $\sum_{\boldsymbol{z}^{'}\in\mathcal{Z}}\frac{1}{|\mathcal{D}_{\mathcal{T}}|}e^{k(\boldsymbol{z}, \boldsymbol{z}^{'})}$.
\end{assumption}
Such an assumption is adopted by \citet{sslhsic} to ensure $k(\boldsymbol{z}, \boldsymbol{z}^{'})\approx \sum_{\boldsymbol{z}^{'}\in\mathcal{Z}}\frac{1}{|\mathcal{D}_{\mathcal{T}}|}e^{k(\boldsymbol{z}, \boldsymbol{z}^{'})}$ so that Taylor expansion can be conducted on exponent in the InfoNCE loss~\cite{infonce}. It can also be adopted in our following analysis results without having to add any other assumption.
\begin{theorem}[\textbf{Lower bound of NCC-based loss}]\label{Theorem:lower_bound_protoloss}
Given a set of normalized support representations $\mathcal{Z}=\{\boldsymbol{z}_i\}_{i=1}^{|\mathcal{D}_{\mathcal{T}}|}=\{h_{\theta}\circ f_{\phi^*}(\boldsymbol{x}_i)\}_{i=1}^{|\mathcal{D}_{\mathcal{T}}|}$ and the corresponding labels $\{y_i\}_{i=1}^{|\mathcal{D}_{\mathcal{T}}|}$ that includes $N_C$ classes from a support set $\mathcal{D}_{\mathcal{T}}$. Let $k(\cdot, \cdot)$ be the cosine similarity function. Then, with Assumption~\ref{Assumption:proximity}, the NCC-based loss (Eq.~\eqref{Eq:prototypical_loss}) owns a lower bound:
\[
\begin{small}
\begin{aligned}
    \mathcal{L}(\theta)\ge &-\frac{1}{|\mathcal{D}_{\mathcal{T}}|}\sum_{i=1}^{|\mathcal{D}_{\mathcal{T}}|}\frac{1}{|\mathcal{C}|}\sum_{\boldsymbol{z}^{+}\in\mathcal{C}}k(\boldsymbol{z}_i, \boldsymbol{z}^{+})\\
    &+\frac{1}{|\mathcal{D}_{\mathcal{T}}|}\sum_{i=1}^{|\mathcal{D}_{\mathcal{T}}|}\sum_{\boldsymbol{z}^{'}\in\mathcal{Z}}\frac{k(\boldsymbol{z}_i, \boldsymbol{z}^{'})}{|\mathcal{D}_{\mathcal{T}}|} + \mathcal{O}\left(k(\boldsymbol{z}, \boldsymbol{z}^{'})\right)+const,
\end{aligned}
\end{small}
\]
where $\boldsymbol{z}^{+}$ denotes the data samples belonging to the same class as $\boldsymbol{z}_i$, $\mathcal{C}$ denotes the class that $\boldsymbol{z}_i$ belongs to, $\mathcal{O}\left(k(\boldsymbol{z}, \boldsymbol{z}^{'})\right)$ denotes a high-order moment term. In addition, $const=\log\alpha_eN_C$, where $N_C$ denotes the number of classes in task, $\alpha_e$ is a constant.
\end{theorem}

For convenience, we define the distance function as the cosine similarity function, though it was defined as Euclidean or Mahalanobias distance functions in previous works~\cite{prototypical,CNAPS}. Cosine similarity has been widely used as distance function~\cite{urt,url,tsa} since it approximates the Euclidean distance function when data are normalized and can be treated as a generalization of Mahalanobis distance computation by decomposing the inverse of the covariance matrix into a lower triangular matrix and its conjugate transpose~\cite{url}. More discussions are available in Appendix~\ref{Appendix:more_related_works}.

The lower bound in Theorem~\ref{Theorem:lower_bound_protoloss} reveals two key insights adopted in the NCC-based loss. On the one hand, the similarities among samples belonging to the same class are maximized via the first term, which facilitates learning class-specific representations for each class in the given task. On the other hand, the similarities between arbitrary two data samples are minimized through the second term. Ideally, the second term potentially helps reduce the similarities derived from trivial information, such as common backgrounds. This further contributes to focusing on the discriminative feature areas. Compared to \citet{sslhsic}, our analysis is conducted on both samples and prototypes instead of samples only. This introduces difficulty in our analysis work. Thus, we propose Lemma~\ref{Lemma:bound_of_JE} to overcome such a problem and then provide a more precise lower bound.

\subsection{HSIC as a Lower Bound of NCC-based Loss}
In this section, we follow \citet{sslhsic} to bridge a connection between NCC-based loss and kernel HSIC measure.
\begin{definition}[Label kernel~\cite{sslhsic}]\label{Def:label_kernel}
Given a support set $\mathcal{D}_{\mathcal{T}}$ that includes $N_C$ classes and $K_n$ shots for each class $n=\{1, 2, ..., N_C\}$, we assume that the label of each data point is a one-hot vector, and each datapoint belonging to class $n$ is sampled from the given support set with the probability $\frac{1}{|\mathcal{D}_{\mathcal{T}}|}$. Then, any kernel that is a function of $y_i^{\top}y_j$ or $||y_i-y_j||$ have the form
\begin{equation}\label{Eq:label_kernel_def}
\begin{small}
l(y_i, y_j)=\left\{
\begin{array}{rcl}
    l_1, &y_i=y_j,\\
    l_0, &otherwise
\end{array}
\right.\equiv \Delta l\mathbb{I}(y_i=y_j)+l_0,
\end{small}
\end{equation}
where $\Delta l=l_1-l_0$.
\end{definition}

\begin{theorem}\label{Theorem:HSIC_Z_Y}
    Given a support representation set $\mathcal{Z}=\{\boldsymbol{z}_i\}_{i=1}^{|\mathcal{D}_{\mathcal{T}}|}=\{h_{\theta}\circ f_{\phi^*}(\boldsymbol{x}_i)\}_{i=1}^{|\mathcal{D}_{\mathcal{T}}|}$ where $N_C$ classes are included, let $k(\cdot, \cdot)$ be a linear kernel function on data representations and $l(\cdot, \cdot)$ be a label kernel defined in Eq.~(\ref{Eq:label_kernel_def}), then ${\rm HSIC}(Z, Y)$ owns a lower bound:
    \[
    \begin{small}
    \begin{aligned}
        {\rm HSIC}(Z, Y) \ge &\frac{\lambda\Delta l}{|\mathcal{D}_{\mathcal{T}}|^2}\sum_{i=1}^{|\mathcal{D}_{\mathcal{T}}|}\sum_{\boldsymbol{z}^{+}\in\mathcal{C}}k(\boldsymbol{z}_i, \boldsymbol{z}^{+}) -\\ &\frac{\lambda\Delta l}{|\mathcal{D}_{\mathcal{T}}|^2}\sum_{i=1}^{|\mathcal{D}_{\mathcal{T}}|}\sum_{\boldsymbol{z}^{'}\in\mathcal{Z}}\frac{1}{|\mathcal{D}_{\mathcal{T}}|}k(\boldsymbol{z}_i, \boldsymbol{z}^{'}),
    \end{aligned}
    \end{small}
    \]
    where $\boldsymbol{z}^{+}$ denotes the data samples belonging to the same class as $\boldsymbol{z}_i$, $\mathcal{C}$ denotes the class that $\boldsymbol{z}_i$ belongs to, $\boldsymbol{z}^{'}$ is an independent copy of $\boldsymbol{z}$, $\lambda$ is a scale constant.
\end{theorem}
\begin{remark}\upshape
Under the setting of few-shot classification task with varied ways and shots, the lower bound of $\mathrm{HSIC}(Z, Y)$ obtained in Theorem~\ref{Theorem:HSIC_Z_Y} shares a similar structure to the lower bound of NCC-based loss obtained in Theorem~\ref{Theorem:lower_bound_protoloss}. Such a phenomenon mainly results from both NCC-based loss and $\mathrm{HSIC}(Z, Y)$ leveraging the label information. Specifically, NCC-based loss implicitly takes advantage of label information in the prototypes while $\mathrm{HSIC}(Z, Y)$ explicitly measures the dependence between representations and labels. 
In addition, an explicit intuition of $\mathrm{HSIC}(Z, Y)$ is that maximizing ${\rm HSIC}(Z, Y)$ is equivalent to guiding models to explore class-specific representations that match the cluster structure indicated by labels.
\end{remark}

Since the constant scaling does not affect the optimization and it is easy to obtain that the high-order moment term satisfies $\mathcal{O}(k(\boldsymbol{z}, \boldsymbol{z}^{'}))\ge \gamma\mathrm{HSIC}(Z, Z)$, where $\gamma=\frac{|\mathcal{D}_{\mathcal{T}}|}{2N_CC_{\rm max}}$, $C_{\rm max}$ is a constant that satisfies $C_{\rm max}\ge |\mathcal{C}_c|$ for $\forall c\in\{1, 2, ..., N_C\}$ (see Appendix~\ref{Appendix:HSICZZ} for more details), we then can build a connection between NCC-based loss and HSIC measure via omitting the scale constants as following: 
\[
    \mathcal{L}(\theta) \ge -\mathrm{HSIC}(Z, Y) + \gamma\mathrm{HSIC}(Z, Z)+const. 
\]

The new lower bound of NCC-based loss is further expressed as a combination of $\mathrm{HSIC}(Z, Y)$ and $\mathrm{HSIC}(Z, Z)$. As revealed in Theorem~\ref{Theorem:HSIC_Z_Y}, the lower bound of ${\rm HSIC}(Z, Y)$ owns the similar structure to that of NCC-based loss and plays the same role to simultaneously maximizes the similarities among samples belonging to the same class and minimizes the similarities between samples from different classes. Moreover, the $\mathrm{HSIC}(Z, Z)$ term measures the dependence among all data samples in the given task. In this case, we can control the dependence between sample representations via scaling $\mathrm{HSIC}(Z, Z)$ term. Based on these, a spontaneous insight into the NCC-based loss is that the lower bound can be adopted as a surrogate loss of NCC-based loss to perform few-shot classification tasks.

\subsection{Perform CFC Tasks with HSIC}

\textbf{Comparison to NCC.} Compared with NCC-based loss, the surrogate loss mentioned above owns the following two desirable merits. First of all, different from NCC-based loss which implicitly and simply leverages label information via class centroids to learn a data cluster for each class, the first term of the surrogate loss explicitly measures the dependence between support data representations and labels that contain cluster structure information of the given set to explore the class-specific representations for each class. Moreover, due to the scarce labeled data for adaptation in few-shot classification tasks, the model usually overfits the data and in turn obtains poor generalization performance. However, such an undesirable phenomenon can be mitigated through scaling $\mathrm{HSIC}(Z, Z)$ term which measures the dependence among all data samples in a given task.

\textbf{Challenges When Using HSIC.} A challenge of applying HSIC to perform cross-domain few-shot classification tasks is that kernel HSIC may fail to accurately measure the dependence between two data samples. 
For example, as shown in Fig.~\ref{Fig:intro_imagenet_pa}, although the similarities among samples within the same class are evidently higher, there still exist high similarities between two samples from different classes. This phenomenon implies that NCC-based loss may fail to accurately measure the dependence between representations and labels, and in turn, fails to learn a set of class-specific representations that match the cluster structure of the support set. Such a phenomenon may potentially induce undesirable uncertainty and further result in the misclassification of samples. To address this problem, a feasible way is improving the capability of the kernel HSIC such that a set of discriminative class-specific representations can be learned.

\vspace{-0.2cm }
\section{Maximizing Optimized Kernel Dependence}
In order to solve the above problem, we propose \emph{maximizing optimized kernel dependence} (MOKD) to perform cross-domain few-shot classification with kernel HSIC measures in which the kernels are optimized to accurately measure the dependence. Specifically, we first maximize the test power of the kernels used in HSIC to improve its capability in dependence detection, and then optimize the dependence respectively between representations and labels, and among representations with the optimized kernel HSIC measures. Intuitively, test power maximization facilitates increasing the sensitivity of kernel HSIC measures to the dependence. In this way, the dependence among samples can be more accurately measured and effectively optimized. In turn, we can learn a set of class-specific and discriminative representations where the similarities among samples belonging to the same class are maximized while the similarities between two samples belonging to different classes are minimized.

\subsection{Backbone Pre-training}
In this paper, we follow URL~\cite{url} and adopt ResNet-18 as the backbone. The pre-training strategy is consistent with that in URL. Specifically, we first pre-train 8 domain-specific backbones respectively for the 8 seen domain datasets. Specifically, in the ``train on ImageNet only'' setting, the backbone pre-trained on the ImageNet dataset is adopted for embedding extraction. Then, a universal backbone is distilled from all domain-specific backbones. Specifically, assume that $\mathcal{D}^{\rm tr}_{\mathcal{S}^{\rm seen}_\tau}$, where $\tau\in\{1, 2, ..., |\mathcal{S}^{\rm seen}|\}$, denotes the training set of the dataset $\mathcal{S}^{\rm seen}_{\tau}$, the distillation of the universal backbone can be formulated as:
\[
\begin{small}
    \min_{\phi, \omega}\sum_{\tau=1}^{|\mathcal{S}^{\rm seen}|}\frac{1}{|\mathcal{D}^{\rm tr}_{\mathcal{S}^{\rm seen}_{\tau}}|}\sum^{|\mathcal{D}^{\rm tr}_{\mathcal{S}^{\rm seen}_{\tau}}|}_{i=1}\ell\left(g_{\omega_{\tau}}\circ f_{\phi}(\boldsymbol{x}_i), y_i\right)+\zeta\mathcal{R}(\phi, \omega_{\tau}),
\end{small}
\]
where $f_{\phi}$ is the universal backbone parameterized with $\phi$, $g_{\omega_{\tau}}$ is the classifier for domain $\mathcal{S}^{\rm seen}_{\tau}$ parameterized with $\omega_{\tau}$, $\mathcal{R}(\cdot, \cdot)$ is the regularization and $\zeta$ is the coefficient. To be specific, in URL~\cite{url}, the regularization is composed of two losses that respectively aim at minimizing the distance between the predictions and maximizing the similarity of representations between the distilled universal backbone and the corresponding domain-specific backbone.

\begin{table*}[t]
    \centering
	\caption{\centering\textbf{Results on Meta-Dataset (Trained on ImageNet Only).} Mean accuracy and 95\% confidence interval are reported.}
	\label{Table:imagenet_only}
	\begin{center}
		\begin{tiny}
			\setlength\tabcolsep{6.7pt}
			\begin{threeparttable}
				\begin{tabular}{lccccccccc|c}
					\toprule
					Datasets & Finetune & ProtoNets & ProtoNets(large) & BOHB & FP-MAML & ALFA+FP-MAML & FLUTE & SSL-HSIC& URL &\bf{MOKD(Ours)}\\
					\midrule
					\rowcolor{green!7}ImageNet    	 &45.8$\pm$1.1	& 50.5$\pm$1.1  & 53.7$\pm$1.1   & 51.9$\pm$1.1	& 49.5$\pm$1.1	& 52.8$\pm$1.1	& 46.9$\pm$1.1	           & 55.5$\pm$1.1  & \bf{57.3$\pm$1.1}     & \bf{57.3$\pm$1.1} \\
					\midrule
					\rowcolor{red!7}Omniglot     	  & 60.9$\pm$1.6	& 60.0$\pm$1.4   & 68.5$\pm$1.3   & 67.6$\pm$1.2	& 63.4$\pm$1.3	& 61.9$\pm$1.5	&61.6$\pm$1.4           & 66.4$\pm$1.2  & 69.4$\pm$1.2	        & \bf{70.9$\pm$1.3}\\
					\rowcolor{red!7}Aircraft    		  & \bf{68.7$\pm$1.3}	& 53.1$\pm$1.0   & 58.0$\pm$1.0   & 54.1$\pm$0.9	& 56.0$\pm$1.0	& 63.4$\pm$1.1	& 48.5$\pm$1.0	& 49.5$\pm$0.9  & 57.6$\pm$1.0	        & \bf{59.8$\pm$1.0}\\
					\rowcolor{red!7}Birds    		      & 57.3$\pm$1.3	& 68.8$\pm$1.0   & \bf{74.1$\pm$0.9}   & 70.7$\pm$0.9	& 68.7$\pm$1.0	& 69.8$\pm$1.1	& 47.9$\pm$1.0	    & 71.6$\pm$0.9   & 72.9$\pm$0.9	        & \bf{73.6$\pm$0.9}\\
					\rowcolor{red!7}Textures     	   & 69.0$\pm$0.9	& 66.6$\pm$0.8   & 68.8$\pm$0.8   & 68.3$\pm$0.8	& 66.5$\pm$0.8	& 70.8$\pm$0.9	& 63.8$\pm$0.8	        & 72.2$\pm$0.7   & 75.2$\pm$0.7	    & \bf{76.1$\pm$0.7}\\
					\rowcolor{red!7}Quick Draw       & 42.6$\pm$1.2	& 49.0$\pm$1.1   & 53.3$\pm$1.0  & 50.3$\pm$1.0	& 51.5$\pm$1.0	& 59.2$\pm$1.2	& 57.5$\pm$1.0	           & 54.2$\pm$1.0    & 57.9$\pm$1.0	        & \bf{61.2$\pm$1.0}\\
					\rowcolor{red!7}Fungi      			   & 38.2$\pm$1.0	& 39.7$\pm$1.1   & 40.7$\pm$1.2   & 41.4$\pm$1.1	& 40.0$\pm$1.1	& 41.5$\pm$1.2	& 31.8$\pm$1.0	    & 43.4$\pm$1.1   & 46.2$\pm$1.0	    & \bf{47.0$\pm$1.1}\\
					\rowcolor{red!7}VGG Flower     & 85.5$\pm$0.7	& 85.3$\pm$0.8   & 87.0$\pm$0.7   & 87.3$\pm$0.6	&87.2$\pm$0.7	& 86.0$\pm$0.8	&80.1$\pm$0.9	            & 85.5$\pm$0.7    & 86.9$\pm$0.6	        & \bf{88.5$\pm$0.6}\\
					\rowcolor{red!7}Traffic Sign      & \bf{66.8$\pm$1.3}	& 47.1$\pm$1.1   & 58.1$\pm$1.1   & 51.8$\pm$1.0	& 48.8$\pm$1.1	& 60.8$\pm$1.3	& 46.5$\pm$1.1	   & 50.5$\pm$1.1    & 61.2$\pm$1.2	        & \bf{61.6$\pm$1.1}\\
					\rowcolor{red!7}MSCOCO         & 34.9$\pm$1.0	& 41.0$\pm$1.1   & 41.7$\pm$1.1  & 48.0$\pm$1.0	& 43.7$\pm$1.1	& 48.1$\pm$1.1	& 41.4$\pm$1.0	           & 51.4$\pm$1.0    & 53.0$\pm$1.0	        & \bf{55.3$\pm$1.0}\\
					\rowcolor{red!7}MNIST              & - & -  &-	&- & -	& -	&80.8$\pm$0.8  &77.0$\pm$0.7   & 86.2$\pm$0.7	        & \bf{88.3$\pm$0.7}\\
					\rowcolor{red!7}CIFAR-10         & - & - &-	& - & -	& -	& 65.4$\pm$0.8	& 71.0$\pm$0.8    & 69.5$\pm$0.8     & \bf{72.2$\pm$0.8}\\
					\rowcolor{red!7}CIFAR-100       & - & - &-	& -	& - & -	& 52.7$\pm$1.1	  & 59.0$\pm$1.0   & 62.0$\pm$1.0	        & \bf{63.1$\pm$1.0}\\
					\midrule
					Average Seen & 45.8	& 50.5 & 53.7 & 51.9	& 49.5	& 52.8	&46.9  & 55.5	& \bf{57.3}	& \bf{57.3}\\
					Average Unseen & - & - &-	& -	& - & -	& 56.5	& 62.5  & 66.6	& \bf{68.1}\\
					Average All & - & - & -	& -	& -	& -	& 55.8	& 62.0 & 65.9	&\bf{67.3}\\ 
					\midrule
					Average Rank  &	7.1	& 8.4	& 4.6	& 5.5	& 6.8	& 4.4 & 8.9 & 4.9 & 2.8&\bf{1.4}\\
					\bottomrule
				\end{tabular}
				\begin{tablenotes}
					\item[1] The results on URL and MOKD are the average of 5 reproductions with different random seeds. 
				\end{tablenotes} 
			\end{threeparttable}
		\end{tiny}
	\end{center}
	\vspace{-2.0em}
\end{table*}

\begin{table*}[t]
	\caption{\textbf{Results on Meta-Dataset (Trained on All Datasets).} Mean accuracy and 95\% confidence interval are reported.}
	\label{Table:all_datasets}
	\begin{center}
		\begin{tiny}
			\setlength\tabcolsep{6pt}
			\begin{threeparttable}
				\begin{tabular}{lcccccccccc|c}
					\toprule
					Datasets & ProtoMAML & CNAPS & S-CNAPS & SUR & URT & Tri-M & FLUTE  & 2LM & SSL-HSIC & URL &\bf{MOKD}\\
					\midrule
					\rowcolor{green!7}ImageNet    	 & 46.5$\pm$ 1.1	& 50.8$\pm$1.1		& 58.4 $\pm$1.1 	& 56.2 $\pm$ 1.0 	& 56.8 $\pm$ 1.1 	& \bf{58.6 $\pm$ 1.0}   & 51.8 $\pm$ 1.1       & 58.0 $\pm$ 3.6 	  & 56.5 $\pm$ 1.2	   & 57.3 $\pm$ 1.1		& 57.3 $\pm$ 1.1 \\
					\rowcolor{green!7}Omniglot     	  & 82.7$\pm$ 1.0	 & 91.7$\pm$0.5 	& 91.6 $\pm$ 0.6 		  & 94.1 $\pm$ 0.4 	  & 94.2 $\pm$ 0.4   & 92.0 $\pm$ 0.6   & 93.2 $\pm$ 0.5		  & \bf{95.3 $\pm$ 1.0} 	 & 92.0 $\pm$ 0.9      & 94.1 $\pm$ 0.4      & \bf{94.2 $\pm$ 0.5} \\
					\rowcolor{green!7}Aircraft    		 & 75.2$\pm$ 0.8	& 83.7$\pm$0.6	  & 82.0 $\pm$ 0.7 		   & 85.5 $\pm$ 0.5    & 85.8 $\pm$ 0.5   & 82.8 $\pm$ 0.7   & 87.2 $\pm$ 0.5		  & 88.2 $\pm$ 0.5 	 & 87.3 $\pm$ 0.7	   & 88.2 $\pm$ 0.5      & \bf{88.4 $\pm$ 0.5} \\
					\rowcolor{green!7}Birds    		      & 69.9$\pm$ 1.0	& 73.6$\pm$0.9	   & 74.8 $\pm$ 0.9 		& 71.0 $\pm$ 1.0 	 & 76.2 $\pm$ 0.8    & 75.3 $\pm$ 0.8   & 79.2 $\pm$ 0.8  & \bf{81.8 $\pm$ 0.6} 	& 78.1 $\pm$ 1.1   & 80.2 $\pm$ 0.7      & \bf{80.4 $\pm$ 0.8}\\
					\rowcolor{green!7}Textures     	   & 68.2$\pm$ 1.0 	 & 59.5$\pm$0.7		& 68.8 $\pm$ 0.9 		 & 71.0 $\pm$ 0.8 	 & 71.6 $\pm$ 0.7     & 71.2 $\pm$ 0.8   & 68.8 $\pm$ 0.8  & 76.3 $\pm$ 2.4 		& 75.2 $\pm$ 0.8   & 76.2 $\pm$ 0.7      & \bf{76.5 $\pm$ 0.7} \\
					\rowcolor{green!7}Quick Draw      & 66.8$\pm$ 0.9   & 74.7$\pm$0.8	 & 76.5 $\pm$0.8  		  & 81.8 $\pm$ 0.6 	  & \bf{82.4 $\pm$ 0.6}    & 77.3 $\pm$ 0.7   & 79.5 $\pm$ 0.7  & 78.3 $\pm$ 0.7    & 81.4 $\pm$ 0.7	   & 82.2 $\pm$ 0.6      & 82.2 $\pm$ 0.6 \\
					\rowcolor{green!7}Fungi      			  & 42.0$\pm$1.2 	& 50.2$\pm$1.1		& 46.6 $\pm$ 1.0 		 & 64.3 $\pm$ 0.9 	 & 64.0 $\pm$ 1.0    & 48.5 $\pm$ 1.0   & 58.1 $\pm$ 1.1	      & \bf{69.6 $\pm$ 1.5}  & 63.5 $\pm$ 1.2 & 68.7 $\pm$ 1.0      & 68.6 $\pm$ 1.0 \\
					\rowcolor{green!7}VGG Flower     & 88.7$\pm$ 0.7 	  & 88.9$\pm$0.5	& 90.5 $\pm$ 0.5 		 & 82.9 $\pm$ 0.8 	 & 87.9 $\pm$ 0.6    & 90.5 $\pm$ 0.5   & 91.6 $\pm$ 0.6  & 90.3 $\pm$ 0.8 		& 90.9 $\pm$ 0.8   & 91.9 $\pm$ 0.5      & \bf{92.5 $\pm$ 0.5} \\
					\midrule
					\rowcolor{red!7}Traffic Sign      & 52.4 $\pm$ 1.1 	 & 56.5 $\pm$1.1     & 57.2 $\pm$ 1.0		   & 51.0 $\pm$ 1.1		 & 48.2 $\pm$ 1.1     & 63.0 $\pm$ 1.0   & 58.4 $\pm$ 1.1		  & 63.6 $\pm$ 1.5	& 59.7 $\pm$ 1.3   & 63.3 $\pm$ 1.2      &\bf{64.5 $\pm$ 1.1}\\
					\rowcolor{red!7}MSCOCO         & 41.7 $\pm$ 1.1 	& 39.4 $\pm$1.0   & 48.9 $\pm$ 1.1			& 52.0 $\pm$ 1.1	  & 51.5 $\pm$ 1.1      & 52.8 $\pm$ 1.1   & 50.0 $\pm$ 1.0		  & \bf{57.0 $\pm$ 1.1}	  & 51.4 $\pm$ 1.1  & 54.2 $\pm$ 1.0      &\bf{55.5 $\pm$ 1.0}\\
					\rowcolor{red!7}MNIST              & - 				& -     		& 94.6 $\pm$ 0.4			 & 94.3 $\pm$ 0.4 	 & 90.6 $\pm$ 0.5      & \bf{96.2 $\pm$ 0.3}   & \bf{95.6 $\pm$ 0.5}		  & 94.7 $\pm$ 0.5  & 93.4 $\pm$ 0.6 & 94.7 $\pm$ 0.4      &\bf{95.1 $\pm$ 0.4}\\
					\rowcolor{red!7}CIFAR-10         & - 				& - 		& 74.9 $\pm$ 0.7			& 66.5 $\pm$ 0.9	& 67.0 $\pm$ 0.8	& 75.4 $\pm$ 0.8   & \bf{78.6 $\pm$ 0.7}		  & 71.5 $\pm$ 0.9		& 70.0 $\pm$ 1.1   & 71.9 $\pm$ 0.8      &\bf{72.8 $\pm$ 0.8}\\
					\rowcolor{red!7}CIFAR-100       & -			 & - 				& 61.3 $\pm$ 1.1			& 56.9 $\pm$ 1.1	  & 57.3 $\pm$ 1.0    & 62.0 $\pm$ 1.0   & \bf{67.1 $\pm$ 1.0}		  & 60.0 $\pm$ 1.1		   & 61.8 $\pm$ 1.1 & 62.9 $\pm$ 1.0      &\bf{63.9 $\pm$1.0}\\
					\midrule
					Average Seen & 67.5 					& 71.6						& 73.7								& 75.9						 & 77.4						  &76.2   & 76.2		& 79.7 & 76.5 & 79.9		&\bf{80.0}\\
					Average Unseen & -  					 & - 							& 67.4							  & 64.1						& 62.9						 & 69.9   & 69.9		& 69.4 & 68.2 & 69.4		& \bf{70.3}\\
					Average All & - 							  & - 							 & 71.2								 & 71.3						   & 71.8   & 73.8						& 73.8		& 75.7 & 74.6 & 75.8		& \bf{76.3}\\ 
					\midrule
					Average Rank &	-	& -	& 7.2	& 7.3	& 6.4	& 5.2 & 5.2 & 3.4 & 5.5 & 3.1	&\bf{2.2} \\
					\bottomrule
				\end{tabular}
				\begin{tablenotes}
					\item[1] Results of URL are the average of 5 reproductions with different random seeds. The reproductions are consistent with the results reported on their \href{https://groups.inf.ed.ac.uk/vico/research/URL/}{website}. The results of our method are the average of 5 random reproduction experiments. The ranks considers all 13 datasets and are calculated only with the methods in the table.
				\end{tablenotes} 
			\end{threeparttable}
		\end{tiny}
	\end{center}
	\vskip -0.1in
 \vspace{-0.5em}
\end{table*}
\begin{table*}[t]
\vspace{-1.0em}
	\caption{\textbf{Comparisons of MOKD with different characteristic kernels.}}
	\label{Table:kernel_type}
	\begin{center}
		\begin{tiny}
		\setlength\tabcolsep{4pt}
			\begin{threeparttable}
				\begin{tabular}{lccccccccccccc}
					\toprule
					Datasets & ImageNet & Omniglot & Aircraft & Birds & DTD & QuickDraw & Fungi & VGG\_Flower & Traffic Sign & MSCOCO & MNIST & CIFAR10 & CIFAR100 \\
					\midrule
					Gaussian  & 57.3$\pm$1.1 & 94.2$\pm$0.5 & \bf{88.4$\pm$0.5} & 80.4$\pm$0.8 & \bf{76.5$\pm$0.7} & 82.2$\pm$0.6 & \bf{68.6$\pm$1.0} & \bf{92.5$\pm$0.5} & \bf{64.5$\pm$1.1} & \bf{55.5$\pm$1.0} & 95.1$\pm$0.4 & 72.8$\pm$0.8 & \bf{63.9$\pm$1.0}\\
					IMQ       & 57.3$\pm$1.1 & 94.3$\pm$0.5 & 88.0$\pm$0.5 & \bf{80.5$\pm$0.8} & 76.2$\pm$0.7 & 82.3$\pm$0.6 & 67.7$\pm$1.0 & 92.1$\pm$0.5 & 63.8$\pm$1.1 & 54.8$\pm$1.0 & \bf{95.4$\pm$0.4} & 72.7$\pm$0.8 & 63.7$\pm$1.0\\
					\bottomrule
				\end{tabular}
			\end{threeparttable}
		\end{tiny}
	\end{center}
	\vskip -0.1in
	\vspace{-0.5em}
\end{table*}
\subsection{Problem Formulation for MOKD}
Consider a set of support representations pairs $\{(\boldsymbol{z}_i, y_i)\}_{i=1}^{m}$, where $m$ denotes the size of the set, the ultimate goal of MOKD is to learn a set of optimal task-specific parameters $\theta^*$ from the given data by performing optimization on the optimized kernel HSIC measures where the test power is maximized to increase their sensitivity to dependence. Thus, MOKD is formulated as a bi-level optimization problem:
\begin{equation}\label{Eq:mokd_problem}
\begin{small}
\begin{aligned}
    &\min_{\theta} -{\rm HSIC}(Z, Y; \sigma_{ZY}^*, \theta) + \gamma{\rm HSIC}(Z, Z; \sigma_{ZZ}^*, \theta),\\
    s.t. &\max_{\sigma_{ZY}}\frac{{\rm HSIC}(Z, Y; \sigma_{ZY}, \theta)}{\sqrt{v_{ZY}+\epsilon}}, \max_{\sigma_{ZZ}}\frac{{\rm HSIC}(Z, Z; \sigma_{ZZ}, \theta)}{\sqrt{v_{ZZ}+\epsilon}},
\end{aligned}
\end{small}
\end{equation}
where $\sigma_{ZY}$ and $\sigma_{ZZ}$ are the bandwidths of Gaussian kernels respectively calculated in ${\rm HSIC}(Z, Y)$ and ${\rm HSIC}(Z, Z)$, $v_{ZY}$ and $v_{ZZ}$ are the variances of estimated ${\rm HSIC}(Z, Y)$ and ${\rm HSIC}(Z, Z)$, $\gamma$ is the scalar coefficient of ${\rm HSIC}(Z, Z)$ term and $\epsilon$ is a scalar that is used to avoid the case $v\leq 0$. 

Since the true distribution of support data features is unknown, in this paper, we follow~\citet{hsic_estimation} and estimate the kernel HSIC, $\widehat{{\rm HSIC}}(Z, Y)$, with a set of finite data samples in an unbiased way:
\begin{equation}\label{Eq:HSIC_estimation_song}
\begin{small}
    \frac{1}{m(m-3)}\left[{\rm tr}\left(\boldsymbol{\Tilde{K}}\boldsymbol{\Tilde{L}}\right)+\frac{\boldsymbol{1}^{\top}\boldsymbol{\Tilde{K}}\boldsymbol{1}\boldsymbol{1}^{\top}\boldsymbol{\Tilde{L}}\boldsymbol{1}}{(m-1)(m-2)}-\frac{2\boldsymbol{1}^{\top}\boldsymbol{\Tilde{K}}\boldsymbol{\Tilde{L}}\boldsymbol{1}}{m-2}\right],
\end{small}
\end{equation}
where $\boldsymbol{\Tilde{K}}$ and $\boldsymbol{\Tilde{L}}$ are kernel matrices where $\boldsymbol{\Tilde{K}}_{i,j}=(1-\delta_{i,j})k(\boldsymbol{z}_i, \boldsymbol{z}_j)$ and $\boldsymbol{\Tilde{L}}_{i,j}=(1-\delta_{i,j})l(y_i, y_j)$, $m$ denotes the number of samples in support set. In practice, the time complexity of the calculation of $\widehat{{\rm HSIC}}(Z, Y)$ is $\mathcal{O}(m^2)$.

The bi-level optimization problem proposed in Eq.~(\ref{Eq:mokd_problem}) mainly contains two aspects: inner optimization for test power maximization and outer optimization for dependence optimization. To be specific, during the inner optimization phase, MOKD performs optimization on the kernel HSIC to maximize its test power via maximizing $\frac{{\rm HSIC}(\cdot, \cdot; \sigma, \theta)}{\sqrt{v+\epsilon}}$. In this way, the optimized kernel HSIC measures are more sensitive to dependence and in turn, the dependence among data samples can be more precisely measured. 
Moreover, during the outer optimization, with the optimized kernel HSIC measures, MOKD maximizes the dependence between representations and labels to explore a set of class-specific representations, where the similarities among samples within the same class are maximized while the similarities between samples belonging to different classes are minimized, to match the cluster structures of the given support set. Meanwhile, the dependence among all representations is minimized as a regularization to penalize the high-variance representations for alleviating the overfitting derived from scarce data.

\textbf{Differences between SSL-HSIC and MOKD.} We notice that the outer optimization objective in Eq.~(\ref{Eq:mokd_problem}) shares the similar format as that in SSL-HSIC~\citep{sslhsic}. From our perspective, in fact, there are two major differences between them. 
Firstly, the most obvious difference between SSL-HSIC and MOKD is that MOKD takes the test power into consideration. This facilitates increasing the sensitivity of the kernel HSIC to the dependence and further contributes to the dependence optimization. 
In addition, SSL-HSIC is derived from unsupervised contrastive learning and focuses on learning discriminative features by contrasting two different views of a sample. However, MOKD is derived from supervised few-shot classification and aims at learning a set of class-specific features where similarities among samples within the same class are maximized while similarities between samples from different classes are minimized. More details and empirical results are available in Appendix~\ref{Appendix:difference_sslhsic_mokd}.

\vspace{-0.2cm}
\subsection{Bandwidth Selection for Test Power Maximization}
In our proposed MOKD method, since dependence optimization significantly depends on the kernels' capability of precise dependence measuring, we first propose to optimize the kernel HSIC measures to improve their capability of detecting dependence between two variables by maximizing their test power.
According to the definition of the test power mentioned above, maximizing the test power of a kernel HSIC measure is equivalent to maximizing the $\frac{{\rm HSIC}(\cdot, \cdot; \sigma, \theta)}{\sqrt{v+\epsilon}}$ term. In practice, we adopt a Gaussian kernel, which contains a bandwidth $\sigma$, as the kernel function in the Problem~\eqref{Eq:mokd_problem}. Thus, test power maximization can be further reformulated as finding an optimal bandwidth $\sigma^*$ for the Gaussian kernel function to maximize the $\frac{{\rm HSIC}(\cdot, \cdot; \sigma, \theta)}{\sqrt{v+\epsilon}}$ term.

A key step of performing test power maximization is estimating the variance of ${\rm HSIC}(\cdot, \cdot, \sigma, \theta)$ with the given bandwidth $\sigma$. According to Theorem 5 proposed by~\citet{hsic_estimation}, $\widehat{{\rm HSIC}}$, which is estimated in an unbiased way following Eq.~(\ref{Eq:HSIC_estimation_song}), converges in distribution to a Gaussian random variable with the mean ${\rm HSIC}$ and the estimated variance:
\begin{equation}\label{Eq:variance_hsic_estimation}
\begin{small}
    v = \frac{16}{m}\left(R-\widehat{{\rm HSIC}}^2\right), R = (4m)^{-1}(m-1)^{-2}_3\boldsymbol{h}^{\top}\boldsymbol{h},
\end{small}
\end{equation}
where $m$ denotes the number of samples, $(m-1)_3$ denotes the Pochhammer symbols $\frac{(m-1)!}{(m-4)!}$, and $\boldsymbol{h}$ is a basic vector for the calculation of $R$, and can be calculated as:
\begin{equation}
\begin{small}
\begin{aligned}
    \boldsymbol{h} = &(m-2)^2\left(\boldsymbol{\Tilde{K}}\circ \boldsymbol{\Tilde{L}}\right)\boldsymbol{1} - m(\boldsymbol{\Tilde{K}}\boldsymbol{1})\circ(\boldsymbol{\Tilde{L}}\boldsymbol{1})\\
    & + (\boldsymbol{1}^{\top}\boldsymbol{\Tilde{L}}\boldsymbol{1})\boldsymbol{\Tilde{K}}\boldsymbol{1} + (\boldsymbol{1}^{\top}\boldsymbol{\Tilde{K}}\boldsymbol{1})\boldsymbol{\Tilde{L}}\boldsymbol{1} - (\boldsymbol{1}^{\top}\boldsymbol{\Tilde{K}}\boldsymbol{\Tilde{L}}\boldsymbol{1})\boldsymbol{1}\\
    &+(m-2)\left(({\rm tr\boldsymbol{\Tilde{K}}\boldsymbol{\Tilde{L}}})\boldsymbol{1}-\boldsymbol{\Tilde{K}}\boldsymbol{\Tilde{L}}\boldsymbol{1}-\boldsymbol{\Tilde{L}}\boldsymbol{\Tilde{K}}\boldsymbol{1}\right),
\end{aligned}
\end{small}
\end{equation}
where $\circ$ denotes elementwise multiplication on matrices, $\boldsymbol{1}\in\mathbb{R}^{m\times 1}$ denotes the vector where all elements are 1.

The maximization of test power can be performed by any optimizer, such as gradient-based optimizers. However, in practice, we perform test power maximization via selecting the optimal bandwidth from a list of candidates in the way of grid search~\citep{jitkrittum2016interpretable} since optimizing the bandwidth with optimizers requires extra hyperparameter selection (e.g., learning rate) and gradient steps, which are time-consuming and may negatively affect the efficiency. To some extent, the grid search can be treated as a variant version of the typical bi-level optimization framework (Eq.~\eqref{Eq:mokd_problem}). Specifically, selecting bandwidth with grid search resembles performing ``stop gradient'' on the Problem~\eqref{Eq:mokd_problem}. Although the typical bi-level optimization facilitates exploring higher-order information, such as the Hessian matrix, it will consume more computational resources. Meanwhile, as demonstrated by previous works (e.g. MAML~\cite{maml}) first-order information is sufficient to achieve good performance without hurting the efficiency of the algorithm.

Generally, the bandwidth selection of kernel HSIC mainly includes two steps. Firstly, the bandwidth $\sigma$ is initialized as the median of the non-zero elements of a kernel matrix. Meanwhile, a list of coefficients, which covers as many potential values as possible, is manually set to scale the median. Then, the bandwidth selection is respectively performed for both ${\rm HSIC}(Z, Y)$ and ${\rm HSIC}(Z, Z)$ by selecting the optimal scale coefficient to generate a scaled bandwidth that is able to maximize $\frac{{\rm HSIC}(\cdot, \cdot; \sigma, \theta)}{\sqrt{v+\epsilon}}$. 
In practice, we directly apply the optimal coefficient of ${\rm HSIC}(Z, Y)$ to ${\rm HSIC}(Z, Z)$. A complete process of MOKD is summarized in Algorithm~\ref{Alg:MOKD}. 

\vspace{-0.2cm}
\section{Experiments}
\vspace{-0.2cm}
In this section, we evaluate our proposed MOKD method on the representative mainstream benchmark Meta-Dataset~\citep{metadatasets} under several task settings in order to answer the following questions: (1). Does MOKD achieve better empirical performance on Meta-Dataset under different task settings? (2). What roles do test power maximization and ${\rm HSIC}(\boldsymbol{Z}, \boldsymbol{Z})$ play in MOKD? (3). Is MOKD efficient? (4). Does MOKD facilitate alleviating the high similarity problem and further learning better clusters?

In this paper, we follow most settings in URL~\citep{url} to train a simple linear head on top of a pre-trained ResNet-18 backbone by initializing it as an identity matrix for each adaptation episode and optimizing it with Adadelta~\citep{adadelta}. In order to validate the performance of MOKD, we compare MOKD with existing state-of-the-art approaches, including Proto-MAML~\citep{metadatasets}, fo-Proto-MAML~\citep{metadatasets}, ALFA+fo-Proto-MAML~\citep{alpha} CNAPS~\citep{CNAPS}, SimpleCNAPS (S-CNAPS)~\citep{SimpleCNAPS}, SUR~\citep{SUR}, URT~\citep{urt}, FLUTE~\citep{FLUTE}, Tri-M~\citep{trim}, 2LM~\citep{2LM} and URL~\citep{url}. More details are available in Appendix~\ref{Appendix:detailed_settings} and~\ref{section:more_experimental_details}.

\begin{figure}
    \centering
    \setlength{\abovecaptionskip}{0cm}
    \subfigure[Running time of MOKD \label{Fig:running_time}]{
			\begin{minipage}[t]{0.483 \linewidth}
				\centering
				\includegraphics[width=1\linewidth]{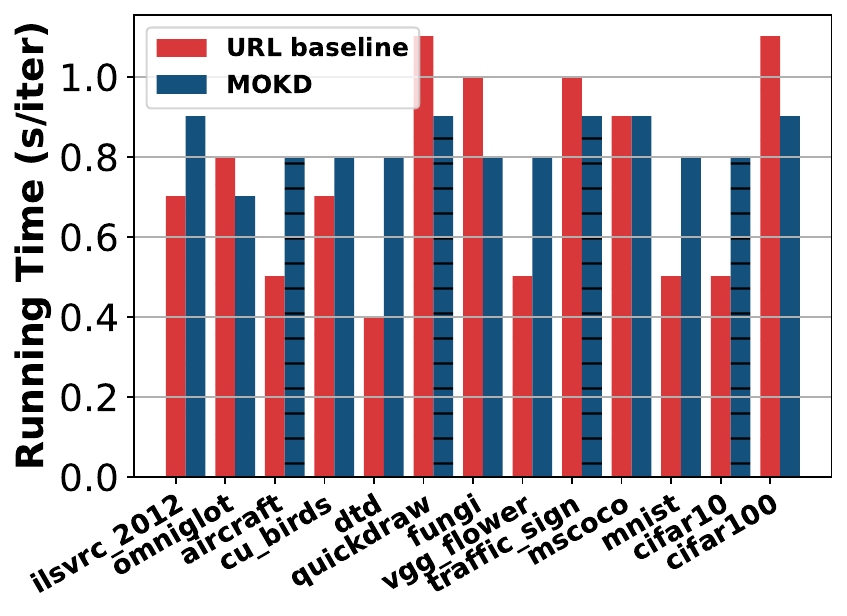}
        \end{minipage}}\vspace{-0.1cm}
    \subfigure[Ablation study on TPM \label{Fig:ablation_tpm}]{
			\begin{minipage}[t]{0.483\linewidth}
				\centering
				\includegraphics[width=1\linewidth]{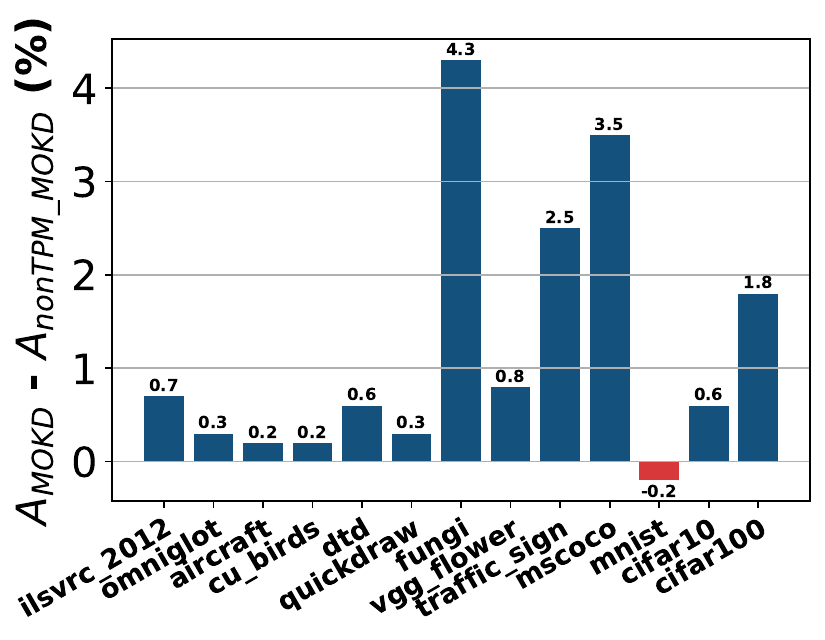}
        \end{minipage}}\vspace{-0.1cm}
    \vspace{-0.2em}
    \caption{\textbf{Results of analysis on MOKD.} \textbf{(a).} Comparison results of running time between MOKD and URL. \textbf{(b).} Accuracy gaps between MOKD with and without test power maximization.}
    \vspace{-2em}
    \label{fig:analysis}
\end{figure}
\begin{figure*}[t]
	\vskip 0.0in
	\begin{center}
	\centering
    \subfigure[$\gamma$ on ImageNet \label{Fig:analysis_imagenet_gamma}]{
			\begin{minipage}[t]{0.24\linewidth}
				\centering
				\includegraphics[width=0.9\linewidth]{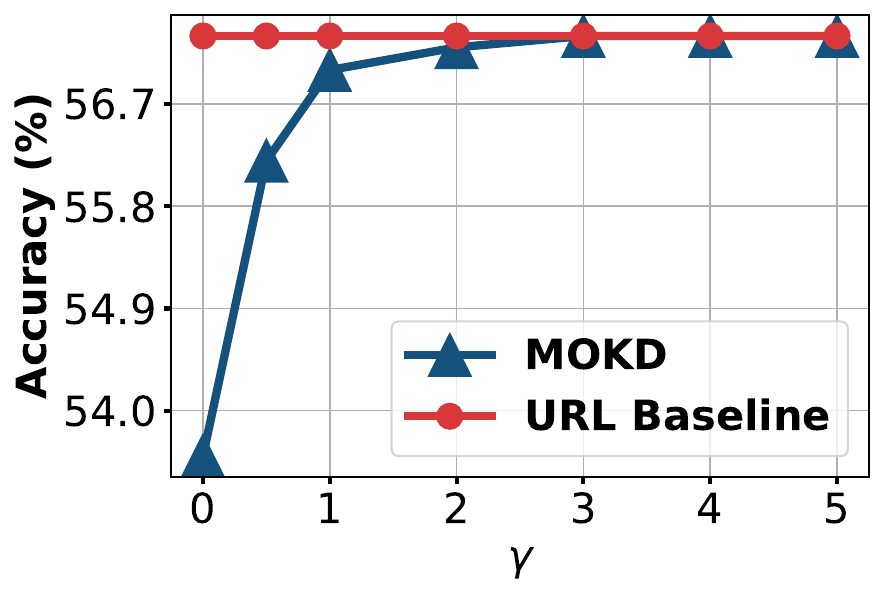}
        \end{minipage}}\vspace{-0.1cm}
    \subfigure[$\gamma$ on MNIST \label{Fig:analysis_mnist_gamma}]{
			\begin{minipage}[t]{0.24\linewidth}
				\centering
				\includegraphics[width=0.9\linewidth]{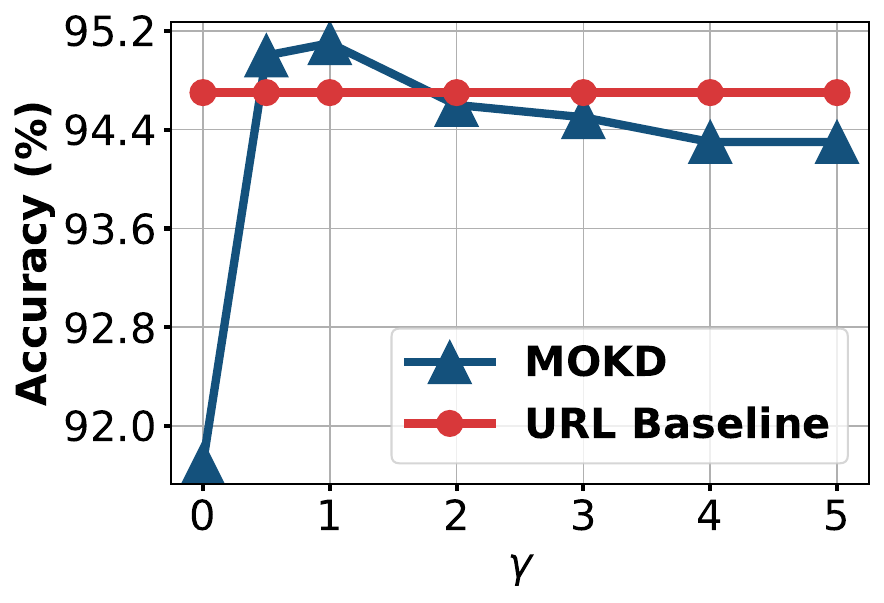}
        \end{minipage}}\vspace{-0.1cm}
	\subfigure[Accuracy gaps\label{Fig:ablation_gamma}]{
			\begin{minipage}[t]{0.23\linewidth}
				\centering
				\includegraphics[width=0.9\linewidth]{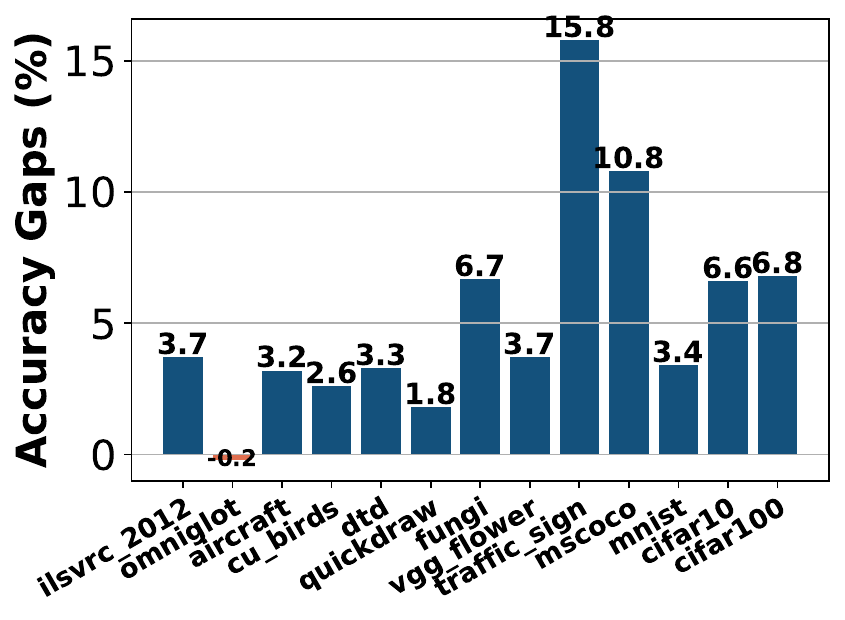}
		\end{minipage}}\vspace{-0.1cm}
    \subfigure[Accuracy curves \label{Fig:test_curves}]{
			\begin{minipage}[t]{0.23\linewidth}
				\centering
				\includegraphics[width=1.0\linewidth]{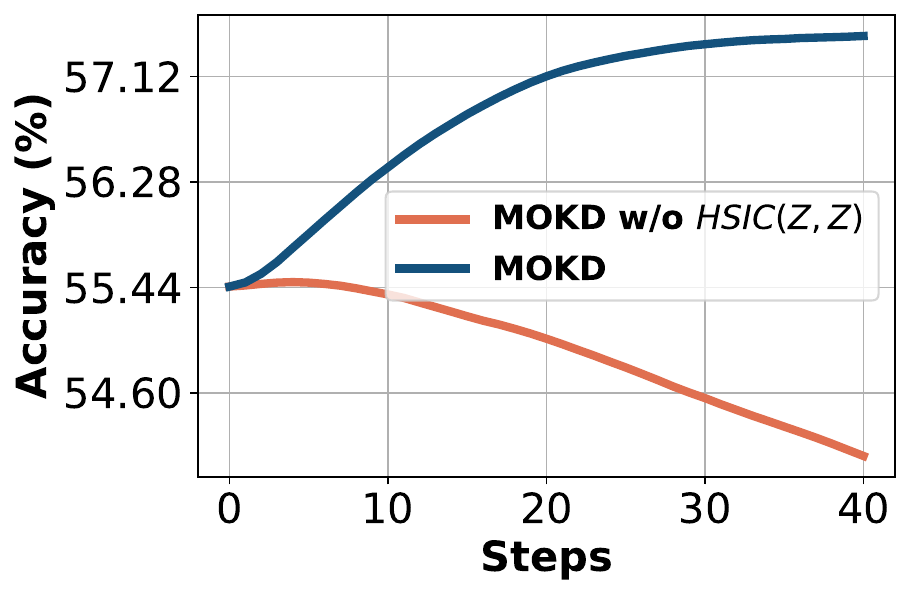}
		\end{minipage}}\vspace{-0.1cm}	
    \caption{\textbf{Quantitative analysis of $\gamma$.} \textbf{(a).} Effect of $\gamma$ on accuracy of ImageNet dataset; \textbf{(b).}  Effect of $\gamma$ on accuracy of MNIST dataset; \textbf{(c).} Performance gaps between MOKD w / w.o. ${\rm HSIC}(Z, Z)$; \textbf{(d).} Test accuracy curves of MOKD w. / w.o. ${\rm HSIC}(Z, Z)$ on ImageNet.}
    \label{Fig:effect_gamma}
	\end{center}
	\vskip -0.2in
	\vspace{-0.3em}
\end{figure*}

\vspace{-0.6em}
\subsection{Main Results}
In this section, we evaluate MOKD on vary-way vary-shot tasks under both ``train on all datasets'' and ``train on ImageNet only'' settings. To be clear, we mark seen domains with green while unseen domains with red. More details about task settings are available in Appendix~\ref{Appendix:task_settings} and~\ref{Appendix:vary_settings}.

\textbf{Train on ImageNet Only.} The results under ``train on ImageNet only'' settings are reported in Table~\ref{Table:imagenet_only}. As shown in the table, MOKD outperforms other baselines on 10 out of 13 datasets and ranks 1.4 on average. Compared with URL, which MOKD is based on, MOKD outperforms on almost all domains with an average improvement of $1.4\%$. In particular, we find that MOKD performs better on unseen domains (all datasets except ImageNet in this case) compared with the performance on seen domains. MOKD roughly achieves about $1.5\%$ improvements on average on unseen domains. Due to large gaps between seen and unseen domains, performing classification on unseen domains is more challenging. Such a phenomenon further reveals that MOKD can generalize well on previously unseen domains.

\textbf{Train on All Datasets.} The results under the ``train on all datasets'' settings are reported in Table~\ref{Table:all_datasets}. As shown in the table, MOKD achieves the best performance on average and ranks 2.2 among all methods. Compared with URL, which MOKD is based on, MOKD achieves better generalization performance on 10 out of 13 datasets. In addition, consistent with that under ``train on ImageNet only'' settings, MOKD performs better on unseen domains as well and achieves nearly $1\%$ improvements on average compared with URL. 

However, although MOKD achieves better results under the ``train on all datasets'' settings, we still notice that there exist some failure cases. For example, MOKD fails to outperform URT on Quick Draw. On the one side, URT contains an attention head that is more powerful in capturing fine-grained features. On the other side, Quick Draw is a dataset composed of doodling created by different individuals, and thus sometimes it is hard to recognize the common features between data within the same class. Besides, we also notice that MOKD also fails on Fungi. According to the curve (Fig.~\ref{Fig:lcurve_fungi}), such a failure may be derived from overfitting.

\textbf{MOKD v.s. SSL-HSIC.} Due to the similarity between MOKD and SSL-HSIC, we further evaluate SSL-HSIC on CFC tasks under both task settings to compare the two approaches empirically. To be fair, we estimate HSIC measures in SSL-HSIC in the same way as MOKD and use the same bandwidth. According to the results reported in Table~\ref{Table:imagenet_only} and \ref{Table:all_datasets}, MOKD outperforms SSL-HSIC under both settings. The complete analysis is available in Appendix~\ref{Appendix:difference_sslhsic_mokd}.

We can observe that the main difference between the two objectives is the ${\rm HSIC}(Z, Z)$ term. Specifically, in SSL-HSIC loss, the term is ${\rm HSIC}(Z, Z)$ is modified to $\sqrt{{\rm HSIC}(Z, Z)}$ for achieving better performance in practice.  However, according to the original theoretical results of SSL-HSIC, the term should be ${\rm HSIC}(Z, Z)$. In this paper, since we propose to maximize the test power through ${\rm HSIC}(\cdot, \cdot)$, the modification may result in a mismatch of test power.

\vspace{-0.6em}
\subsection{Experimental Analysis}
\textbf{Effect of Kernel Type.} HSIC is a valid statistical measure with characteristic kernels (e.g., Gaussian kernel).
In order to study the effect of kernel type, we further run MOKD with inverse multiquadric kernel (IMQ) under ``train on all datasets'' settings with the same random seeds. Since linear kernels, such as cosine similarity, are not characteristic kernels, we do not consider them in this study. The results are reported in Table~\ref{Table:kernel_type}. According to the table, it is easy to observe that MOKD with Gaussian kernel generally achieves better generalization performance than MOKD with IMQ.

\textbf{Running Time.} To discuss the efficiency of MOKD, we assume that the number of data in the given support set is $m$, the length of the bandwidth list is $k$ and the number of adaptation steps is $s$. Then, in each adaptation episode, the time complexity can be roughly expressed as $(4k+2s)\mathcal{O}(m^2)$ according to Algorithm~\ref{Alg:MOKD}. 
To further quantitatively evaluate the efficiency of MOKD, empirical results are reported in Fig.~\ref{Fig:running_time} and Table~\ref{Table:analyses_on_runtime}. We run the experiment on the same NVIDIA RTX 3090 GPU with the same seeds for fairness. 

According to the results, we find that the time that MOKD consumes for each adaptation is acceptable. In some cases, such as datasets like Omniglot and Fungi, MOKD is even more efficient compared with URL. The main reason for such a phenomenon is that MOKD explicitly leverages the label information via label kernel instead of prototypes. In this way, MOKD does not have to calculate the prototypes repeatedly during adaptation. Thus, although MOKD is a bi-level optimization algorithm that includes HSIC and variance estimations, the algorithm is still efficient in total. In addition, we also notice that MOKD fails to consume less time in some cases, such as ImageNet, Aircraft, and CU\_Birds. The main reason here is that the estimations of HSIC measures and variances depend on the size of the support data. Since the size of support data varies among tasks during adaptation, the time consumption also changes.

\textbf{Ablation Study: Test Power.}
In order to highlight the importance of test power maximization proposed in MOKD, we perform an ablation study on it. 
Fig.~\ref{Fig:ablation_tpm} shows the performance gaps between MOKD with and without test power maximization (TPM) on Meta-dataset benchmark. As shown in the figure, MOKD with test power maximization performs better. 
Besides, MOKD with TPM performs better on unseen domains and complicated datasets like Fungi.

{\textbf{Analysis of $\gamma$.}} 
In order to figure out how datasets in Meta-Dataset react to $\gamma$, we propose to run MOKD on all datasets of Meta-Dataset with different $\gamma$ values under the ``train on all datasets'' settings. According to the results, we can observe that 
complicated datasets, such as ImageNet (Fig.~\ref{Fig:analysis_imagenet_gamma}), prefer large $\gamma$ while simple datasets, such as MNIST (Fig. \ref{Fig:analysis_mnist_gamma}), prefer small $\gamma$. A potential reason for such a phenomenon is the semantic information contained in the images of different datasets. For complicated datasets, due to the huge amounts of semantic information, large $\gamma$ is required to simultaneously learn discriminative features and penalize high-variance representations. 
In contrast, since simple datasets, such as MNIST, own evident and definite semantic areas, small $\gamma$ is enough to focus on the correct semantic areas and achieve better generalization performance.

In addition, an interesting case is that it is equivalent to performing an ablation study on the ${\rm HSIC}(Z, Z)$ term when $\gamma$ is set to zero. According to Fig. \ref{Fig:ablation_gamma}, when ${\rm HSIC}(Z, Z)$ is removed, the performance of MOKD drops significantly. According to Fig. \ref{Fig:test_curves} and \ref{Fig:ablearn_curves}, it is easy to find that the reason for the performance drop is overfitting. Thus, these results demonstrate that ${\rm HSIC}(Z, Z)$ facilitates alleviating overfitting and improving the generalization performance.

\begin{figure}
    \centering
    \setlength{\abovecaptionskip}{0cm}
    \centering
    \includegraphics[width=0.9\linewidth]{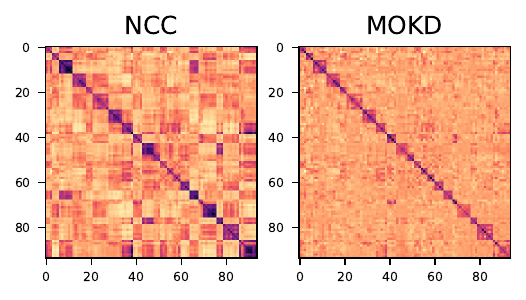}
    \vspace{-1.5em}
    \caption{\textbf{Heatmap visualization of representation similarity on Omniglot.}. The results indicate that MOKD does help learn more discriminative sample clusters than those learned with NCC loss.}
    \vspace{-1.5em}
    \label{fig:visualization_similarity}
\end{figure}
\textbf{Visualization Results.}
Fig.~\ref{fig:visualization_similarity} visualizes the heatmap of the similarity matrices of support data representations respectively learned with NCC-based loss and our proposed MOKD on Omniglot dataset. According to the figure, MOKD learns more definite and clear data clusters compared with those learned with NCC-based loss (URL), which demonstrates that MOKD facilitates capturing the cluster structures of the given support set and learning better class-specific features. Meanwhile, the visualization results on other datasets, such as Fig.~\ref{Fig:intro_imagenet_mokd} and Fig.~\ref{Fig:visual_mnist_sim}, further reveal that MOKD is able to alleviate the undesirable high similarities between two samples from different classes.

\vspace{-0.5em}
\section{Conclusion}
\vspace{-0.2em}
In this paper, we propose an efficient bi-level framework, maximizing optimized kernel dependence, to perform classification on cross-domain few-shot tasks. Specifically, MOKD first maximizes the test power of kernel HSIC to maximize its sensitivity to dependence and then optimizes the optimized kernel HSIC to learn class-specific representations. Extensive experimental results on the Meta-Dataset benchmark demonstrate that MOKD can simultaneously achieve good generalization performance and mitigate the undesirable high similarities for better data clusters.

\vspace{-0.8em}
\section*{Impact Statement}
In this paper, we provide a new interpretation of NCC-based loss from the perspective of dependence measure and propose to solve cross-domain few-shot classification tasks with the dependence measures where the test power is maximized so that the dependence between samples can be more accurately detected. Two advantages are worth noticing in the MOKD framework. For one thing, such a framework is efficient though it is a bi-level optimization problem. For another thing, by defining an appropriate label kernel, more information, such as cluster structure, can be utilized for model adaptation. Thus, such a framework can be generalized or specified to any other existing pipelines, where cross-entropy loss with softmax function is adopted, to learn more class-specific features where the similarity among samples within the same class is maximized while the similarity between samples from different classes is minimized. There are many potential societal consequences of our work, none of which we feel must be specifically highlighted here.

\section*{Acknowledgement}
HDT and BH were supported by the NSFC General Program No.~62376235, Guangdong Basic and Applied Basic Research Foundation Nos.~2022A1515011652 and 2024A1515012399, HKBU Faculty Niche Research Areas No.~RC-FNRA-IG/22-23/SCI/04, and HKBU CSD Departmental Incentive Grant. FL was supported by the Australian Research Council with grant numbers DP230101540 and DE240101089, and the NSF\&CSIRO Responsible AI program with grant number 2303037. BD was supported by the National Natural Science Foundation of China under Grants 62225113, the National Key Research and Development Program of China 2023YFC2705700. YMC was supported in part by NSFC/Research Grants Council (RGC) Joint Research Scheme under Grant: N\_HKBU214/21; in part by the General Research Fund of RGC under Grants: 12201321, 12202622, and 12201323; and in part by RGC Senior Research Fellow Scheme under Grant: SRFS2324-2S02. TLL was partially supported by the following Australian Research Council projects: FT220100318, DP220102121, LP220100527, LP220200949, and IC190100031. 

\bibliography{MOKD}
\bibliographystyle{icml2024}

\newpage
\appendix
\onecolumn

\renewcommand*\contentsname{Appendix}
\tableofcontents

\addtocontents{toc}{\protect\setcounter{tocdepth}{3}}

\clearpage

\section{More Related Work}\label{Appendix:more_related_works}
\textbf{Cross-domain Few-shot Classification.} 
Cross-domain few-shot classification aims to perform few-shot classification on tasks that are sampled from not only previously unseen data sets with the same distribution of seen domains (e.g. test set) but also previously unseen domains with different distributions. Compared with conventional few-shot classification~\citep{MatchingNet,ravi2016optimization,maml,prototypical,reptile,tian2020rethinking}, CFC is much more challenging mainly due to two aspects. First of all, the distribution gaps between source and target domains are sometimes quite large. For example, the distribution of ImageNet (i.e. nature scenes) is quite different from that of Omniglot (i.e. handwritings). Thus, since the feature patterns of ImageNet has never been observed by the model, the model may fail to achieve good generalization performance if it was pre-trained on Omniglot dataset~\citep{omniglot} but evaluated on ImageNet~\citep{imagenet}. On the other side, the task setting of CFC is more difficult~\cite{yang2021learning}. For example, under the vary-way vary-shot setting, the number of classes and the number of shots for each class are randomly determined at the beginning of each episode.

Generally, although many works have been done in this field, existing methods are built in the way of typical meta-learning frameworks~\cite{maml,prototypical} and can be mainly divided into two genres according to the ways of training feature encoders. One of the genres trains the feature encoder and the classifier from scratch. For example, \citet{metadatasets} proposes Proto-MAML which combines Prototypical Nets~\citep{prototypical} and MAML~\citep{maml} by treating the prototypes obtained from the feature encoder as the parameters of the linear classification layer to perform cross-domain few-shot classification. Based on Proto-MAML, ALFA+fo-Proto-MAML~\citep{alpha} further proposes to adaptively generate task-specific hyperparameters, such as learning rate and weight decay, from a small model for each task. Besides, CNAPS~\citep{CNAPS} proposes to leverage the FiLM~\citep{FiLM} module to adapt the parameters of both the feature encoder and classifier to new tasks. Further, SimpleCNAPS proposes to substitute the parametric classifier used in CNAPS with a class-covariance-based distance metric to improve the efficiency and generalization performance with fewer parameters. Currently, as far as we know, the state-of-the-art method in this genre is CrossTransformer~\citep{crosstransformer} which proposes to learn spatial correspondence with a self-attention module from tasks sampled from ImageNet and generalize the prior knowledge to other unseen domains. 

The other genre in cross-domain few-shot classification aims to transfer the prior knowledge of a pre-trained backbone to tasks sampled from previously unseen domains by further training or fine-tuning a module on top of the frozen backbone. An intuition behind these methods is that a pre-trained backbone is able to extract good representations from the given new tasks sampled from previously unseen domains since it has been well-trained on some large datasets, such as ImageNet~\citep{imagenet}). Statistically, such an intuition can be further explained by an assumption that the distribution learned by a pre-trained model is similar to the distribution where downstream tasks are sampled. To some extent, this assumption can be demonstrated by \citet{raghu2019rapid} that the success of meta-learning ought to be attributed to feature reuse. Thus, the semantic feature space embedded in the pre-trained model can partially or fully cover the feature space of target domains and recognize as many feature patterns as possible. Specifically, SUR~\citep{SUR} proposes to learn a weight vector to select features from the output of several pre-trained backbones and combine these features in a linear way with the corresponding weights. Later, URT~\citep{urt} proposes to train a universal representation Transformer~\citep{attention} to select features from the embeddings extracted from 8 pre-trained domain-specific backbones. Besides, FLUTE~\citep{FLUTE} proposes to treat the convolutional layers of a model as universal templates which will be frozen during the test phase while training specific batch normalization layers for each of the 8 seen domain datasets. Then, a `Blender' model is trained to learn to combine the statistical information from all specific BN layers. During meta-test, a new specific BN is generated by feeding the support data in the Blender model. However, since the forward passes of several backbones consume too much time during the test phase, \citet{url} proposes URL to learn a universal multi-domain backbone from the 8 pre-trained backbones via knowledge distillation. During the test phase, URL fine-tunes a simple linear transformation head on top of the distilled backbone for adaptation of the new tasks. As a further version of URL, TSA~\citep{tsa}, which is the state-of-the-art method in this genre, plugs additional learning modules into the backbone so that more task-specific features can be learned.

\paragraph{Distance Function.} Distance function is an essential component of approaches that are based on the prototypes. Originally, the distance function adopted is Euclidean distance function~\cite{metadatasets,CNAPS}. However, \citet{SimpleCNAPS} points out that the Euclidean distance function, which corresponds to the square Mahalanobis distance, implicitly assumes each cluster is distributed according to a unit norm. Based on this, squared Mahalanobias distance is adopted in SimpleCNAPs~\cite{SimpleCNAPS} to consider cluster covariance when computing distances to the cluster centers. In addition, in most pre-training based pipelines~\cite{urt,url,tsa}, the cosine similarity function is adopted. On the one side, when data sampled are normalized to a unit sphere, the cosine similarity is approximately equivalent to the Euclidean distance function and the relationship can be expressed as $(x-y)^{\top}(x-y)=2(1-\cos(x, y))$. It also has been demonstrated that directly matching uniformly sampled points on the unit hypersphere contributes to learning good representations~\cite{liu2023inducing,wang2020understanding}. In addition, as argued by \citet{url}, the cosine similarity function can be treated as a generalization of the Mahalanobis distance computation via decomposing the inverse of the covariance matrix into a product of a lower triangle matrix and its conjugate transpose.

\textbf{Feature Selection via Dependence Measures} 
Dependence measure has been well studied in statistics, and a series of measures have been proposed in previous works, such as constrained covariance (COCO)~\citep{coco}, kernel canonical correlation~\citep{bach2002kernel} and Hilbert-Schmidt Independence Criterion (HSIC)~\citep{hsic}. Recently, these tools, especially HSIC, have been widely applied in deep learning. Specifically, \citet{hsic_estimation} firstly proposed to select features by maximizing the dependence between the selected features and the labels. \citet{kumagai2022few} follows previous works~\citep{yamada2014high,freidling2021post,koyama2022effective} and propose to select relevant features and remove the redundant features by solving an $\ell_1$-regularized regression problem. Besides, \citet{sslhsic} proposes to replace InfoNCE~\citep{infonce} with SSL-HSIC to directly optimize statistical dependence.

\section{Proof Results}\label{Appendix:Proof_results}
\subsection{A Necessary Lemma}
\begin{lemma}\label{Lemma:bound_of_JE}
    Suppose that $a_i\in[0, 1]$ for $\forall i\in\{1, 2, ..., n\}$, then $\frac{1}{n}\sum_{i=1}^{n}e^{a_i}$ and $e^{\frac{1}{n}\sum_{i=1}^{n}a_i}$ satisfy that
    \[
        e^{\frac{1}{n}\sum_{i=1}^{n}a_i} \ge \frac{1}{n}\sum_{i=1}^{n}e^{a_i} - \left(e+(e-1)\log(e-1)\right).
    \]
\end{lemma}
\vspace{-1.5em}
\begin{proof}
    Since $\exp(\cdot)$ is a convex function, according to Jesen's Inequality, we have $e^{\frac{1}{n}\sum_{i=1}^{n}a_i}\leq \frac{1}{n}\sum_{i=1}^{n}e^{a_i}$. In addition, in the interval $[0, 1]$, $\exp(x)$ can be approximated to a linear function $(e-1)x+1$. Thus, in this way, we have $\frac{1}{n}\sum_{i=1}^{n}e^{a_i}=(e-1)(\frac{1}{n}\sum_{i=1}^{n}a_i)+1\ge e^{\frac{1}{n}\sum_{i=1}^{n}a_i}$.

    Then, let $b=\sum_{i=1}^{n}a_i$, we have a function $f(b)=(e-1)b+1-e^x$. By computing $\frac{df}{db}=0$, we can obtain that $f(b)$ achieves its maximum $e+(e-1)\log(e-1)$ at $b=\log(e-1)$. 
    Thus, we have $e^{\frac{1}{n}\sum_{i=1}^{n}a_i} \ge \frac{1}{n}\sum_{i=1}^{n}e^{a_i} - \left(e+(e-1)\log(e-1)\right)$.
\end{proof}
\subsection{Proof of Theorem~\ref{Theorem:lower_bound_protoloss}}\label{Appendix:proof_lowerbound_protoloss}
\begin{proof}\label{Proof:lower_bound_protoloss} 
We first unfold the NCC-based loss as follows:

\begin{equation}\label{Eq:single_loss_unfold}
\begin{aligned}
-\frac{1}{|\mathcal{D}_{\mathcal{T}}|}\sum_{i=1}^{|\mathcal{D}_{\mathcal{T}}|}\log\left(p(y=y_i|\boldsymbol{z}_i, \theta)\right) &= -\frac{1}{|\mathcal{D}_{\mathcal{T}}|}\sum_{i=1}^{|\mathcal{D}_{\mathcal{T}}|}\log\frac{\exp(k(\boldsymbol{z}_i, \boldsymbol{c}))}{\sum_{j=1}^{N_C}\exp(k(\boldsymbol{z}_i, \boldsymbol{c}_j))}\\
&=-\frac{1}{|\mathcal{D}_{\mathcal{T}}|}\sum_{i=1}^{|\mathcal{D}_{\mathcal{T}}|}k(\boldsymbol{z}_i, \boldsymbol{c}) + \frac{1}{|\mathcal{D}_{\mathcal{T}}|}\sum_{i=1}^{|\mathcal{D}_{\mathcal{T}}|}\log\sum_{j=1}^{N_C}\exp(k(\boldsymbol{z}_i, \boldsymbol{c}_j))\\
&=-\frac{1}{|\mathcal{D}_{\mathcal{T}}|}\sum_{i=1}^{|\mathcal{D}_{\mathcal{T}}|}\left(\boldsymbol{z}_i^{\top}\frac{1}{|\mathcal{C}|}\sum_{\boldsymbol{z}^{+}\in\mathcal{C}}\boldsymbol{z}^{+}\right) + \frac{1}{|\mathcal{D}_{\mathcal{T}}|}\sum_{i=1}^{|\mathcal{D}_{\mathcal{T}}|}\log\sum_{j=1}^{N_C}\exp\left(\boldsymbol{z}_i^{\top}\frac{1}{|\mathcal{C}_j|}\sum_{\boldsymbol{z}^{'}\in\mathcal{C}_j}\boldsymbol{z}^{'}\right)\\
&\overset{(1)}{\ge}-\frac{1}{|\mathcal{D}_{\mathcal{T}}|}\sum_{i=1}^{|\mathcal{D}_{\mathcal{T}}|}\frac{1}{|\mathcal{C}|}\sum_{\boldsymbol{z}^{+}\in\mathcal{C}}\boldsymbol{z}_i^{\top}\boldsymbol{z}^{+} + \frac{1}{|\mathcal{D}_{\mathcal{T}}|}\sum_{i=1}^{|\mathcal{D}_{\mathcal{T}}|}\log\sum_{j=1}^{N_C}\frac{1}{|\mathcal{C}_j|}\sum_{\boldsymbol{z}^{'}\in\mathcal{C}_j}\exp\left(\boldsymbol{z}_i^{\top}\boldsymbol{z}^{'}\right)+\log\alpha_{e}\\
&=-\frac{1}{|\mathcal{D}_{\mathcal{T}}|}\sum_{i=1}^{|\mathcal{D}_{\mathcal{T}}|}\frac{1}{|\mathcal{C}|}\sum_{\boldsymbol{z}^{+}\in\mathcal{C}}\boldsymbol{z}_i^{\top}\boldsymbol{z}^{+} + \frac{1}{|\mathcal{D}_{\mathcal{T}}|}\sum_{i=1}^{|\mathcal{D}_{\mathcal{T}}|}\log\sum_{j=1}^{N_C}\sum_{\boldsymbol{z}^{'}\in\mathcal{C}_j}\frac{1}{N_C|\mathcal{C}_j|}\exp\left(\boldsymbol{z}_i^{\top}\boldsymbol{z}^{'}\right)+\log \alpha_eN_C,
\end{aligned}
\end{equation}
where $(1)$ follows Lemma~\ref{Lemma:bound_of_JE} and $\alpha_{e}\in\mathbb{R^{+}}$ is a real constant that satisfies that $\alpha_e\left(\sum_{j=1}^{N_C}\frac{1}{|\mathcal{C}_j|}\sum_{\boldsymbol{z}^{'}\in\mathcal{C}_j}\exp\left(\boldsymbol{z}_i^{\top}\boldsymbol{z}^{'}\right)\right)\leq \sum_{j=1}^{N_C}\left(\frac{1}{|\mathcal{C}_j|}\sum_{\boldsymbol{z}^{'}\in\mathcal{C}_j}\exp\left(\boldsymbol{z}_i^{\top}\boldsymbol{z}^{'}\right)-\left(e+(e-1)\log(e-1)\right)\right)$, $\boldsymbol{c}$ denote the class centroid representations $\boldsymbol{z}_i$ belonging to, $\boldsymbol{c}_j$ denotes the $j$-th class centroid, $\mathcal{C}_j$ denotes the $j$-th class sets. 

Then, we perform Taylor expansion on the second term of Eq.~\eqref{Eq:single_loss_unfold} around $\mu=\sum_{\boldsymbol{z}^{'}\in\mathcal{Z}}\frac{1}{|\mathcal{D}_{\mathcal{T}}|}k\left(\boldsymbol{z}_i \boldsymbol{z}^{'}\right)$. Based on Assumption~\ref{Assumption:proximity}, we can approximately assume that $k(\boldsymbol{z}, \boldsymbol{z}^{'})\approx\mu$. Then, we have:
\[\label{Eq:taylor_expansion_on_2_term}
\begin{aligned}
    \frac{1}{|\mathcal{D}_{\mathcal{T}}|}\sum_{i=1}^{|\mathcal{D}_{\mathcal{T}}|}\log\sum_{j=1}^{N_C}\sum_{\boldsymbol{z}^{'}\in\mathcal{C}_j}\frac{1}{N_C|\mathcal{C}_j|}\exp\left(\boldsymbol{z}_i^{\top}\boldsymbol{z}^{'}\right) &= \frac{1}{|\mathcal{D}_{\mathcal{T}}|}\sum_{i=1}^{|\mathcal{D}_{\mathcal{T}}|}\log\sum_{j=1}^{N_C}\sum_{\boldsymbol{z}^{'}\in\mathcal{C}_j}\frac{e^{\mu}\left[1+\frac{\left(k(\boldsymbol{z}, \boldsymbol{z}^{'})-\mu\right)^2}{2}\right]}{N_C|\mathcal{C}_j|}\\
    &\overset{(2)}{\ge}\frac{1}{|\mathcal{D}_{\mathcal{T}}|}\sum_{i=1}^{|\mathcal{D}_{\mathcal{T}}|}\sum_{\boldsymbol{z}^{'}\in\mathcal{Z}}\frac{1}{|\mathcal{D}_{\mathcal{T}}|}k\left(\boldsymbol{z}_i, \boldsymbol{z}^{'}\right) + \mathcal{O}\left(k(\boldsymbol{z}, \boldsymbol{z}^{'})\right)\\
\end{aligned}
\]
where $(2)$ follows Jesen's Inequality for the convex function $-\log(\cdot)$ the expansion of $\log(1+x)$ around $x=0$, $\mathcal{O}\left(k(\boldsymbol{z}, \boldsymbol{z}^{'})\right)=\frac{1}{|\mathcal{D}_{\mathcal{T}}|}\sum_{i=1}^{|\mathcal{D}_{\mathcal{T}}|}\sum_{j=1}^{N_C}\sum_{\boldsymbol{z}^{'}\in\mathcal{C}_j}\frac{\left[k(\boldsymbol{z}_i, \boldsymbol{z}^{'})-\sum_{\boldsymbol{z}^{'}\in\mathcal{Z}}\frac{1}{|\mathcal{D}_{\mathcal{T}}|}k(\boldsymbol{z}_i, \boldsymbol{z}^{'})\right]^2}{2N_C|\mathcal{C}_j|}$ denotes a high-order moment term regarding $k(\boldsymbol{z}, \boldsymbol{z}^{'})$. In this way, we conclude that Problem~\ref{Eq:prototypical_loss} owns a lower bound:

\[
\begin{aligned}
    \mathcal{L}(\theta)&\ge-\frac{1}{|\mathcal{D}_{\mathcal{T}}|}\sum_{i=1}^{|\mathcal{D}_{\mathcal{T}}|}\frac{1}{|\mathcal{C}|}\sum_{\boldsymbol{z}^{+}\in\mathcal{C}}k(\boldsymbol{z}_i, \boldsymbol{z}^{+})
    +\frac{1}{|\mathcal{D}_{\mathcal{T}}|}\sum_{i=1}^{|\mathcal{D}_{\mathcal{T}}|}\sum_{\boldsymbol{z}^{'}\in\mathcal{Z}}\frac{k(\boldsymbol{z}_i, \boldsymbol{z}^{'})}{|\mathcal{D}_{\mathcal{T}}|} + \mathcal{O}\left(k(\boldsymbol{z}, \boldsymbol{z}^{'})\right)+\log\alpha_eN_C,\\
\end{aligned}
\]
where $\mathcal{O}\left(k(\boldsymbol{z}, \boldsymbol{z}^{'})\right)=\frac{1}{|\mathcal{D}_{\mathcal{T}}|}\sum_{i=1}^{|\mathcal{D}_{\mathcal{T}}|}\sum_{j=1}^{N_C}\sum_{\boldsymbol{z}^{'}\in\mathcal{C}_j}\frac{\left[k(\boldsymbol{z}_i, \boldsymbol{z}^{'})-\sum_{\boldsymbol{z}^{'}\in\mathcal{Z}}\frac{1}{|\mathcal{D}_{\mathcal{T}}|}k(\boldsymbol{z}_i, \boldsymbol{z}^{'})\right]^2}{2N_C|\mathcal{C}_j|}$ denotes a high-order moment term regarding $k(\boldsymbol{z}, \boldsymbol{z}^{'})$.
\end{proof}

\subsection{Proof of Theorem~\ref{Theorem:HSIC_Z_Y}}
\begin{proof}
Following~\cite{sslhsic}, we compute HSIC by directly calculating the three terms in Eq.~(\ref{Eq:empirical_hsic}) under cross-domain few-shot classification setting respectively. 
Given a set of support representations $\mathcal{Z}=\{\boldsymbol{z}_i\}_{i=1}^{|\mathcal{D}_{\mathcal{T}}|}=\{h_{\theta}\circ f_{\phi^*}(\boldsymbol{x}_i)\}_{i=1}^{|\mathcal{\mathcal{D}_{\mathcal{T}}}|}$, each data sample in $\mathcal{Z}$ is randomly sampled with the probability $\frac{1}{|\mathcal{D}_{\mathcal{T}}|}$, where $N_C$ denotes the number of classes. Then, for the first term, we can obtain
\begin{align*}
    \mathbb{E}\left[k(Z, Z^{'})l(Y, Y^{'})\right]&=\mathbb{E}_{Z, Z^{'}, Y, Y^{'}}\left[k(Z, Z^{'})l(Y, Y^{'})\right] \\
    &=\Delta l\mathbb{E}_{\boldsymbol{Z}, \boldsymbol{Z}^{'}, Y, Y^{'}}\left[k(Z, Z^{'})\mathbb{I}(Y, Y^{'})\right]+l_0\mathbb{E}_{\boldsymbol{Z}, \boldsymbol{Z}^{'}}\left[k(\boldsymbol{Z}, \boldsymbol{Z}^{'})\right]\\
    &=\Delta l\sum_{i=1}^{|\mathcal{D}_{\mathcal{T}}|}\sum_{j=1}^{|\mathcal{D}_{\mathcal{T}}|}\mathbb{E}_{Z|y_i, Z^{'}|y^{'}_j}\left[\frac{1}{|\mathcal{D}_{\mathcal{T}}|}\cdot\frac{1}{|\mathcal{D}_{\mathcal{T}}|}\cdot k(Z, Z^{'})\mathbb{I}(y_i=y^{'}_j)\right]+l_0\mathbb{E}_{Z, Z^{'}}\left[k(Z, Z^{'})\right]\\
    &=\Delta l\sum_{i=1}^{|\mathcal{D}_{\mathcal{T}}|}\mathbb{E}_{Z|y_i, Z^{'}|y_i}\left[\frac{1}{|\mathcal{D}_{\mathcal{T}}|}\cdot\frac{|\mathcal{C}_{y_i}|}{|\mathcal{D}_{\mathcal{T}}|} k(Z, Z^{'})\right]+l_0\mathbb{E}_{Z, Z^{'}}\left[k(Z, Z^{'})\right]\\
    &= \Delta l\sum_{i=1}^{|\mathcal{D}_{\mathcal{T}}|}\frac{|\mathcal{C}_{y_i}|}{|\mathcal{D}_{\mathcal{T}}|^2}\mathbb{E}_{Z|y_i, Z^{'}|y_i}\left[k(Z, Z^{'})\right]+l_0\mathbb{E}_{Z, Z^{'}}\left[k(Z, Z^{'})\right]
\end{align*}
\begin{align*}
    &~~~~~~~~~~~= \frac{\Delta l}{|\mathcal{D}_{\mathcal{T}}|}\sum_{i=1}^{|\mathcal{D}_{\mathcal{T}}|}\frac{|\mathcal{C}_{y_i}|}{|\mathcal{D}_{\mathcal{T}}|}\sum_{m=1}^{|\mathcal{C}_{y_i}|}\sum_{n=1}^{|\mathcal{C}_{y_i}|}\frac{1}{|\mathcal{C}_{y_i}|^2}\left[k(\boldsymbol{z}_m, \boldsymbol{z}^{'}_n)\right]+l_0\mathbb{E}_{Z, Z^{'}}\left[k(Z, Z^{'})\right]\\
    &~~~~~~~~~~~= \frac{\Delta l}{|\mathcal{D}_{\mathcal{T}}|}\sum_{c=1}^{N_C}\frac{|\mathcal{C}_c|^2}{|\mathcal{D}_{\mathcal{T}}|}\sum_{m=1}^{|\mathcal{C}_c|}\sum_{n=1}^{|\mathcal{C}_c|}\frac{1}{|\mathcal{C}_c|^2}k(\boldsymbol{z}_m, \boldsymbol{z}^{'}_n)+l_0\mathbb{E}_{Z, Z^{'}}\mathbb{E}\left[k(Z, Z^{'})\right]\\
    &~~~~~~~~~~~= \frac{\Delta l}{|\mathcal{D}_{\mathcal{T}}|^2}\sum_{i=1}^{|\mathcal{D}_{\mathcal{T}}|}\sum_{\boldsymbol{z}^{+}\in\mathcal{C}}k(\boldsymbol{z}_i, \boldsymbol{z}^{+}) + l_0\mathbb{E}_{Z, Z^{'}}\left[k(Z, Z^{'})\right]
\end{align*}
where $\mathcal{C}$ denotes the class set that $y_i$ belongs to.

Then, due to the independence between $\boldsymbol{Z}^{'}$ and $Y^{''}$, the second term can be calculated as:
\[
\begin{aligned}
    \mathbb{E}\left[k(Z, Z^{'})l(Y, Y^{''})\right]&=\mathbb{E}_{ZY}\left[\mathbb{E}_{Z^{'}}[k(Z, Z^{'})]\mathbb{E}_{Y^{''}}[l(Y, Y^{''})]\right]\\
    &=\sum_{i=1}^{|\mathcal{D}_{\mathcal{T}}|}\mathbb{E}_{Z|y_i}\left[\mathbb{E}_{Z^{'}}k(Z, Z^{'})\left(\frac{\Delta l|\mathcal{C}_{y_i}|}{|\mathcal{D}_{\mathcal{T}}|^2}+l_0\right)\right]\\
    &=\sum_{i=1}^{|\mathcal{D}_{\mathcal{T}}|}\frac{\Delta l|\mathcal{C}_{y_i}|}{|\mathcal{D}_{\mathcal{T}}|^2}\sum_{m=1}^{|\mathcal{C}_{y_i}|}\frac{1}{|\mathcal{C}_{y_i}|}\sum_{n=1}^{|\mathcal{D}_{\mathcal{T}}|}\frac{1}{|\mathcal{D}_{\mathcal{T}}|}\left[k(\boldsymbol{z}_m, \boldsymbol{z}^{'}_n)\right]+l_0\mathbb{E}_{Z, Z^{'}}\left[k(Z, Z^{'})\right]\\
    &= \frac{\Delta l}{|\mathcal{D}_{\mathcal{T}}|}\sum_{c=1}^{N_C}\frac{|\mathcal{C}_c|^2}{|\mathcal{D}_{\mathcal{T}}|}\sum_{m=1}^{|\mathcal{C}_c|}\frac{1}{|\mathcal{C}_c|}\sum_{n=1}^{|\mathcal{D}_{\mathcal{T}}|}\frac{1}{|\mathcal{D}_{\mathcal{T}}|}k(\boldsymbol{z}_m, \boldsymbol{z}^{'}_{n})+l_0\mathbb{E}_{Z, Z^{'}}\left[k(Z, Z^{'})\right]\\
    &= \frac{\Delta l}{|\mathcal{D}_{\mathcal{T}}|^2}\sum_{c=1}^{N_C}\sum_{m=1}^{|\mathcal{C}_c|}\sum_{n=1}^{|\mathcal{D}_{\mathcal{T}}|}\frac{|\mathcal{C}_c|}{|\mathcal{D}_{\mathcal{T}}|}k(\boldsymbol{z}_m, \boldsymbol{z}^{'}_{n})+l_0\mathbb{E}_{Z, Z^{'}}\left[k(Z, Z^{'})\right]\\
    &= \frac{\Delta l}{|\mathcal{D}_{\mathcal{T}}|^2}\sum_{i=1}^{|\mathcal{D}_{\mathcal{T}}|}\sum_{j=1}^{N_C}\sum_{\boldsymbol{z}^{'}\in\mathcal{C}_j}\frac{|\mathcal{C}_j|}{|\mathcal{D}_{\mathcal{T}}|}k(\boldsymbol{z}_i, \boldsymbol{z}^{'})+l_0\mathbb{E}_{Z, Z^{'}}\left[k(Z, Z^{'})\right]\\
    &= \frac{\Delta l}{|\mathcal{D}_{\mathcal{T}}|^2}\sum_{i=1}^{|\mathcal{D}_{\mathcal{T}}|}\sum_{\boldsymbol{z}^{'}\in\mathcal{Z}}\frac{1}{|\mathcal{D}_{\mathcal{T}}|}k(\boldsymbol{z}_i, \boldsymbol{z}^{'})+ \frac{\Delta l}{|\mathcal{D}_{\mathcal{T}}|^2}\sum_{i=1}^{|\mathcal{D}_{\mathcal{T}}|}\sum_{j=1}^{N_C}\sum_{\boldsymbol{z}^{'}\in\mathcal{C}_j}\frac{|\mathcal{C}_c|-1}{|\mathcal{D}_{\mathcal{T}}|}k(\boldsymbol{z}_i, \boldsymbol{z}^{'}) +l_0\mathbb{E}_{Z, Z^{'}}\left[k(Z, Z^{'})\right]
\end{aligned}
\]

For the third part, we can obtain:
\begin{align*}
    \mathbb{E}\left[k(Z, Z^{'})\right]\mathbb{E}\left[l(Y, Y^{'})\right]&=\mathbb{E}_{Z, Z^{'}}\left[k(Z, Z^{'})\right]\mathbb{E}_{Y, Y^{'}}\left[l(Y, Y^{'})\right]\\
    &=\mathbb{E}_{Z, Z^{'}}\left[k(Z, Z^{'})\right]\left(\sum_{i=1}^{|\mathcal{D}_{\mathcal{T}}|}\sum_{j=1}^{|\mathcal{D}_{\mathcal{T}}|}\frac{1}{|\mathcal{D}_{\mathcal{T}}|}\cdot\frac{1}{|\mathcal{D}_{\mathcal{T}}|}\cdot\Delta l\mathbb{I}(y_i=y^{'}_j) + l_0\right)\\
    &=\sum_{i=1}^{|\mathcal{D}_{\mathcal{T}}|}\frac{1}{|\mathcal{D}_{\mathcal{T}}|}\cdot\frac{\Delta l|\mathcal{C}_{y_i}|}{|\mathcal{D}_{\mathcal{T}}|}\mathbb{E}_{Z, Z^{'}}\left[k(Z, Z^{'})\right] + l_0\mathbb{E}_{Z, Z^{'}}\left[k(Z, Z^{'})\right]\\
    &=\sum_{i=1}^{|\mathcal{D}_{\mathcal{T}}|}\frac{1}{|\mathcal{D}_{\mathcal{T}}|}\cdot\frac{\Delta l|\mathcal{C}_{y_i}|}{|\mathcal{D}_{\mathcal{T}}|}\sum_{m=1}^{|\mathcal{D}_{\mathcal{T}}|}\frac{1}{|\mathcal{D}_{\mathcal{T}}|}\sum_{n=1}^{|\mathcal{D}_{\mathcal{T}}|}\frac{1}{|\mathcal{D}_{\mathcal{T}}|}k(\boldsymbol{z}_m, \boldsymbol{z}^{'}_n) + l_0\mathbb{E}_{Z, Z^{'}}\left[k(Z, Z^{'})\right]\\
    &=\frac{\Delta l}{|\mathcal{D}_{\mathcal{T}}|^2}\sum_{i=1}^{|\mathcal{D}_{\mathcal{T}}|}\frac{|\mathcal{C}_j|}{|\mathcal{D}_{\mathcal{T}}|^2}\sum_{m=1}^{|\mathcal{D}_{\mathcal{T}}|}\sum_{n=1}^{|\mathcal{D}_{\mathcal{T}}|}k(\boldsymbol{z}_m, \boldsymbol{z}^{'}_n) + l_0\mathbb{E}_{Z, Z^{'}}\left[k(Z, Z^{'})\right]
\end{align*}
\begin{align*}
    &~~~~~~~~~~~~~~~=\frac{\Delta l}{|\mathcal{D}_{\mathcal{T}}|^2}\frac{\sum_{c=1}^{N_C}|\mathcal{C}_c|}{|\mathcal{D}_{\mathcal{T}}|^2}\sum_{m=1}^{|\mathcal{D}_{\mathcal{T}}|}\sum_{n=1}^{|\mathcal{D}_{\mathcal{T}}|}k(\boldsymbol{z}_m, \boldsymbol{z}^{'}_n) + l_0\mathbb{E}_{Z, Z^{'}}\left[k(Z, Z^{'})\right]\\
    &~~~~~~~~~~~~~~~=\frac{\Delta l}{|\mathcal{D}_{\mathcal{T}}|^2}\sum_{m=1}^{|\mathcal{D}_{\mathcal{T}}|}\sum_{n=1}^{|\mathcal{D}_{\mathcal{T}}|}\frac{1}{|\mathcal{D}_{\mathcal{T}}|}k(\boldsymbol{z}_m, \boldsymbol{z}^{'}_n) + l_0\mathbb{E}_{Z, Z^{'}}\left[k(Z, Z^{'})\right]\\
    &~~~~~~~~~~~~~~~=\frac{\Delta l}{|\mathcal{D}_{\mathcal{T}}|^2}\sum_{i=1}^{|\mathcal{D}_{\mathcal{T}}|}\sum_{\boldsymbol{z}^{'}\in\mathcal{Z}}\frac{1}{|\mathcal{D}_{\mathcal{T}}|}k(\boldsymbol{z}_i, \boldsymbol{z}^{'}) + l_0\mathbb{E}_{Z, Z^{'}}\left[k(Z, Z^{'})\right],
\end{align*}

Thus, ${\rm HSIC}(Z, Y)$ can be specifically reformulated as:
\[
\begin{small}
\begin{aligned}
    {\rm HSIC}(Z, Y) &= \mathbb{E}[k(Z, Z^{'})l(Y, Y^{'})]-2\mathbb{E}[k(Z, Z^{'})l(Y, Y^{''})]+\mathbb{E}[k(Z, Z^{'})]\mathbb{E}[l(Y, Y^{'})]\\
    & = \frac{\Delta l}{|\mathcal{D}_{\mathcal{T}}|}\left(\frac{1}{|\mathcal{D}_{\mathcal{T}}|}\sum_{i=1}^{|\mathcal{D}_{\mathcal{T}}|}\sum_{\boldsymbol{z}^{+}\in\mathcal{C}}k(\boldsymbol{z}_i, \boldsymbol{z}^{+}) - \frac{1}{|\mathcal{D}_{\mathcal{T}}|}\sum_{i=1}^{|\mathcal{D}_{\mathcal{T}}|}\sum_{\boldsymbol{z}^{'}\in\mathcal{Z}}\frac{1}{|\mathcal{D}_{\mathcal{T}}|}k(\boldsymbol{z}_i, \boldsymbol{z}^{'})\right) - \frac{2\Delta l}{|\mathcal{D}_{\mathcal{T}}|^2}\sum_{i=1}^{|\mathcal{D}_{\mathcal{T}}|}\sum_{j=1}^{N_C}\sum_{\boldsymbol{z}^{'}\in\mathcal{C}_j}\frac{|\mathcal{C}_c|-1}{|\mathcal{D}_{\mathcal{T}}|}k(\boldsymbol{z}_i, \boldsymbol{z}^{'})\\
    &\ge \lambda\cdot\frac{\Delta l}{|\mathcal{D}_{\mathcal{T}}|}\left(\frac{1}{|\mathcal{D}_{\mathcal{T}}|}\sum_{i=1}^{|\mathcal{D}_{\mathcal{T}}|}\sum_{\boldsymbol{z}^{+}\in\mathcal{C}}k(\boldsymbol{z}_i, \boldsymbol{z}^{+}) - \frac{1}{|\mathcal{D}_{\mathcal{T}}|}\sum_{i=1}^{|\mathcal{D}_{\mathcal{T}}|}\sum_{\boldsymbol{z}^{'}\in\mathcal{Z}}\frac{1}{|\mathcal{D}_{\mathcal{T}}|}k(\boldsymbol{z}_i, \boldsymbol{z}^{'})\right),
\end{aligned}
\end{small}
\]
where $\lambda$ is a scale constant.
\end{proof}

\subsection{Relation between $\mathrm{HSIC}(Z, Z)$ and High-order Moment Term}\label{Appendix:HSICZZ}
We first decompose $\mathrm{HSIC}(Z, Z)$. According to the definition of HSIC and Cauchy-Schwarz Inequality, we can expand $\mathrm{HSIC}(Z, Z)$ under the setting of vary-way vary-shot few-shot classification task as:
\[
\begin{aligned}
    \mathrm{HSIC}(Z, Z) &= \mathbb{E}_{Z, Z^{'}}k(Z, Z^{'})^2 - 2\mathbb{E}_{Z}(\mathbb{E}_{Z^{'}}k(Z, Z^{'}))^2 + (\mathbb{E}_{Z, Z^{'}}k(Z, Z^{'}))^2\\
    &\leq \mathbb{E}_{Z, Z^{'}}k(Z, Z^{'})^2 - \mathbb{E}_{Z}(\mathbb{E}_{Z^{'}}k(Z, Z^{'}))^2\\
    &= \sum_{i=1}^{|\mathcal{D}_{\mathcal{T}}|}\frac{1}{|\mathcal{D}_{\mathcal{T}}|}\sum_{j=1}^{\mathcal{D}_{\mathcal{T}}}\frac{1}{|\mathcal{D}_{\mathcal{T}}|}k(\boldsymbol{z}_i, \boldsymbol{z}^{'}_j)^2-\sum_{i=1}^{|\mathcal{D}_{\mathcal{T}}|}\frac{1}{|\mathcal{D}_{\mathcal{T}}|}\left(\sum_{j=1}^{|\mathcal{D}_{\mathcal{T}}|}\frac{1}{|\mathcal{D}_{\mathcal{T}}|}k(\boldsymbol{z}_i, \boldsymbol{z}^{'}_j)\right)^2\\
    &= \sum_{i=1}^{|\mathcal{D_{\mathcal{T}}|}}\frac{1}{|\mathcal{D}_{\mathcal{T}}|}\left[\sum_{j=1}^{|\mathcal{D}_{\mathcal{T}}|}\frac{1}{|\mathcal{D}_{\mathcal{T}}|}k(\boldsymbol{z}_i, \boldsymbol{z}^{'}_j)^2-\left(\sum_{j=1}^{|\mathcal{D}_{\mathcal{T}}|}\frac{1}{|\mathcal{D}_{\mathcal{T}}|}k(\boldsymbol{z}_i, \boldsymbol{z}^{'}_j)\right)^2\right].
\end{aligned}
\]

Then, we study the high-order moment term:
\[
\begin{aligned}
    \mathcal{O}\left(k(\boldsymbol{z}, \boldsymbol{z}^{'})\right)&=\frac{1}{|\mathcal{D}_{\mathcal{T}}|}\sum_{i=1}^{|\mathcal{D}_{\mathcal{T}}|}\sum_{j=1}^{N_C}\sum_{\boldsymbol{z}^{'}\in\mathcal{C}_j}\frac{\left[k(\boldsymbol{z}_i, \boldsymbol{z}^{'})-\sum_{\boldsymbol{z}^{'}\in\mathcal{Z}}\frac{1}{|\mathcal{D}_{\mathcal{T}}|}k(\boldsymbol{z}_i, \boldsymbol{z}^{'})\right]^2}{2N_C|\mathcal{C}_j|}\\
    &\ge \frac{1}{|\mathcal{D}_{\mathcal{T}}|}\sum_{i=1}^{|\mathcal{D}_{\mathcal{T}}|}\sum_{\boldsymbol{z}^{'}\in\mathcal{Z}}\frac{\left[k(\boldsymbol{z}_i, \boldsymbol{z}^{'})-\sum_{\boldsymbol{z}^{'}\in\mathcal{Z}}\frac{1}{|\mathcal{D}_{\mathcal{T}}|}k(\boldsymbol{z}_i, \boldsymbol{z}^{'})\right]^2}{2N_CC_{\rm max}}\\
    &=\frac{|\mathcal{D}_{\mathcal{T}}|}{2N_CC_{\rm max}}\sum_{i=1}^{|\mathcal{D}_{\mathcal{T}}|}\frac{1}{|\mathcal{D}_{\mathcal{T}}|}\left[\sum_{j=1}^{|\mathcal{D}_{\mathcal{T}}|}\frac{1}{|\mathcal{D}_{\mathcal{T}}|}k(\boldsymbol{z}_i, \boldsymbol{z}^{'}_j)^2-\left(\sum_{j=1}^{|\mathcal{D}_{\mathcal{T}}|}\frac{1}{|\mathcal{D}_{\mathcal{T}}|}k(\boldsymbol{z}_i, \boldsymbol{z}^{'}_j)\right)^2\right]\\
    &\ge \frac{|\mathcal{D}_{\mathcal{T}}|}{2N_CC_{\rm max}}\mathrm{HSIC}(Z, Z),
\end{aligned}
\]
where $C_{\rm max} \ge |\mathcal{C}_c|$ for $\forall c\in\{1, 2, ..., N_C\}$.

\begin{algorithm}[tb]
   \caption{Maximizing Optimized Kernel Dependence Algorithm}
   \label{Alg:MOKD}
    \begin{algorithmic}
   \STATE {\bfseries Input:} pre-trained backbone $f_{\phi^*}$, number of inner iterations $n$, learning rate $\eta$, linear transformation parameters $h_{\theta}$, a list of bandwidths $\Sigma=\{\sigma_1, \sigma_2, ..., \sigma_T\}$, and $\epsilon=1e-5$.
   
   \STATE {\bfseries Output:} the optimal parameters for linear transformation head $\theta^*$.

   \emph{\textcolor{blue}{\# Sample a task}}
   \STATE \textbf{Sample} a new task $\mathcal{T}=\{\{\boldsymbol{X}^{\rm s}, Y^{\rm s}\}, \{\boldsymbol{X}^{\rm q}, Y^{\rm q}\}\}$;\\
   \STATE \textbf{Obtain} the representations: $\mathcal{Z}=\{h_{\theta}\circ f_{\phi^*}(\boldsymbol{x}_i)\}_{i=1}^{|\boldsymbol{X}^{\rm s}|}$;
   
   \emph{\textcolor{blue}{\# Inner optimization for test power maximization}}
   \STATE \textbf{Maximize} the test power of $\widehat{{\rm HSIC}}(Z, Y; \sigma_{ZY}, \theta)$ and $\widehat{{\rm HSIC}}(Z, Z; \sigma_{ZZ}, \theta)$ with Eq.~(\ref{Eq:HSIC_estimation_song}) and~(\ref{Eq:variance_hsic_estimation}):\\
   \quad\quad $\sigma^*_{ZY}=\max_{\Sigma}\frac{\widehat{{\rm HSIC}}(Z, Y; \sigma_{ZY}, \theta)}{\sqrt{v_{ZY}+\epsilon}}$; $\sigma^*_{ZZ}=\max_{\Sigma}\frac{\widehat{{\rm HSIC}}(Z, Z; \sigma_{ZZ}, \theta)}{\sqrt{v_{ZZ}+\epsilon}}$
   
   \emph{\textcolor{blue}{\# Outer optimization for dependence optimization}}

   \FOR{$i=1$ {\bfseries to} $n$}
    \STATE \textbf{Obtain} the representations: $\mathcal{Z}=\{h_{\theta}\circ f_{\phi^*}(\boldsymbol{x}_i)\}_{i=1}^{|\boldsymbol{X}^{\rm s}|}$
    \STATE \textbf{Compute} $\widehat{{\rm HSIC}}(Z, Y, \sigma^*_{ZY}, \theta)$ and $\widehat{{\rm HSIC}}(Z, Z; \sigma^*_{ZZ}, \theta)$ with Eq.~(\ref{Eq:HSIC_estimation_song}) for loss:\\
    \quad\quad$\mathcal{L}(Z, Y; \theta)=-\widehat{{\rm HSIC}}(Z, Y, \sigma^*_{ZY}, \theta)+\gamma\widehat{{\rm HSIC}}(Z, Z; \sigma^*_{ZZ}, \theta)$

    \STATE \textbf{Update} parameters:\\
    \quad\quad $\theta \gets \theta - \eta\nabla_{\theta}\mathcal{L}(Z, Y; \theta)$
   \ENDFOR

\end{algorithmic}
\end{algorithm}

\section{Differences between SSL-HSIC and MOKD}\label{Appendix:difference_sslhsic_mokd}
In this paper, we propose a bi-level optimization framework MOKD, which is inspired by a new interpretation of NCC-based loss from the perspective of kernel dependence measure. We find that the core insight of NCC-based loss is learning a set of class-specific representations, where the similarities among samples within the same class are maximized while the similarities between samples from different classes are minimized. Our proposed MOKD method achieves the same goal by optimizing the dependence respectively between representations and labels and among all representations based on the optimized HSIC measures where the test power of the kernels used is maximized. 

However, we notice that the outer optimization objective in Eq.~(\ref{Eq:mokd_problem}) of our proposed MOKD method shares a similar format as the objective of SSL-HSIC~\citep{sslhsic}. From our perspective, such a similar format mainly results from two aspects. On the one side, the outer optimization objectives of NCC-based loss and InfoNCE share the same softmax-like structure. On the other hand, we adopt a similar label kernel as SSL-HSIC by adapting it to few-shot classification settings. Even though, actually, there are two major differences between these two objectives.

Firstly, the most obvious difference between SSL-HSIC and MOKD is that MOKD takes the test power of kernel HSIC measures into consideration. As aforementioned, a challenge of applying HSIC to few-shot classification tasks is that the kernels used may sometimes fail to accurately measure the dependence between the given two data samples. As a result, the transformation model may fail to learn a set of class-specific representations where the similarities among samples belonging to the same class are maximized while the similarities between samples from different classes are minimized. Such a phenomenon may further induce uncertainty and result in misclassification of samples. Thus, by introducing test power maximization in HSIC, kernels' capability of detecting dependence between data samples is improved. This facilitates increasing the sensitivity of kernel HSIC to dependence and further contributes to dependence optimization. 

In addition, SSL-HSIC and MOKD are derived from different learning frameworks and are designed for different task settings. To be specific, SSL-HSIC is derived from the InfoNCE loss~\citep{infonce} that is designed for unsupervised contrastive learning and focuses on learning robust and discriminative representations of a sample by contrasting two different views of samples. The ultimate goal of SSL-HSIC is to learn a good feature encoder for downstream tasks. However, MOKD is derived from NCC-based loss (a.k.a., Prototypical loss)~\citep{prototypical} that is designed for supervised few-shot classification and aims at learning a set of class-specific representations where the similarities among samples within the same class are maximized while the similarities between samples from different classes are minimized. The ultimate goal of MOKD is to learn the optimal task-specific parameters (of a linear transformation head) for each task to extract a set of class-specific representations where the data clusters are well learned and the undesirable high similarities are alleviated.

In order to compare SSL-HSIC with MOKD, we further conduct an experiment to reveal the differences between the two learning frameworks. In this experiment, HSIC measures used in SSL-HSIC are estimated in the same unbiased way as MOKD. The results are reported in Table~\ref{Table:comparision_mokd_sslhsic}. As we can observe, MOKD outperforms SSL-HSIC and SSL-HSIC with Test Power Maximization on all datasets of Meta-Datasets. Moreover, an interesting phenomenon is that SSL-HSIC achieves better performance when applying test power maximization to kernels used in SSL-HSIC. This strongly demonstrates that test power facilitates capturing the dependence between data samples and in turn, learning better representations for each class in the given support set. Compared with MOKD, the main difference between the two objectives is the ${\rm HSIC}(Z, Z)$ term. Specifically, in SSL-HSIC loss, the term is ${\rm HSIC}(Z, Z)$ is modified to $\sqrt{{\rm HSIC}(Z, Z)}$ to achieve better performance in practice.  However, in the original theoretical results of SSL-HSIC, the term should be ${\rm HSIC}(Z, Z)$. In this paper, since we propose to maximize the test power via ${\rm HSIC}(\cdot, \cdot)$, the modification may potentially result in a mismatch of test power.

\begin{table}[t]
\vspace{-1.0em}
	\caption{\textbf{Comparisons of MOKD and SSL-HSIC.}}
	\label{Table:comparision_mokd_sslhsic}
	\begin{center}
		\begin{tiny}
		\setlength\tabcolsep{4pt}
			\begin{threeparttable}
				\begin{tabular}{lccccccccccccc}
					\toprule
					Datasets & ImageNet & Omniglot & Aircraft & Birds & DTD & QuickDraw & Fungi & VGG\_Flower & Traffic Sign & MSCOCO & MNIST & CIFAR10 & CIFAR100 \\
					\midrule
					MOKD  & \bf{57.3$\pm$1.1} & \bf{94.2$\pm$0.5} & \bf{88.4$\pm$0.5} & \bf{80.4$\pm$0.8} & \bf{76.5$\pm$0.7} & \bf{82.2$\pm$0.6} & \bf{68.6$\pm$1.0} & \bf{92.5$\pm$0.5} & \bf{64.5$\pm$1.1} & \bf{55.5$\pm$1.0} & \bf{95.1$\pm$0.4} & \bf{72.8$\pm$0.8} & \bf{63.9$\pm$1.0}\\
					SSL-HSIC       & 56.5$\pm$1.2 & 92.0$\pm$0.9 & 87.3$\pm$0.7 & 78.1$\pm$1.1 & 75.2$\pm$0.8 & 81.4$\pm$0.7 & 63.5$\pm$1.2 & 90.9$\pm$0.8 & 59.7$\pm$1.3 & 51.4$\pm$1.1 & 93.4$\pm$0.6 & 70.0$\pm$1.1 & 61.8$\pm$1.1\\
                    SSL-HSIC(TPM) & 56.9$\pm$1.1 & 92.6$\pm$0.9 & 87.5$\pm$0.6 & 79.8$\pm$0.9 & 75.7$\pm$0.7 & 82.0$\pm$0.7 & 67.1$\pm$1.0 & 91.4$\pm$0.6 & 62.4$\pm$1.0 & 53.6$\pm$1.0 & 94.3$\pm$0.5 & 71.5$\pm$0.8 & 63.5$\pm$1.0 \\
					\bottomrule
				\end{tabular}
			\end{threeparttable}
		\end{tiny}
	\end{center}
	\vskip -0.1in
	\vspace{-1.0em}
\end{table}

\section{More Settings for CFC}\label{Appendix:detailed_settings}
In this section, we provide more details about cross-domain few-shot classification task settings. Specifically, detailed information on Meta-Dataset, experimental settings, data split settings, and vary-way vary-shot settings are introduced.

\subsection{Introduction to Meta-Dataset}
Meta-Dataset was first proposed by~\citet{metadatasets} as a CFC benchmark. The selected datasets are free and easy to obtain and span various visual concepts with different degrees in fine-grain. The original Meta-Dataset is composed of 10 datasets that are ILSVRC\_2012~(ImageNet)~\citep{imagenet}, Omniglot~\citep{omniglot}, FGVC\_Aircraft~(Aircraft)~\citep{aircraft}, CUB\_200-2011~(CU\_Birds)~\citep{cu_birds}, Describable Textures~(DTD)~\citep{dtd}, Quick Draw~\citep{quickdraw}, FGVCx Fungi~(Fungi)~\citep{fungi}, VGG\_Flower~(Flower)~\citep{vgg_flower}, Traffic Sign~\citep{traffic_sign}, MSCOCO~\citep{mscoco}. Then, MNIST~\citep{mnist}, CIFAR-10~\citep{cifar} and CIFAR-100~\citep{cifar} were added by \citet{CNAPS}.

\textbf{ILSVRC\_2012}. ImageNet is a dataset composed of natural images from 1000 categories. In Meta-Dataset, some images that are duplicates of other datasets (e.g. 43 images in CU\_Birds) are removed.

\textbf{Omniglot.} Omniglot is a dataset composed of 1623 handwritten characters. The dataset contains 50 classes with 20 examples in each class. The split of Omniglot dataset follows \citet{omniglot}.

\textbf{Aircraft.} Aircraft is a dataset of images of aircraft spanning 102 model variants, and each class contains 100 images. In Meta-Dataset, the images are cropped according to the provided bounding boxes in case of including other aircrafts or the copyright texts.

\textbf{CU\_Birds (CUB-200-2011).} CU\_Birds is a dataset for fine-grained classification of 200 different bird species. The images in CU\_Birds did not use the provided bounding boxes for harder challenges.

\textbf{Describable Textures.} DTD dataset consists of 5640 images of textures, which are organized according to a list of 47 categories inspired from human perception.

\textbf{Quick Draw.} A dataset of 50 million black-and-white drawings across 345 categories.

\textbf{Fungi.} Fungi is a dataset of 1500 wild mushroom species and contains about 100K images.

\textbf{VGG Flower.} VGG Flower is a dataset of natural images of 102 flower categories. Each class contains 40$\sim$258 images.

\textbf{Traffic Sign.} Traffic Sign is a dataset of 50K images of German road signs in 43 classes.

\textbf{MSCOCO.} MSCOCO is a dataset of images collected from Flickr with 1.5 million object instances belonging to 80 categories labeled and localized using bounding boxes. The version adopted in Meta-Dataset is the train2017 split and the images are created images from original images using each object instance's groundtruth bounding box.

\textbf{MNIST.} MNIST is a dataset of handwritten digits. MNIST is composed of 60K training images and 10K test images.

\textbf{CIFAR-10 \& CIFAR-100.} CIFAR-10 is a dataset of 60K 32$\times$32 color images in 10 different classes that are airplanes, cars, birds, cats, deer, dogs, frogs, horses, ships, and trucks. CIFAR-100 is a dataset with 100 classes and each class contains 500 training data and 100 test data. Each image in CIFAR-100 owns both a ``fine'' label and a ``coarse'' label.

\subsection{Task Settings}\label{Appendix:task_settings}
In this paper, there are two main experimental settings for our main results: ``train on all datasets'' and ``train on ImageNet only''. In ``train on all datasets'' settings, the backbone we use is the multi-domain backbone which has observed training data of all 8 domains (ImageNet, Omniglot, Aircraft, CU\_Birds, DTD, QuickDraw, Fungi and VGG\_Flower). In the ``train on ImageNet only'' settings, the backbone we use is the single-domain backbone which is trained only on the training data of the ImageNet dataset. For both sets of settings, during the meta-test phase, the evaluation is performed on the test data of seen domains and data from unseen domains.

Moreover, for simplicity, the default setting in this paper is ``train on all datasets'' settings if not any specific clarification.

\subsection{Split Settings}

The splits of the datasets in this paper are consistent with those in Meta-Dataset. For example, under ``train on all datasets'' settings, ImageNet, Omniglot, Aircraft, Birds, DTD, QuickDraw, Fungi, and VGG Flower are preserved as `seen domains' where the training set of each dataset are accessible for training the backbone. Each dataset of the seen domain is divided into a training set, a validation set, and a test set roughly with the proportions of 75\%, 15\%, and 15\%. Specifically, for ImageNet, Meta-Dataset constructs a sub-graph of the overall DAG that describes the relationships among all 82115 `synsets' in ILSVRC\_2012. Then, the entire graph is cut into three pieces for training, validation, and testing without overlap.

\subsection{Vary-way Vary-shot Settings}\label{Appendix:vary_settings}
Vary-way vary-shot task is a popular and basic task setting in cross-domain few-shot classification~\citep{metadatasets,SUR,urt,crosstransformer,url,tsa}. Such task settings stimulate the common daily situations where there exist distribution gaps among tasks and the data in the given task are imbalanced. Compared with conventional few-shot classification task settings where the numbers of ways and shots are fixed and tasks for the test are sampled from unseen data sets with the same distribution, the vary-way vary-shot task is more challenging due to imbalanced data and distributional discrepancies between source and target domains.

In the context of cross-domain few-shot classification, a vary-way vary-shot task is sampled from a single dataset for each learning episode.
Generally, the sampling process of a vary-way vary-shot task mainly includes two independent steps: sampling a set of classes and sampling support and query data from the sampled classes. 
We only provide a brief introduction to the task sampling process, for more details, please refer to the paper of Meta-Dataset~\cite{metadatasets}.

\paragraph{Class Sampling} Given a dataset, the number of ways (classes) $N_C$ is sampled uniformly from the interval $[5, N_{\rm max}]$, where $N_{\rm max}$ denotes the maximum of the number of classes. Usually, $N_{\rm max}$ is either 50 or as many classes as available.

\paragraph{Support and Query Data Sampling} After a set of classes is sampled, the numbers of shots for support and query sets are respectively determined by the following rules.

\underline{\textbf{Compute query set size.}} In vary-way vary-shot task settings, the number of query data of each class in a task is fixed to the same number. The fixed number should be no more than half of the total number of data in the given class so that there are still roughly 50\% of data being used as support data. The process is formulated as:
\[
    q = \min\left\{10, \left(\min_{c\in \mathcal{C}}\lfloor0.5*|c|\rfloor\right) \right\},
\]
where $\mathcal{C}$ denotes a set of selected classes, $c$ denotes a single class and $|c|$ denotes the number of images in the given class $c$. In order to avoid a too large query set, the maximum number of query data of each class is set to 10.

\underline{\textbf{Compute support set size.}} The computation of support set size is formulated as:
\[
    s = \min\left\{500, \sum_{c\in \mathcal{C}}\lceil\beta\min\left\{100, |c|-q\right\}\rceil\right\},
\]
where $\beta$ is a coefficient sampled uniformly from $(0, 1]$. In vary-way vary-shot task settings, the total number of data in a support set of a task is no more than 500. For each class in the selected set, the number of shots is determined by its remaining data where query data have been excluded. The maximum number of shots for each class is 100. The coefficient $\beta$ is used to sample a smaller number of support data and generate a task with an imbalanced number of shots.

\underline{\textbf{Data Sampling for Each Class.}} After the support set size is determined, the number of shots for each class is calculated. First of all, $N_C$ random scalars $\{\alpha_1, \alpha_2, ..., \alpha_{N_C}\}$ are uniformly sampled from the interval $[\log(0.5), \log(2))$. Then, their `contributions' to the support set are calculated as:
\[
    R_c = \frac{\exp(\alpha_c)|c|}{\sum_{c^{'}\in\mathcal{C}}\exp(\alpha_{c^{'}})|c^{'}|}.
\]
Then, the number of shots for class $c$ can be calculated by:
\[
    K_c = \min\left\{\lfloor R_c * (s-|\mathcal{C}|)\rfloor+1, |c|-q\right\}.
\]
The term $R_c * (s-|\mathcal{C}|)\rfloor+1$ is to guarantee that there is at least one sample being select for the class.

\section{More Experimental Settings}\label{section:more_experimental_details}
\subsection{Pre-trained Backbone} 
In this paper, we directly use both multi-domain and single-domain ResNet-18~\citep{resnet} backbones provided by URL repository\footnote{\href{https://github.com/VICO-UoE/URL}{https://github.com/VICO-UoE/URL}} for simplicity and fairness. Two kinds of backbones are respectively applied in our experiments according to different experimental settings. For ``train on ImageNet only'' settings, the pre-trained backbone applied is a single domain-specific backbone that is trained only on the ImageNet dataset. For ``train on all datasets'' settings, the pre-trained backbone applied is a multi-domain backbone. The multi-domain backbone is distilled from 8 single domain-specific pre-trained backbones. More details about model distillation are available in \cite{url}.

For simplicity, except for specific clarification, the experiments are conducted on the multi-domain backbone under the ``train on all datasets'' settings.
In practice, we directly use both multi-domain and single-domain backbones provided in URL repository in order to make fair comparisons.

\subsection{More Implementation Details}
In this paper, we follow most settings in URL~\citep{url} to train a simple linear head on top of a pre-trained backbone. 

\paragraph{Initialization \& Learning rate.} For each adaptation episode,  we re-initialize the linear transformation layer with an identity matrix and learn a set of task-specific parameters for the given task. The optimizer used in MOKD is Adadelta~\citep{adadelta}. The learning rate is 1.0 for Traffic Sign and MNIST and 0.25 for the remaining datasets. Besides, the weight decay is set to 0.25 for seen domains and 0.0 for unseen domains.

\paragraph{Values of $\gamma$.} In vary-way vary-shot task settings, we intuitively set $\gamma$ to 1.0 for Omniglot, Aircraft, CU\_Birds, Quick Draw and MNIST while 3.0 for other datasets. Since datasets like Omniglot and Aircraft are simple and the main object of each image is salient, small $\gamma$ is enough. In contrast, since datasets like ImageNet and Fungi are complex and each image contains too much semantic information, large $\gamma$ is required to penalize the high-variance representations and alleviate the overfitting.

In addition, in vary-way 5-shot and 5-way 1-shot task settings, we respectively set $\gamma$ to 1.0 and 0 for all datasets since there are only a few data samples in each task.

\paragraph{Hardware \& Seed settings.} In this paper, all experiments are performed on an NVIDIA GeForce RTX 3090 GPU. The GPU memory required for running MOKD is about 5 GB. For fairness, all baselines of URL and experiments on MOKD are performed with seeds 41, 42, 43, 44, 45.

\subsection{Adaptive Bandwidth Selection}
Bandwidth is an essential component of a kernel (such as Gaussian kernel) since it closely corresponds to the test power of the kernel as shown in this paper. It is widely believed that kernels with large test power are more sensitive to the dependence among data. As the ultimate goal of this paper is to learn class-specific representations by maximizing dependence among samples belonging to the same class and minimizing the dependence among all samples, a viable way is to select a suitable bandwidth to maximize the test power of the kernel. 
To this end, we first perform bandwidth selection before optimizing the objective loss to maximize its test power so that the optimized kernel is much more sensitive to the dependence. To be concrete, the test power is maximized by selecting an optimal bandwidth to maximize $\frac{{\rm HSIC}(\cdot, \cdot; \sigma, \theta)}{\sqrt{v+\epsilon}}$, where, $\sigma$ denotes the bandwidth, $v$ denotes the variance of HSIC, and $\epsilon$ is a constant that aims to avoid $v\leq 0$.

First of all, bandwidth is initialized as the median of the Gram matrix obtained with the data. Then, we manually set a list of coefficients to scale the median as the new bandwidth. To be concrete, the scale coefficient is selected from the list [0.001, 0.01, 0.1, 0.2, 0.25, 0.5, 0.75, 0.8, 0.9, 1.0, 1.25, 1.5, 2.0, 5.0, 10.0]. Finally, by iteratively calculating $\frac{{\rm HSIC}(\cdot, \cdot; \sigma, \theta)}{\sqrt{v+\epsilon}}$ where $\sigma$ is a scaled median, we select the $\sigma$ which obtains the largest test value of $\frac{{\rm HSIC}(\cdot, \cdot; \sigma, \theta)}{\sqrt{v+\epsilon}}$ as the optimal bandwidth. 

The reason that we chose the grid search method for the optimal bandwidth is the efficiency of MOKD. Optimizing bandwidth with auto optimizer requires extra hyperparameter selection and gradient descent steps, and this extra work will make the algorithm complicated and time-consuming.

\section{Detailed Experimental Results}
\subsection{Results Under Vary-way Vary-shot Settings}
In this section, we evaluate MOKD on vary-way vary-shot tasks under both ``train on all datasets'' and ``train on ImageNet only'' settings. To be clear, we mark seen domains with green while unseen domains with red.

\subsubsection{Results under Train on ImageNet Only Settings}
The empirical results under ``train on ImageNet only'' settings are reported in Table~\ref{Table:imagenet_only} with mean accuracy and 95\% confidence. Here, we provide a more detailed analysis of the results.

Generally, MOKD achieves the best performance among all approaches on 10 out of 13 datasets, including ImageNet, Omniglot, Textures (DTD), Quick Draw, Fungi, VGG\_Flower, MSCOCO, MNIST, CIFAR10, and CIFAR100, and ranks 1.3 on average of all datasets. Compared with URL, where MOKD is based, MOKD outperforms URL on almost all datasets. Specifically, compared with URL, MOKD obtains 1.5\%, 0.2\%, 0.7\%, 0.9\%, 3.3\%, 0.8\%, 1.6\%, 0.4\%, 2.3\%, 2.1\%, 1.2\% and 1.1\% improvements respectively from Omniglot to CIFAR-100. For ImageNet, which is the seen domain, MOKD gets the same results as URL and outperforms other previous works. 

An interesting phenomenon is that MOKD performs better than URL on unseen domains compared with the results on seen domains. As we can see from the table, MOKD achieves 1.5\% improvements on average on unseen domains. Such a phenomenon indicates that MOKD has better generalization ability than previous works. Due to there exist distribution gaps between seen and unseen domains, it is challenging for a model to perform well on the domains that it has never observed before. We guess the reason for such a phenomenon is that MOKD directly optimizes the dependence respectively between representations and labels and representation themselves with the optimized kernel HSIC where the test power is maximized to be more sensitive to dependence. Thus, it is able to learn a set of better representations where similarities among samples within the same class are maximized while similarities between samples from different classes are minimized via capturing the accurate dependence between representations and labels.

\subsubsection{Results Under Train on All datasets Settings}\label{Appendix:complete_res_all_datasets}
The results under ``train on all datasets'' settings are reported in Table~\ref{Table:all_datasets} with mean accuracy and 95\% confidence. Here, we intend to provide a more detailed analysis of empirical results.

According to the table, it is easy to observe that MOKD achieves the best performance on average and ranks 1.8 among all baselines. Compared with URL where our proposed MOKD is based, MOKD outperforms URL on 10 out of 13 datasets. Specifically, MOKD achieves 0.1\%, 0.2\%, 0.2\%, 0.3\%, 0.6\%, 1.2\%, 1.1\%, 0.4\%, 0.9\% and 1.0\% improvements respectively on Omniglot, Aircraft, CU\_Birds, Textures (DTD), VGG\_Flower, Traffic Sign, MSCOCO, MNIST, CIFAR10 and CIFAR100 datasets. Besides, compared with 2LM which is a recent new state-of-the-art method in the cross-domain few-shot classification community, MOKD still achieves better performance on 8 out of 13 datasets. 

Consistent with the results under ``train on ImageNet only'' settings, MOKD also obtained better performance on unseen domains (Traffic Sign, MSCOCO, MNIST, CIFAR10, and CIFAR100) under ``train on all datasets'' settings. Specifically, under ``train on all datasets'' settings, MOKD achieves 1.2\%, 1.3\%, 0.4\%, 0.9\%, and 1.0\% improvements on Traffic Sign, MSCOCO, MNIST, CIFAR10 and CIFAR100 datasets. Such a phenomenon consistently demonstrates that MOKD is able to obtain better generalization performance on previously unseen domains with only a few learning adaptation steps.

Although MOKD achieves impressive performance on the Meta-Dataset benchmark, we also notice that slight overfitting happens on the Fungi dataset (see Fig.~\ref{Fig:lcurve_fungi}).

\subsubsection{Discussion about Why MOKD Generalizes Well on Unseen Domains}
According to the results under both ``train on all datasets'' and ``train on ImageNet only'' settings, we observe that MOKD achieves better generalization performance on unseen domains than on seen domains. From our perspective, the reasons for this phenomenon are collectively determined by both pre-trained backbones and the optimization objective of MOKD. Specifically, on the one side, the optimization objective of MOKD proposed in Eq. (\ref{Eq:mokd_problem}) is more powerful in exploring class-specific representations; on the other side, the pre-trained backbones limit feature exploration to some extent.

From the perspective of the optimization objective of MOKD, as we have mentioned in Theorem \ref{Theorem:HSIC_Z_Y} of our paper, maximizing ${\rm HSIC}(Z, Y)$ is equivalent to exploring a set of representations that matches the cluster structure of the given task. Meanwhile, since test power maximization is further taken into consideration, MOKD has a more powerful ability to explore such representations compared with NCC-based loss. Our visualization results in Fig. \ref{Fig:intro_fig}, \ref{Fig:remaining_visual_sim_seen}, and \ref{Fig:remaining_visual_sim_unseen} have demonstrated this.

However, the performance is not only simply determined by the optimization objective in Eq. (\ref{Eq:mokd_problem}), but also decided by the pre-trained backbones. In our paper, the backbone used under ``train on all datasets'' settings is pre-trained on 8 datasets, including ILSVRC\_2012, Omniglot, Aircraft, CU\_Birds, DTD, Quick Draw, Fungi and VGG Flowers. Since the distribution is shared between the training and test sets of a single dataset, it is easy for the pre-trained backbone to extract good features, where the cluster structures are definite, from test data of seen domains. However, such an advantage may somewhat constrain the function space that can be explored by the linear transformation head. An intuitive explanation for this conjecture is that a loss will converge to the local optimal if the initial point is close to that local optimal. In contrast, since data from unseen domains have never been observed by the pre-trained model, the features extracted from the pre-trained backbone are not so good. Thus, it is probable that MOKD can find better results from a relatively bad initialization.

\begin{figure}[t]
	\vskip 0.0in
	\begin{center}
	\centering
	\subfigure[ImageNet\label{Fig:lcurve_imagenet}]{
			\begin{minipage}[t]{0.32\linewidth}
				\centering
				\includegraphics[width=1.0\linewidth]{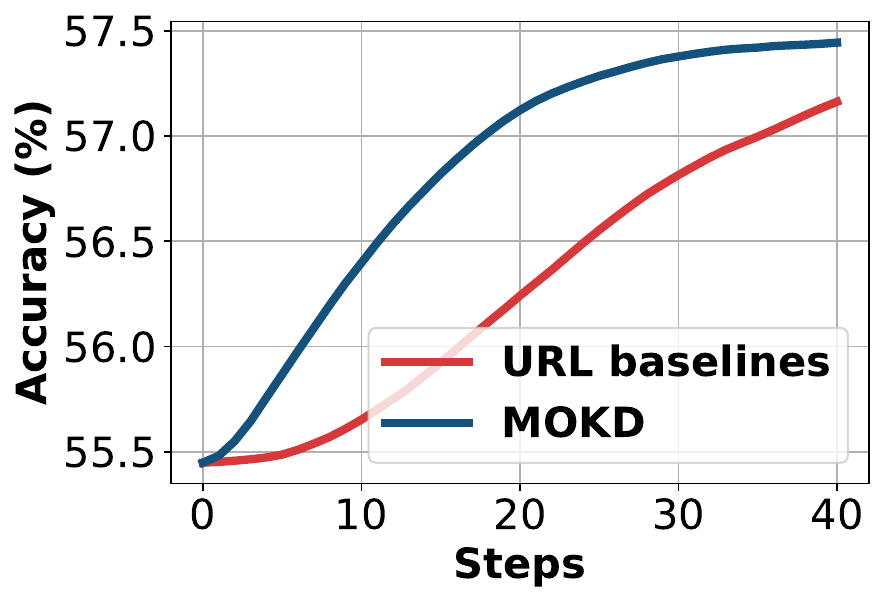}
		\end{minipage}}\vspace{-0.1cm}
	\subfigure[Omniglot\label{Fig:lcurve_omniglot}]{
			\begin{minipage}[t]{0.32\linewidth}
				\centering
				\includegraphics[width=1.0\linewidth]{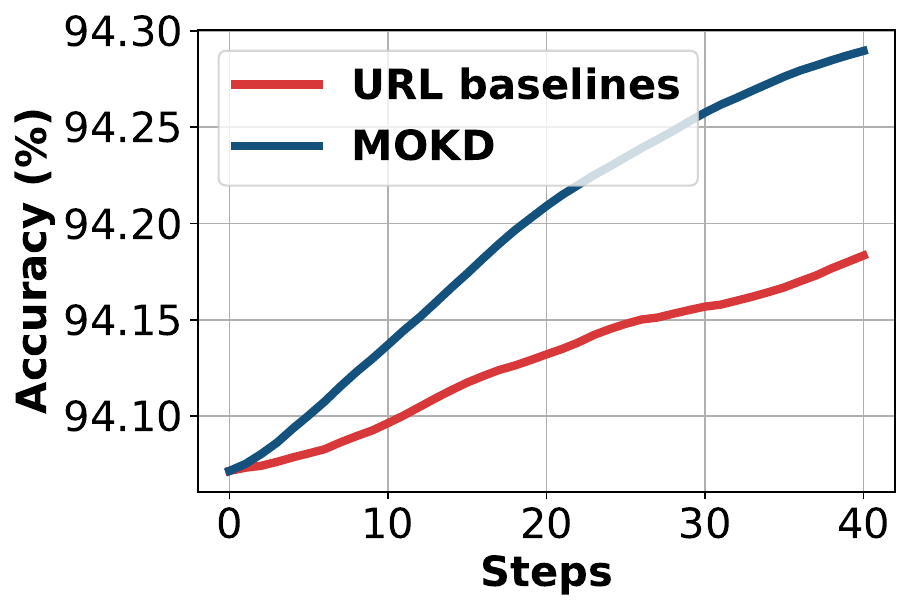}
		\end{minipage}}\vspace{-0.1cm}
	\subfigure[Aircraft\label{Fig:lcurve_aircraft}]{
			\begin{minipage}[t]{0.32\linewidth}
				\centering
				\includegraphics[width=1.0\linewidth]{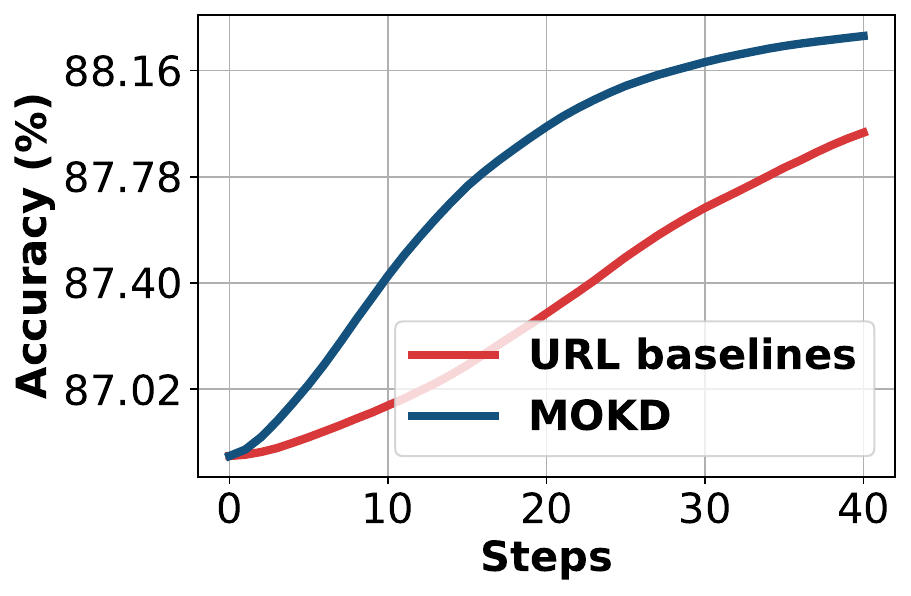}
		\end{minipage}}\vspace{-0.1cm}
    \subfigure[CU\_Birds\label{Fig:lcurve_cubirds}]{
			\begin{minipage}[t]{0.32\linewidth}
				\centering
				\includegraphics[width=1.0\linewidth]{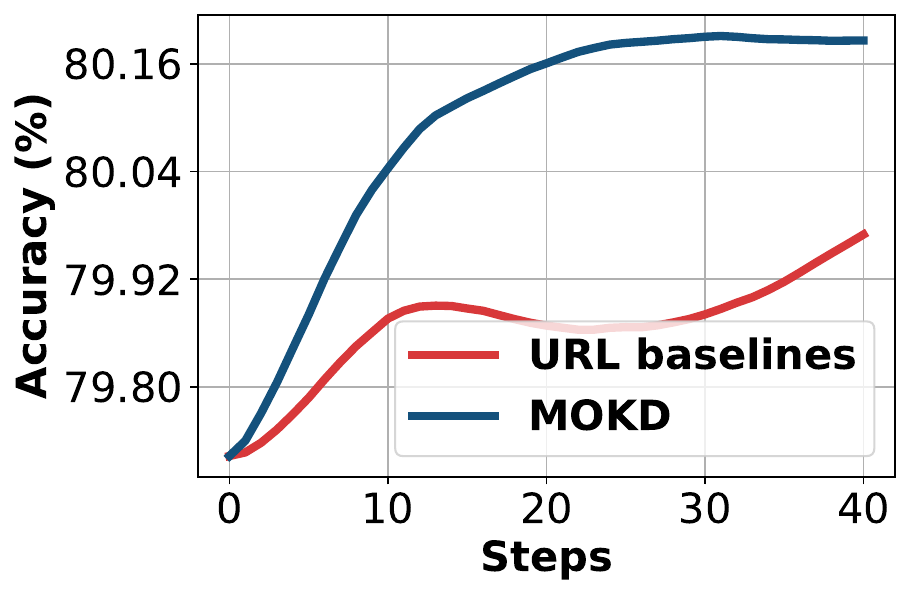}
		\end{minipage}}\vspace{-0.1cm}
    \subfigure[DTD\label{Fig:lcurve_dtd}]{
			\begin{minipage}[t]{0.32\linewidth}
				\centering
				\includegraphics[width=1.0\linewidth]{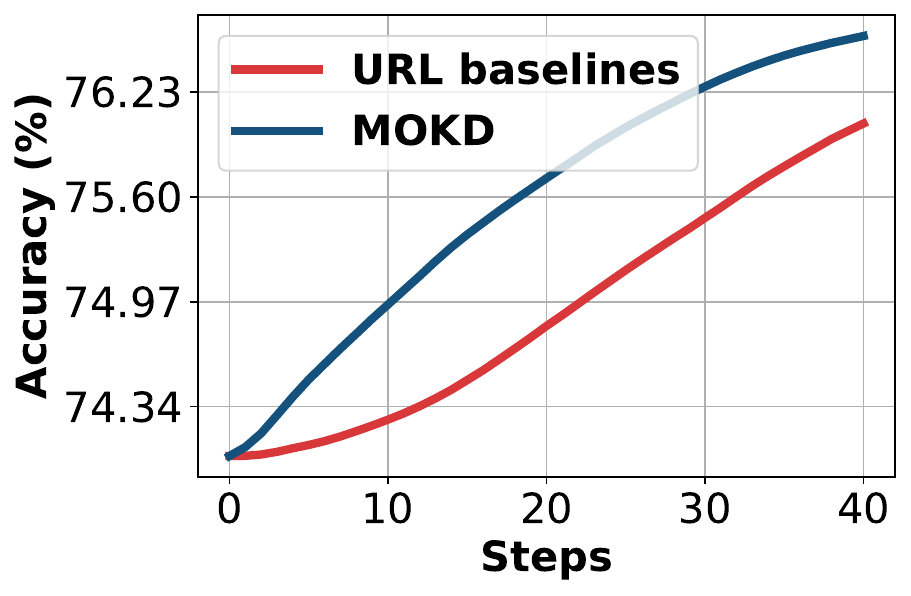}
		\end{minipage}}\vspace{-0.1cm}
    \subfigure[Quick Draw\label{Fig:lcurve_quickdraw}]{
			\begin{minipage}[t]{0.32\linewidth}
				\centering
				\includegraphics[width=1.0\linewidth]{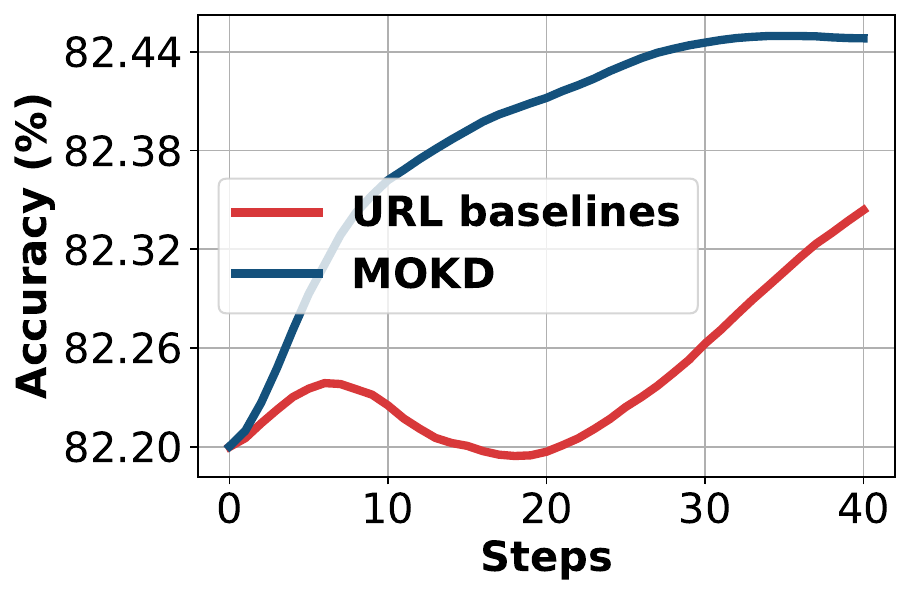}
		\end{minipage}}\vspace{-0.1cm}
	\subfigure[Fungi\label{Fig:lcurve_fungi}]{
			\begin{minipage}[t]{0.32\linewidth}
				\centering
				\includegraphics[width=1.0\linewidth]{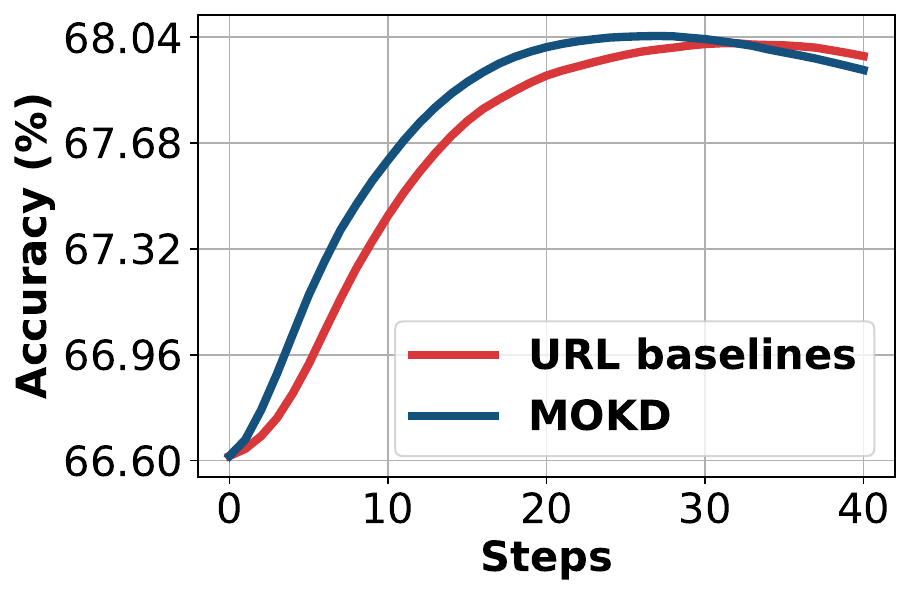}
		\end{minipage}}\vspace{-0.1cm}
	\subfigure[VGG Flower\label{Fig:lcurve_vgg_flower}]{
			\begin{minipage}[t]{0.32\linewidth}
				\centering
				\includegraphics[width=1.0\linewidth]{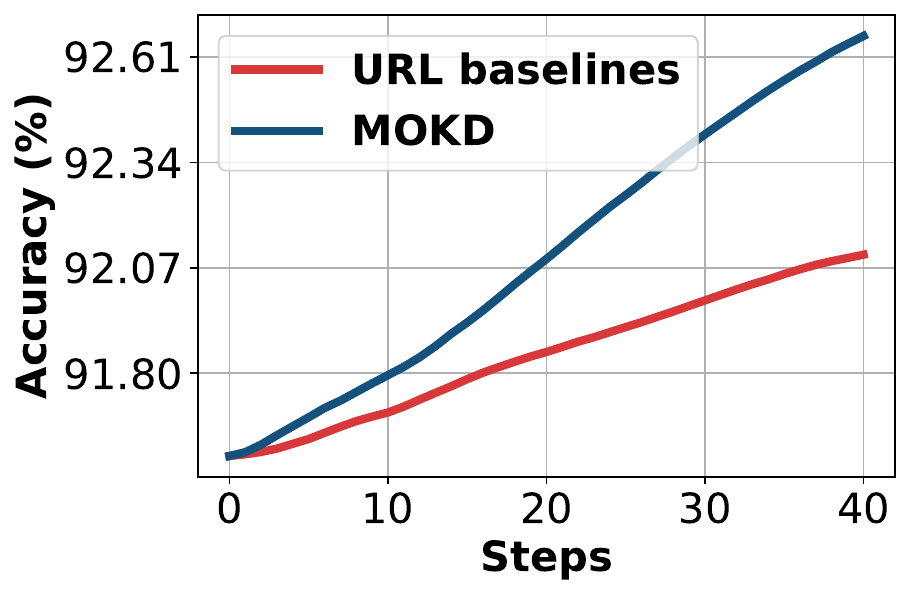}
		\end{minipage}}\vspace{-0.1cm}
    \subfigure[Traffic Sign\label{Fig:lcurve_traffic_sign}]{
			\begin{minipage}[t]{0.32\linewidth}
				\centering
				\includegraphics[width=1.0\linewidth]{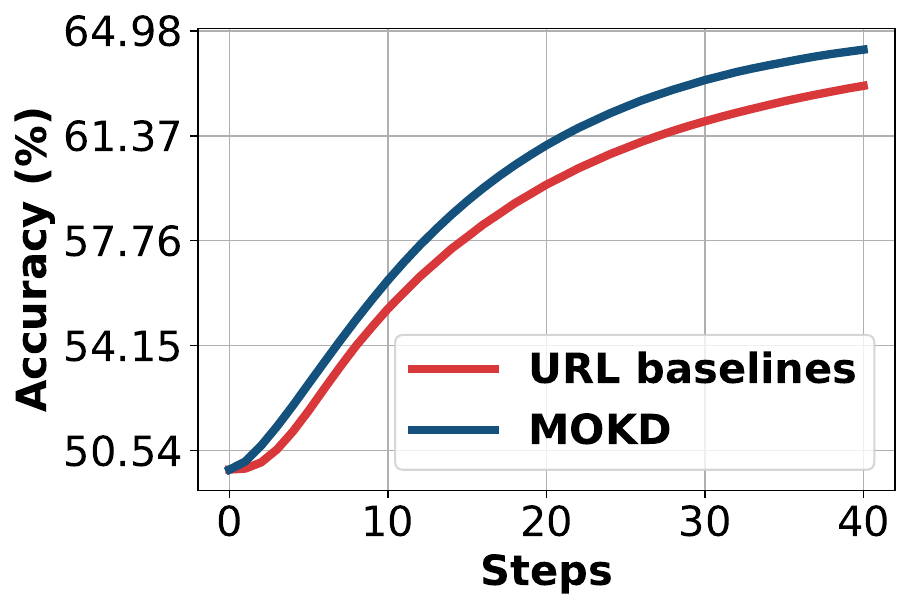}
		\end{minipage}}\vspace{-0.1cm}
    \subfigure[MSCOCO\label{Fig:lcurve_mscoco}]{
			\begin{minipage}[t]{0.32\linewidth}
				\centering
				\includegraphics[width=1.0\linewidth]{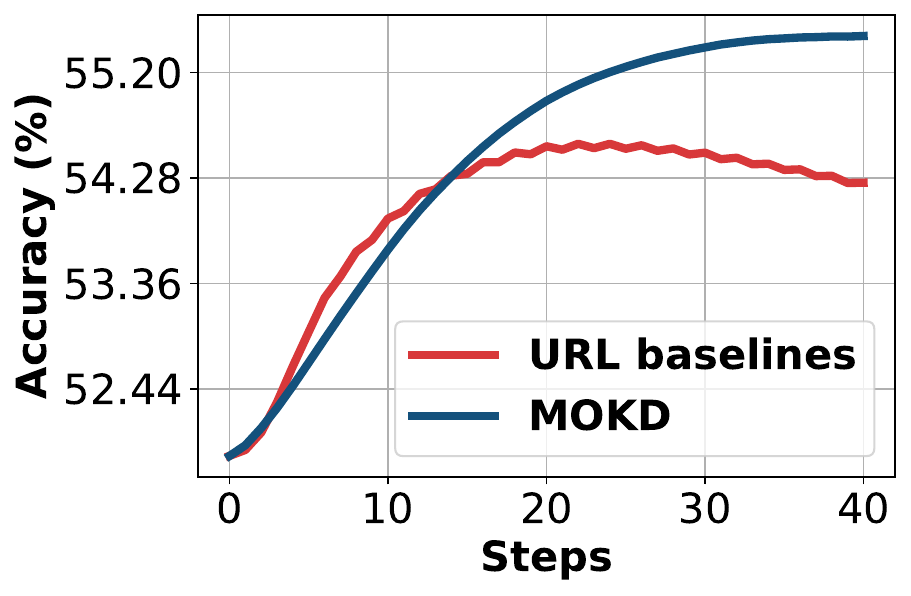}
		\end{minipage}}\vspace{-0.1cm}
    \subfigure[MNIST\label{Fig:lcurve_mnist}]{
			\begin{minipage}[t]{0.32\linewidth}
				\centering
				\includegraphics[width=1.0\linewidth]{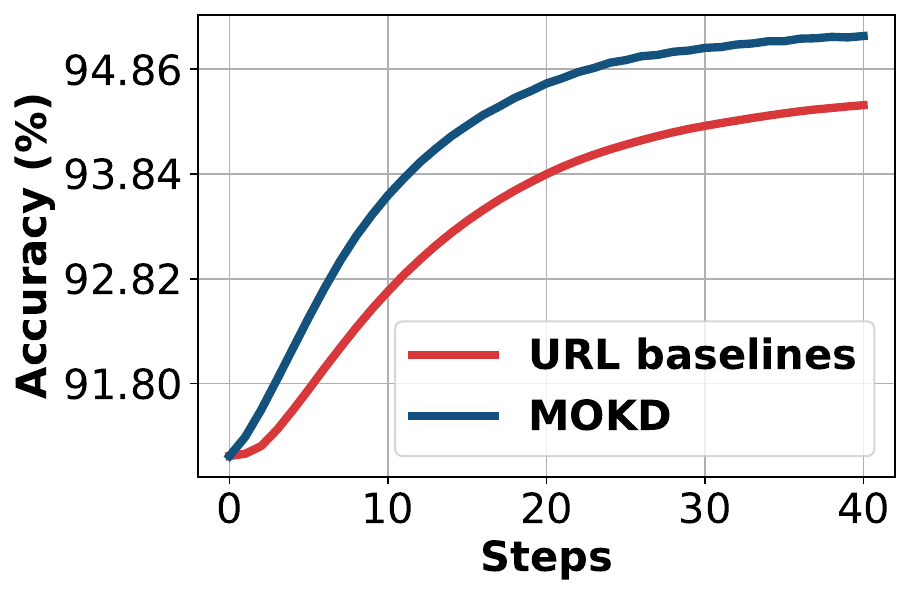}
		\end{minipage}}\vspace{-0.1cm}
    \subfigure[CIFAR 10\label{Fig:lcurve_cifar10}]{
			\begin{minipage}[t]{0.32\linewidth}
				\centering
				\includegraphics[width=1.0\linewidth]{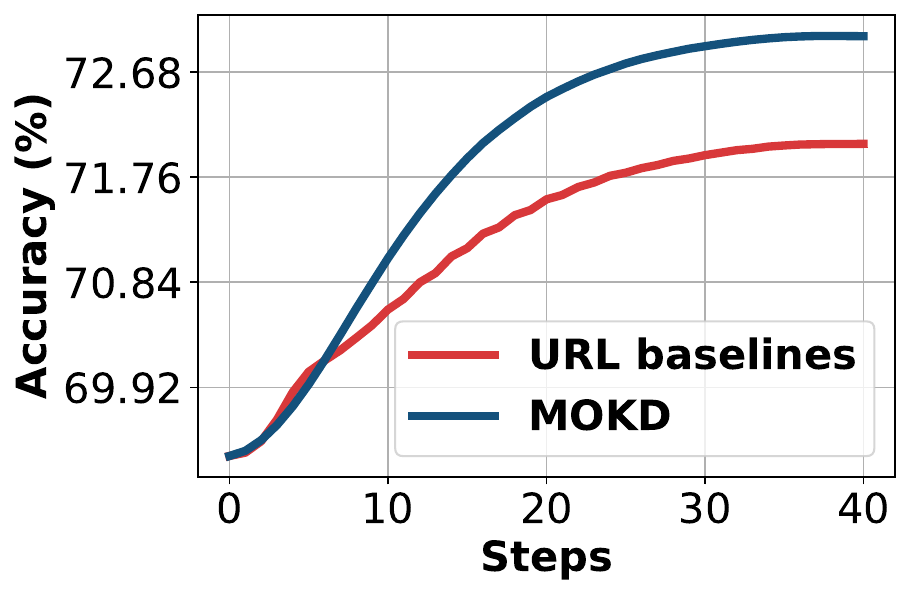}
		\end{minipage}}\vspace{-0.1cm}
	\subfigure[CIFAR 100\label{Fig:lcurve_cifar100}]{
			\begin{minipage}[t]{0.32\linewidth}
				\centering
				\includegraphics[width=1.0\linewidth]{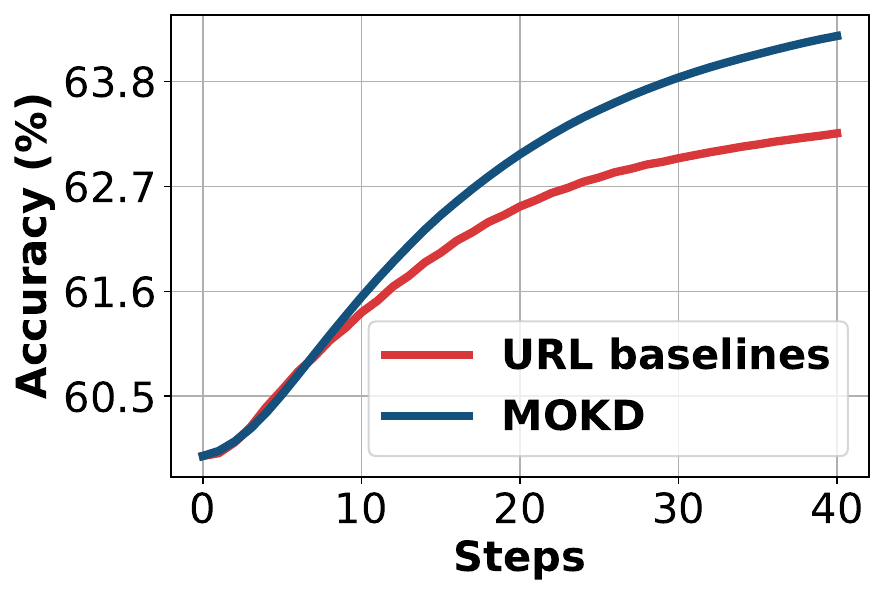}
		\end{minipage}}\vspace{-0.1cm}
    \caption{Test accuracy curves of Meta-Dataset with respect to the steps under ``train on all datasets'' settings. As shown in the figures, MOKD evidently achieves a better learning process and convergence performance compared with URL baseline.}
    \label{Fig:remaining_learn_curves}
	\end{center}
	\vskip -0.2in
	\vspace{1em}
\end{figure}

As a simple demonstration, we compare the performance gaps between the initial and final adaptation steps. The results are obtained with random seed 42 from both URL and MOKD methods. Since both URL and MOKD initialize the linear head with an identity matrix, their initial accuracies before performing adaptation are the same (as shown in Fig. \ref{Fig:remaining_learn_curves} in our paper).

Intuitively, a small performance gap means the extracted features from the pre-trained backbone are good enough for direct classification. Otherwise, the extracted features are not so good. According to the table, we notice that the gaps on seen domains are generally smaller than unseen domains for both URL and MOKD, which demonstrates that the extracted features from unseen domain data are not so good. Based on this observation, as MOKD is more powerful in exploring a set of representations that matches the cluster structure of the given task, better improvements are obtained on unseen domains.

\subsection{Effect of More Trainable Modules}
As demonstrated in previous work~\cite{tsa}, plugging extra trainable modules into the frozen pre-trained backbone contributes to achieving better generalization performance on Meta-Dataset. As MOKD can be seen as a variant case of URL, we also evaluate our proposed MOKD with the TSA strategy.

Specifically, since fine-tuning 4 extra trainable modules consumes more running time (about 30s per iteration), we plug the extra trainable modules only into the second and third resnet block. In addition, in this experiment, we set the learning rate for the trainable module in the backbone to 0.5 for Traffic Sign, MNIST, and CIFAR-100 and 0.05 for the remaining datasets. We also set the learning rate for the transformation head to 1.0 for Traffic Sign, MNIST, and CIFAR-100 and 0.1 for the remaining datasets. The results are reported in Table~\ref{Table:mokd_tsa}.

According to the results in the table, we can observe that MOKD+TSA fails to outperform TSA in most cases. According to our further visualization results of learning curves, we find that severe overfitting happens in the failure cases. Such a phenomenon implies that MOKD strategy may be more suitable for the efficient transformation head adaptation. In the case of more trainable modules in the backbone, MOKD may limit the performance due to overfitting.

\begin{table}[t]
\vspace{-1.0em}
	\caption{\textbf{Comparisons of MOKD and MOKD TSA.}}
	\label{Table:mokd_tsa}
	\begin{center}
		\begin{tiny}
		\setlength\tabcolsep{4.0pt}
			\begin{threeparttable}
				\begin{tabular}{lccccccccccccc}
					\toprule
					Datasets & ImageNet & Omniglot & Aircraft & Birds & DTD & QuickDraw & Fungi & VGG\_Flower & Traffic Sign & MSCOCO & MNIST & CIFAR10 & CIFAR100 \\
					\midrule
					TSA  & \bf{57.7$\pm$1.1} & 94.7$\pm$0.4 & \bf{88.9$\pm$0.5} & \bf{80.7$\pm$0.8} & \bf{77.2$\pm$0.7} & \bf{82.3$\pm$0.6} & \bf{66.8$\pm$1.0} & \bf{92.7$\pm$0.5} & 83.6$\pm$0.9 & \bf{55.2$\pm$1.0} & 96.8$\pm$0.3 & \bf{80.6$\pm$0.8} & \bf{71.2$\pm$0.9}\\
					MOKD+TSA       & 56.2$\pm$1.1 & \bf{94.9$\pm$0.4} & 88.1$\pm$0.6 & 78.6$\pm$0.8 & 73.0$\pm$0.7 & 81.2$\pm$0.5 & 63.9$\pm$1.0 & 92.1$\pm$0.6 & \bf{87.3$\pm$0.8} & 54.5$\pm$1.0 & \bf{97.4$\pm$0.3} & 78.9$\pm$0.8 & 64.7$\pm$0.8\\
					\bottomrule
				\end{tabular}
			\end{threeparttable}
		\end{tiny}
	\end{center}
	\vskip -0.1in
	\vspace{-0.5em}
\end{table}

\begin{table}[t]
\vspace{-1.0em}
	\caption{\textbf{Comparisons of performance gaps between initial and final steps.}}
	\label{Table:performance_gap_init_final}
	\begin{center}
		\begin{tiny}
		\setlength\tabcolsep{6.0pt}
			\begin{threeparttable}
				\begin{tabular}{lccccccccccccc}
					\toprule
					Datasets & ImageNet & Omniglot & Aircraft & Birds & DTD & QuickDraw & Fungi & VGG\_Flower & Traffic Sign & MSCOCO & MNIST & CIFAR10 & CIFAR100 \\
					\midrule
					URL  & 2.00 & 0.15 & 1.36 & 0.27 & 2.42 & 0.19 & 1.40 & 0.58 & 13.94 & 2.30 & 3.59 & 2.77 & 3.52\\
					MOKD       & 2.00 & 0.23 & 1.55 & 0.46 & 2.68 & 0.25 & 1.22 & 1.22 & 14.99 & 3.70 & 4.19 & 3.67 & 4.64\\
					\bottomrule
				\end{tabular}
			\end{threeparttable}
		\end{tiny}
	\end{center}
	\vskip -0.1in
	\vspace{-0.5em}
\end{table}

\subsection{Further Studies on Vary-way 5-shot and 5-way 1-shot}
In this section, we further conduct experiments on more challenging 5-way 1-shot and vary-way 5-shot tasks under the ``train on all datasets'' settings. These tasks are difficult for our proposed MOKD since scarce data have a negative effect on HSIC estimation. For example, less accurate HSIC measures can be estimated with fewer data samples. Besides, maximizing dependence between representations and labels with only extremely few data samples may result in learning biased class-specific representations. The results are reported in Table~\ref{Table:complete_fixedshot}.

\textbf{Vary-way 5-shot.} According to Table~\ref{Table:complete_fixedshot}, MOKD achieves best performance on 6 out of 13 datasets and ranks 1.8 among all baselines on vary-way 5-shot task settings. Generally, MOKD obtains comparable results on Omniglot and Quick Draw compared with the best results but obtains better results on Aircraft, CU\_Birds, VGG\_Flower, and Traffic Sign with improvements $0.5\%$, $0.7\%$, $0.6\%$ and $2.5\%$ respectively. Such results show that MOKD can obtain good performance even if the available data are relatively scarce. However, compared with the results under vary-way vary-shot settings, MOKD fails to achieve impressive performance in general.

\textbf{Five-way One-shot.} The results regarding 5-way 1-shot tasks are reported in Table~\ref{Table:complete_fixedshot}. According to the table, we observe that MOKD fails to achieve the best performance under 5-way 1-shot task settings. We notice that overfitting took place when performing MOKD on 5-way 1-shot tasks. The reason for such a phenomenon is that there is only one sample available for training and MOKD tends to learn excessively biased representations for each class. Even so, MOKD still outperforms other baselines except for URL.

\textbf{Some remarks regarding the empirical results.} As shown in both vary-way 5-shot and 5-way 1-shot task settings, MOKD fails to significantly outperform all baselines. According to the learning curves in our experiments, we find that MOKD tends to overfit the data under these two settings. A reasonable cause for such a phenomenon is that too few data samples negatively affect the estimation of HSIC measures. For example, in 5-way 1-shot task settings, ${\rm HSIC}(Z, Y)$ has to explore representations from only 5 data samples to match the cluster structures of the given support set. Thus, it is highly possible that the learned representations will be extremely biased since there is only one reference sample for each class. Thus, the generalization performance drastically drops. From this perspective, we know that the proposed MOKD is unsuitable for tasks with extremely scarce data samples in each class since the estimated HSIC is not reliable enough to learn a set of good representations that match the cluster structures.

\begin{table}[t]
	\caption{\textbf{Results on vary-way 5-shot and 5-way 1-shot task settings (Trained on All Datasets).} Mean accuracy, 95\% confidence interval reported.}
	\label{Table:complete_fixedshot}
	\begin{center}
		\begin{tiny}
			\setlength\tabcolsep{4.5pt}
			\begin{threeparttable}
				\begin{tabular}{lccccc|ccccc}
					\toprule
					\multirow{2}{*}{Datasets}&
					\multicolumn{5}{c}{\bf Vary-way 5-shot} \vline&
					\multicolumn{5}{c}{\bf 5-way 1-shot}\\
					& Sim-CNAPS & SUR  & URT  & URL &\bf{MOKD}  & Sim-CNAPS & SUR  & URT  & URL &\bf{MOKD}\\
					\midrule
					\rowcolor{green!7}ImageNet    	 & 47.2$\pm$1.0	       & 46.7$\pm$1.0	     & \bf{48.6$\pm$1.0}	& 47.8$\pm$1.0         & 47.5$\pm$1.0          & 42.6$\pm$0.9    & 40.7$\pm$1.0    & \bf{47.4$\pm$1.0}    & 46.5$\pm$1.0    & 46.0$\pm$1.0\\
					\rowcolor{green!7}Omniglot     	 & 95.1$\pm$0.3  	   & 95.8$\pm$0.3	     & \bf{96.0$\pm$0.3}	& 95.8$\pm$0.3	       & \bf{96.0$\pm$0.3}          & 93.1$\pm$0.5    & 93.0$\pm$0.7    & \bf{95.6$\pm$0.5}    & 95.5$\pm$0.5    & 95.5$\pm$0.5\\
					\rowcolor{green!7}Aircraft       & 74.6$\pm$0.6        & 82.1$\pm$0.6	     & 81.2$\pm$0.6	        & 83.9$\pm$0.5	       & \bf{84.4$\pm$0.5}     & 65.8$\pm$0.9    & 67.1$\pm$1.4    & 77.9$\pm$0.9    & \bf{78.6$\pm$0.9}    & \bf{78.6$\pm$0.9}\\
					\rowcolor{green!7}Birds    	     & 69.6$\pm$0.7	       & 62.8$\pm$0.9	     & 71.2$\pm$0.7	        & 76.1$\pm$0.7	       & \bf{76.8$\pm$0.6}     & 67.9$\pm$0.9    & 59.2$\pm$1.0    & 70.9$\pm$0.9    & \bf{76.2$\pm$0.9}    & 75.9$\pm$0.9\\
					\rowcolor{green!7}Textures     	 & 57.5$\pm$0.7	       & 60.2$\pm$0.7	     & 65.2$\pm$0.7	        & \bf{66.8$\pm$0.6}	   & 66.3$\pm$0.6          & 42.2$\pm$0.8    & 42.5$\pm$0.8    & 49.4$\pm$0.9    & \bf{52.0$\pm$0.9}    & 51.4$\pm$0.9\\
					\rowcolor{green!7}Quick Draw     & 70.9$\pm$0.6	       & 79.0$\pm$0.5	     & \bf{79.2$\pm$0.5}	& 78.3$\pm$0.5	       & 78.9$\pm$0.5          & 70.5$\pm$0.9    & \bf{79.8$\pm$0.9}    & 79.6$\pm$0.9    & 79.1$\pm$0.9    & 78.9$\pm$0.9\\
					\rowcolor{green!7}Fungi      	 & 50.3$\pm$1.0	       & 66.5$\pm$0.8	     & 66.9$\pm$0.9	        & 68.7$\pm$0.9	       & \bf{68.8$\pm$0.9}     & 58.3$\pm$0.1    & 64.8$\pm$1.1    & 71.0$\pm$1.0    & \bf{71.4$\pm$1.0}    & 71.1$\pm$1.0\\
					\rowcolor{green!7}VGG Flower     & 86.5$\pm$0.4	       & 76.9$\pm$0.6        & 82.4$\pm$0.5	        & 88.5$\pm$0.4	       & \bf{89.1$\pm$0.4}     & 79.9$\pm$0.7    & 65.0$\pm$1.0    & 72.7$\pm$1.0    & \bf{80.3$\pm$0.8}    & 79.8$\pm$0.8\\
					\midrule
					\rowcolor{red!7}Traffic Sign     & 55.2$\pm$0.8        & 44.9$\pm$0.9	     & 45.1$\pm$0.9	        & 56.7$\pm$0.8	       & \bf{59.2$\pm$0.8}     & 55.3$\pm$0.9    & 44.6$\pm$0.9    & 52.7$\pm$0.9    & \bf{57.4$\pm$0.9}    & 57.0$\pm$0.9\\
					\rowcolor{red!7}MSCOCO           & 49.2$\pm$0.8	       & 48.1$\pm$0.9	     & \bf{52.3$\pm$0.9}	& 51.3$\pm$0.8	       & 51.8$\pm$0.8          & 48.8$\pm$0.9    & 47.8$\pm$1.1    & \bf{56.9$\pm$1.1}    & 52.1$\pm$1.0    & 50.9$\pm$0.8\\
					\rowcolor{red!7}MNIST            & 88.9$\pm$0.4        & \bf{90.1$\pm$0.4}	 & 86.5$\pm$0.5	        & 88.5$\pm$0.4	       & 89.4$\pm$0.3          & \bf{80.1$\pm$0.9}    & 77.1$\pm$0.9    & 75.6$\pm$0.9    & 73.3$\pm$0.8    & 72.5$\pm$0.9\\
					\rowcolor{red!7}CIFAR-10         & \bf{66.1$\pm$0.7}   & 50.3$\pm$1.0	     & 61.4$\pm$0.7     	& 59.6$\pm$0.7	       & 58.8$\pm$0.7          & \bf{50.3$\pm$0.9}    & 35.8$\pm$0.8    & 47.3$\pm$0.9    & 48.6$\pm$0.8    & 47.3$\pm$0.8\\
					\rowcolor{red!7}CIFAR-100        & 53.8$\pm$0.9        & 46.4$\pm$0.9	     & 52.5$\pm$0.9	        & \bf{55.8$\pm$0.9}	   & 55.3$\pm$0.9     & 53.8$\pm$0.9    & 42.9$\pm$1.0    & 54.9$\pm$1.1    & \bf{61.5$\pm$1.0}    & 60.2$\pm$1.0\\
					\midrule
					Average Seen                     & 69.0 	   & 71.2	     & 73.8	        & 75.7	       & \bf{76.0}	   & 65.0	 & 64.0    & 70.6    & \bf{72.5}    & 72.2\\
					Average Unseen                   & 62.6        & 56.0	     & 59.6	        & 62.3	       & \bf{63.0}	   & 57.7	 & 49.6    & 57.5    & \bf{58.4}    & 57.5\\
					Average All                      & 66.5        & 65.4	     & 68.3	        & 70.6		   & \bf{71.0}	   & 62.2	 & 58.5    & 65.5    & \bf{67.1}    & 66.5\\ 
					\midrule
					Average Rank                     & 4.1	       & 3.8	     & 2.8	        & 2.2	       & \bf{1.8}      & 3.6     & 4.2     & 2.5     & \bf{1.7}     & 2.8\\
					\bottomrule
				\end{tabular}
				\begin{tablenotes}
					\item[1] Both the results on URL and MOKD are the average of 5 random seed. The ranks only consider the first 10 datasets.
				\end{tablenotes} 
			\end{threeparttable}
		\end{tiny}
	\end{center}
	\vspace{-1.5em}
\end{table}

\begin{table}[t]
	\caption{\textbf{Analyses on $\gamma$ (Trained on All Datasets).} Mean accuracy, 95\% confidence interval are reported.}
	\label{Table:analyses_on_gamma}
	\begin{center}
		\begin{small}
			\begin{threeparttable}
				\begin{tabular}{lccccccc}
					\toprule
					Datasets & $\gamma=0.0$  & $\gamma=0.5$ & $\gamma=1.0$ & $\gamma=2.0$ & $\gamma=3.0$ & $\gamma=4.0$ & $\gamma=5.0$ \\
					\midrule
					\rowcolor{green!7}ImageNet    	 & 53.6$\pm$1.0    & 56.2$\pm$1.1    & 57.0$\pm$1.1    & 57.2$\pm$1.1    & \bf{57.3$\pm$1.1}    & \bf{57.3$\pm$1.1} & \bf{57.3$\pm$1.1}\\
					\rowcolor{green!7}Omniglot     	 & \bf{94.4$\pm$0.5}    & \bf{94.4$\pm$0.5}    & 94.2$\pm$0.5    & 93.9$\pm$0.5    & 93.7$\pm$0.5    & 93.5$\pm$0.5    & 93.3$\pm$0.5\\
					\rowcolor{green!7}Aircraft       & 85.2$\pm$0.5    & 87.6$\pm$0.5    & \bf{88.4$\pm$0.5}    & 88.3$\pm$0.5    & 88.2$\pm$0.5    & 88.0$\pm$0.5    & 87.9$\pm$0.5\\
					\rowcolor{green!7}Birds    		 & 77.8$\pm$0.7    & 80.3$\pm$0.7    & \bf{80.4$\pm$0.8}    & 80.3$\pm$0.8    & 80.1$\pm$0.8    & 79.9$\pm$0.8    & 79.8$\pm$0.8\\
					\rowcolor{green!7}Textures     	 & 73.2$\pm$0.7    & 75.4$\pm$0.7    & 76.1$\pm$0.7    & 76.3$\pm$0.7    & \bf{76.5$\pm$0.7}    & \bf{76.5$\pm$0.7}    & \bf{76.5$\pm$0.7}\\
					\rowcolor{green!7}Quick Draw     & 80.5$\pm$0.6    & 82.1$\pm$0.6    & \bf{82.3$\pm$0.6}    & \bf{82.3$\pm$0.6}    & 82.2$\pm$0.6    & 82.1$\pm$0.6    & 82.0$\pm$0.6\\
					\rowcolor{green!7}Fungi      	 & 61.9$\pm$0.9    & 65.1$\pm$1.0    & 66.8$\pm$1.0    & 68.1$\pm$1.0    & 68.6$\pm$1.0    & \bf{68.7$\pm$1.0}    & \bf{68.7$\pm$1.0}\\
					\rowcolor{green!7}VGG Flower     & 88.8$\pm$0.5    & 91.5$\pm$0.5    & 92.1$\pm$0.5    & 92.4$\pm$0.5    & \bf{92.5$\pm$0.5}    & \bf{92.5$\pm$0.5}    & 92.3$\pm$0.5\\
					\midrule
					\rowcolor{red!7}Traffic Sign     & 48.7$\pm$1.0    & 62.9$\pm$1.1    & 64.1$\pm$1.1    & \bf{64.6$\pm$1.1}    & 64.5$\pm$1.1    & 64.2$\pm$1.1    & 64.0$\pm$1.1\\
					\rowcolor{red!7}MSCOCO           & 44.7$\pm$1.0    & 51.3$\pm$1.0    & 53.4$\pm$1.0    & 55.0$\pm$1.0    & \bf{55.5$\pm$1.0}    & 55.4$\pm$1.0    & 55.3$\pm$1.0\\
					\rowcolor{red!7}MNIST            & 91.7$\pm$0.5    & 95.0$\pm$0.4    & \bf{95.1$\pm$0.4}    & 94.6$\pm$0.4    & 94.5$\pm$0.4    & 94.3$\pm$0.4    & 94.3$\pm$0.4\\
					\rowcolor{red!7}CIFAR-10         & 66.2$\pm$0.8    & 71.0$\pm$0.8    & 72.3$\pm$0.8    & 72.4$\pm$0.8    & 72.8$\pm$0.8    & \bf{72.9$\pm$0.8}    & \bf{72.9$\pm$0.8}\\
					\rowcolor{red!7}CIFAR-100        & 57.1$\pm$1.0    & 62.5$\pm$1.0    & 63.5$\pm$1.0    & \bf{64.0$\pm$1.0}    & 63.9$\pm$1.0    & 63.8$\pm$1.0    & 63.6$\pm$1.0\\
					\midrule
					Average Seen                     & 76.9    & 79.1   & 79.7    & 79.8    & \bf{79.9}     & 79.8     & 79.7\\
					Average Unseen                   & 61.7    & 68.4   & 69.7    & 70.1    & \bf{70.2}     & 70.1     & 70.0\\
					Average All                      & 71.1    & 75.0   & 75.8    & 76.1    & \bf{76.2}     & 76.1     & 76.0\\ 
					\bottomrule
				\end{tabular}
			\end{threeparttable}
		\end{small}
	\end{center}
	\vskip -0.1in
\end{table}

\begin{table}[t]
	\vspace{-1.0em}
	\caption{\textbf{Comparisons of running time between MOKD and URL}. (sec. per task)}
	\label{Table:analyses_on_runtime}
	\begin{center}
		\begin{tiny}
		\setlength\tabcolsep{3pt}
			\begin{threeparttable}
				\begin{tabular}{lccccccccccccc}
					\toprule
					Datasets & ImageNet & Omniglot & Aircraft & Birds & DTD & QuickDraw & Fungi & VGG\_Flower & Traffic Sign & MSCOCO & MNIST & CIFAR10 & CIFAR100 \\
					\midrule
					URL    	 & 0.7 & 0.8 & 0.5 & 0.7 & 0.4 & 1.1 & 1.0 & 0.5 & 1.0 & 0.9 & 0.5 & 0.5 & 1.1\\
					MOKD     & 0.9 & 0.7 & 0.8 & 0.8 & 0.8 & 0.9 & 0.8 & 0.8 & 0.9 & 0.9 & 0.8 & 0.8 & 0.9\\
					\bottomrule
				\end{tabular}
			\end{threeparttable}
		\end{tiny}
	\end{center}
	\vskip -0.1in
	\vspace{-0.5em}
\end{table}

\begin{table}[t]
	\caption{\textbf{Ablation study on test power maximization.} Mean accuracy, 95\% confidence interval are reported.}
	\label{Table:ablation_on_testpower}
	\begin{center}
		\begin{small}
			\begin{threeparttable}
				\begin{tabular}{lcc}
					\toprule
					Datasets  & MOKD w/o Test Power & MOKD w Test Power \\
					\midrule
					\rowcolor{green!7}ImageNet    	 & 56.4$\pm$1.1         & \bf{57.2$\pm$1.1}\\
					\rowcolor{green!7}Omniglot     	 & 93.9$\pm$0.5         & \bf{94.2$\pm$0.5}\\
					\rowcolor{green!7}Aircraft       & 88.1$\pm$0.5         & \bf{88.3$\pm$0.5}\\
					\rowcolor{green!7}Birds    		 & 80.1$\pm$0.8         & \bf{80.2$\pm$0.8}\\
					\rowcolor{green!7}Textures     	 & \bf{75.9$\pm$0.7}    & 75.7$\pm$0.7\\
					\rowcolor{green!7}Quick Draw     & 82.0$\pm$0.6         & \bf{82.1$\pm$0.6}\\
					\rowcolor{green!7}Fungi      	 & 64.1$\pm$1.1         & \bf{68.9$\pm$1.0}\\
					\rowcolor{green!7}VGG Flower     & \bf{91.9$\pm$0.5}    & 91.8$\pm$0.5\\
					\midrule
					\rowcolor{red!7}Traffic Sign     & 62.8$\pm$1.2         & \bf{64.1$\pm$1.0}\\
					\rowcolor{red!7}MSCOCO           & 52.9$\pm$1.1         & \bf{55.3$\pm$1.0}\\
					\rowcolor{red!7}MNIST            & \bf{95.3$\pm$0.4}    & 95.0$\pm$0.4\\
					\rowcolor{red!7}CIFAR-10         & 72.1$\pm$0.8         & \bf{73.0$\pm$0.8}\\
					\rowcolor{red!7}CIFAR-100        & 62.0$\pm$1.0         & \bf{63.0$\pm$1.0}\\
					\midrule
					Average Seen                     & 79.0                 & \bf{79.8}\\
					Average Unseen                   & 69.0                 & \bf{70.1}\\
					Average All                      & 75.2                 & \bf{76.1}\\ 

					\bottomrule
				\end{tabular}
			\end{threeparttable}
		\end{small}
	\end{center}
	\vskip -0.1in
\end{table}

\subsection{Analyses on Gamma}\label{section:gamma_analysis}

\begin{figure}[ht]
	\vskip 0.0in
	\begin{center}
	\centering
	\subfigure[ImageNet\label{Fig:ablcurve_ilsvrc_2012}]{
			\begin{minipage}[t]{0.32\linewidth}
				\centering
				\includegraphics[width=1.0\linewidth]{./figs/ab_gamma_test_curves_ilsvrc_2012.pdf}
		\end{minipage}}\vspace{-0.1cm}
  \subfigure[Omniglot\label{Fig:ablcurve_omniglot}]{
			\begin{minipage}[t]{0.32\linewidth}
				\centering
				\includegraphics[width=1.0\linewidth]{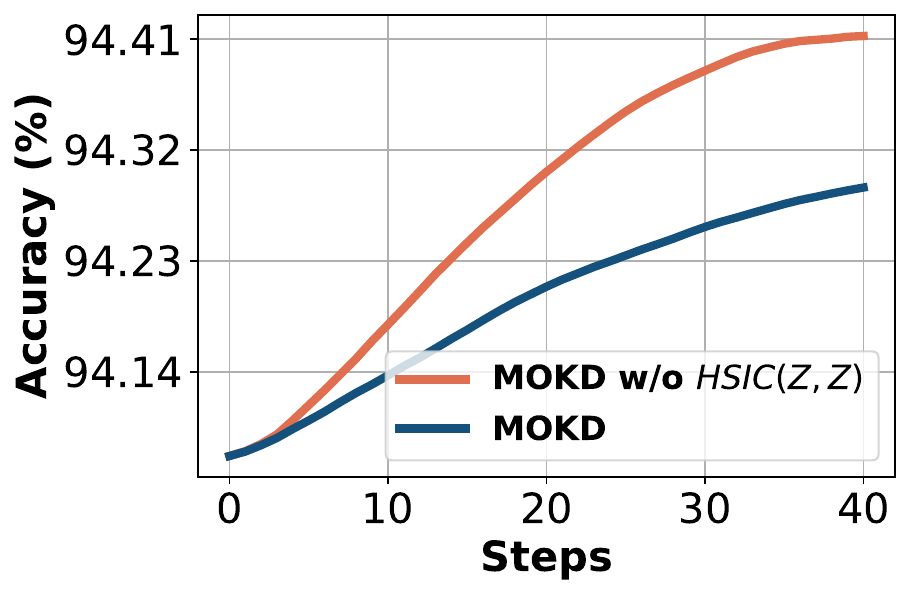}
		\end{minipage}}\vspace{-0.1cm}
  \subfigure[Aircraft\label{Fig:ablcurve_aircraft}]{
			\begin{minipage}[t]{0.32\linewidth}
				\centering
				\includegraphics[width=1.0\linewidth]{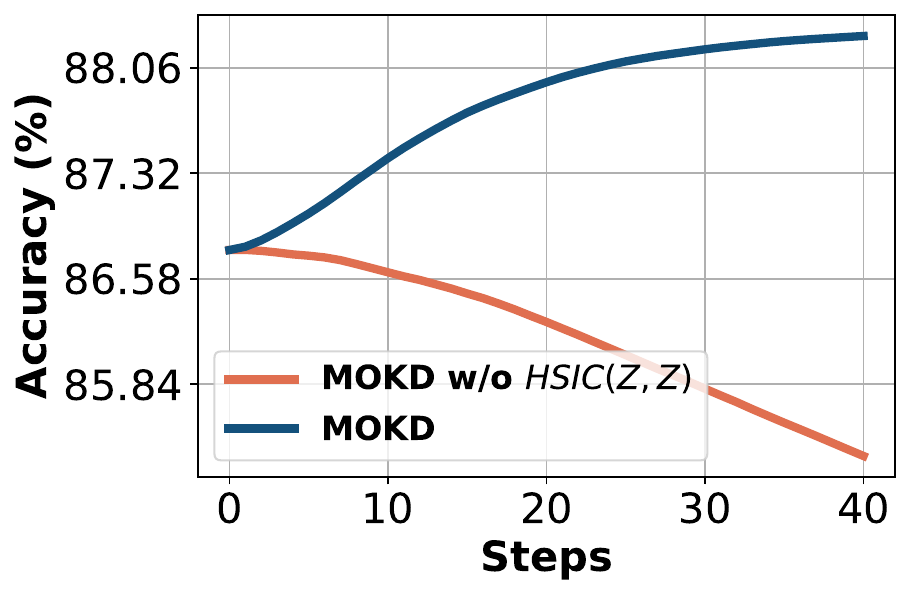}
		\end{minipage}}\vspace{-0.1cm}
  \subfigure[CU\_Birds\label{Fig:ablcurve_cu_birds}]{
			\begin{minipage}[t]{0.32\linewidth}
				\centering
				\includegraphics[width=1.0\linewidth]{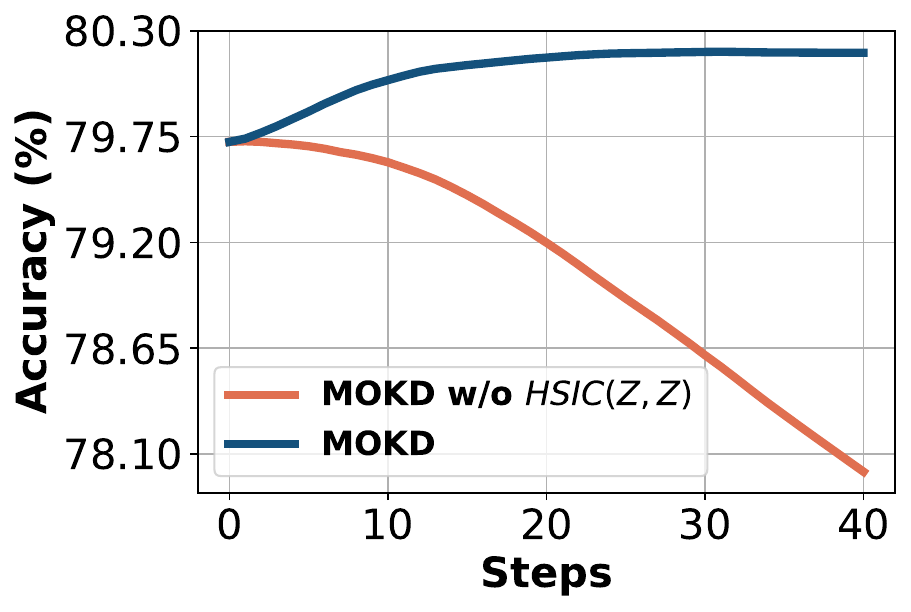}
		\end{minipage}}\vspace{-0.1cm}
  \subfigure[DTD\label{Fig:ablcurve_dtd}]{
			\begin{minipage}[t]{0.32\linewidth}
				\centering
				\includegraphics[width=1.0\linewidth]{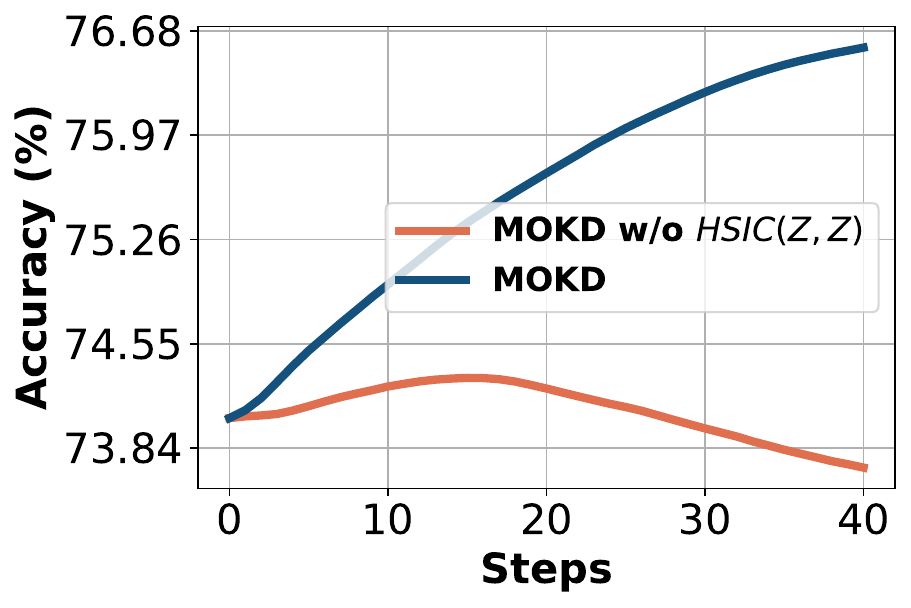}
		\end{minipage}}\vspace{-0.1cm}
  \subfigure[Quick Draw\label{Fig:ablcurve_quickdraw}]{
			\begin{minipage}[t]{0.32\linewidth}
				\centering
				\includegraphics[width=1.0\linewidth]{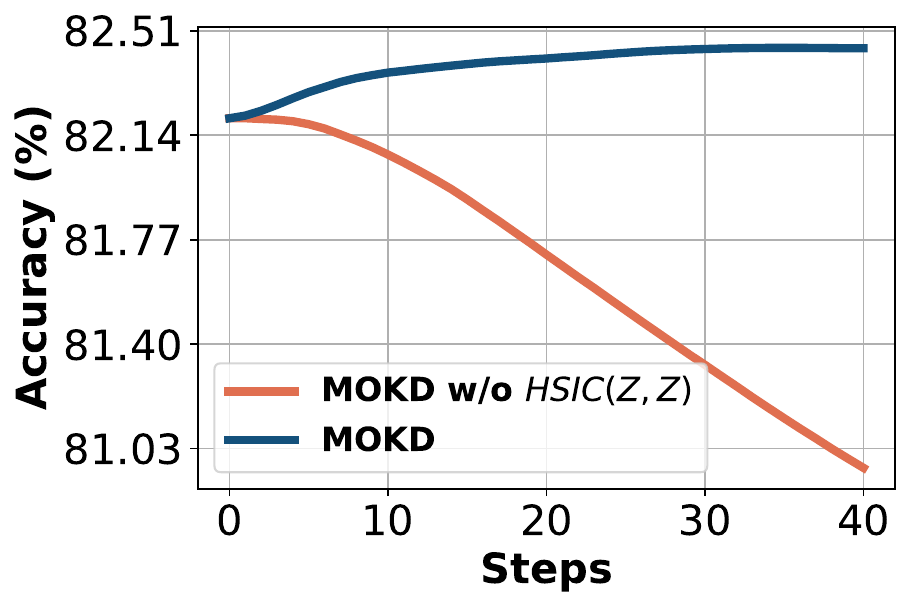}
		\end{minipage}}\vspace{-0.1cm}
	\subfigure[Fungi\label{Fig:ablcurve_fungi}]{
			\begin{minipage}[t]{0.32\linewidth}
				\centering
				\includegraphics[width=1.0\linewidth]{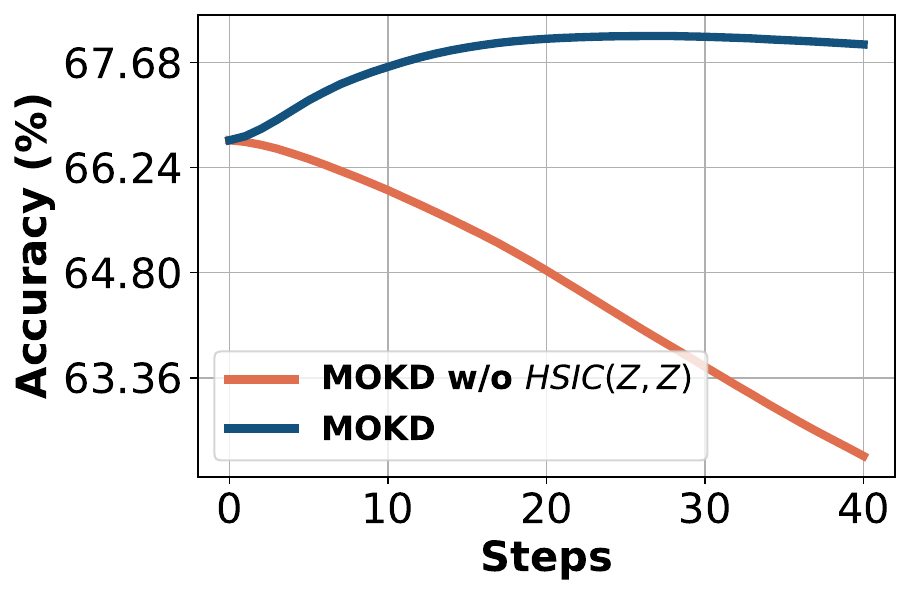}
		\end{minipage}}\vspace{-0.1cm}
  \subfigure[VGG\_Flower\label{Fig:ablcurve_vgg_flower}]{
			\begin{minipage}[t]{0.32\linewidth}
				\centering
				\includegraphics[width=1.0\linewidth]{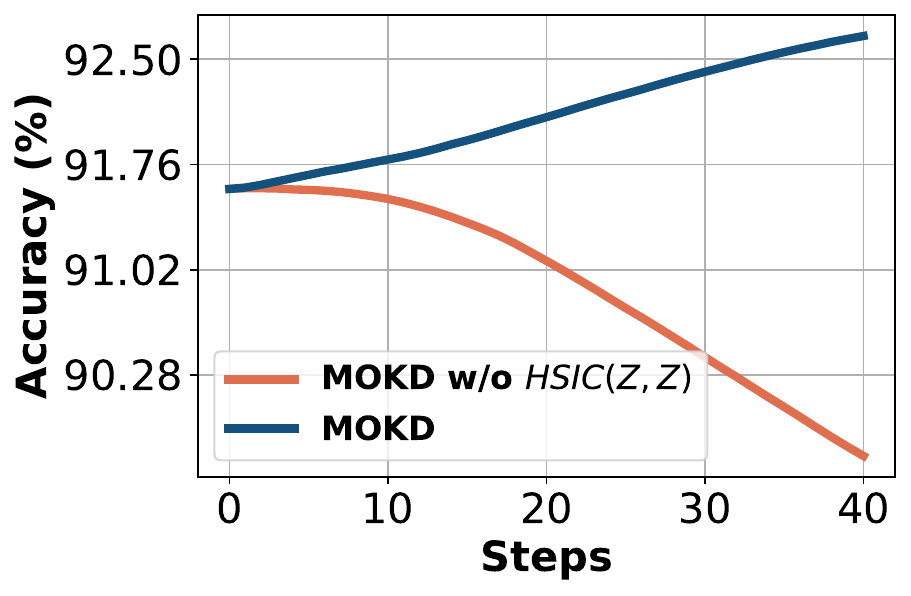}
		\end{minipage}}\vspace{-0.1cm}
	\subfigure[Traffic Sign\label{Fig:ablcurve_traffic}]{
			\begin{minipage}[t]{0.32\linewidth}
				\centering
				\includegraphics[width=1.0\linewidth]{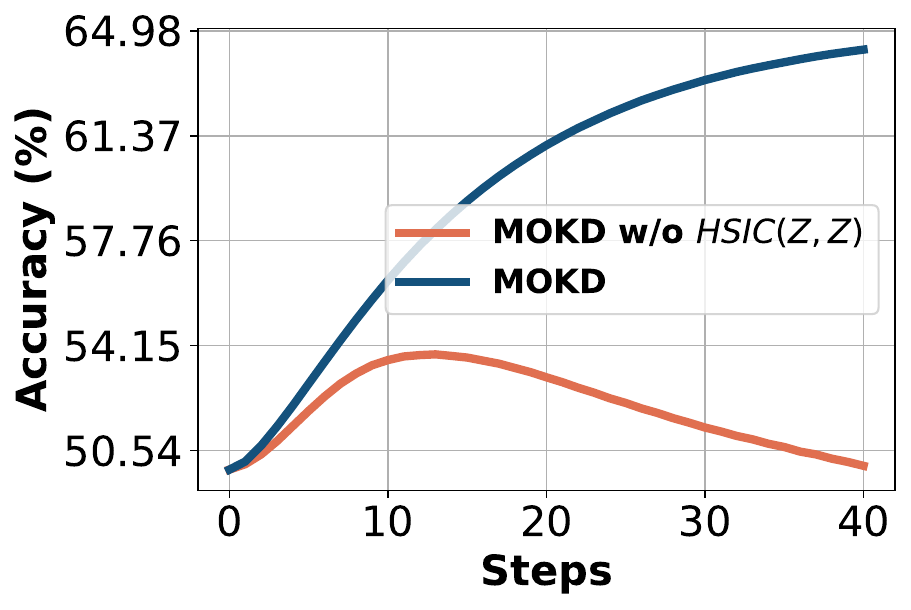}
		\end{minipage}}\vspace{-0.1cm}
	\subfigure[MOCOCO\label{Fig:ablcurve_mscoco}]{
			\begin{minipage}[t]{0.32\linewidth}
				\centering
				\includegraphics[width=1.0\linewidth]{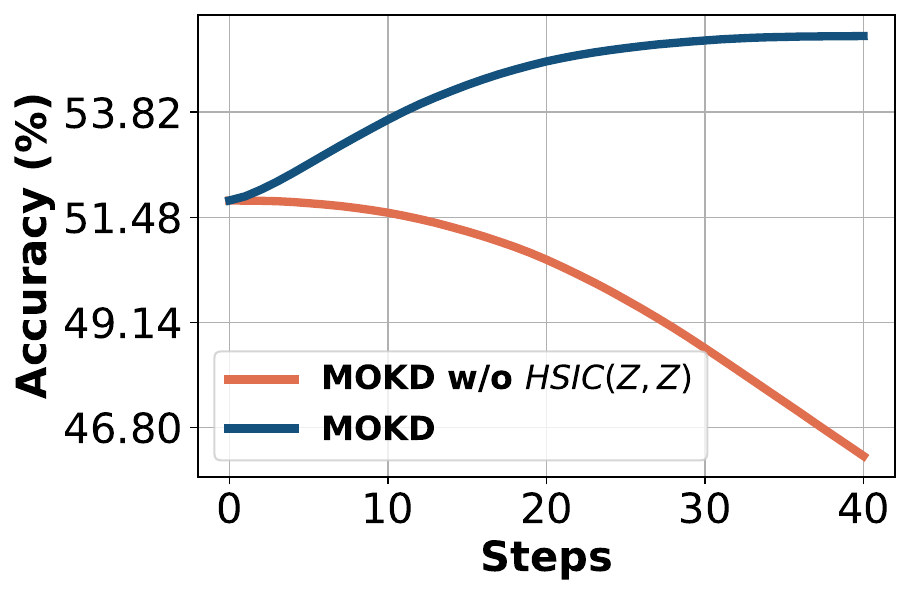}
		\end{minipage}}\vspace{-0.1cm}
	\subfigure[MNIST\label{Fig:ablcurve_mnist}]{
			\begin{minipage}[t]{0.32\linewidth}
				\centering
				\includegraphics[width=1.0\linewidth]{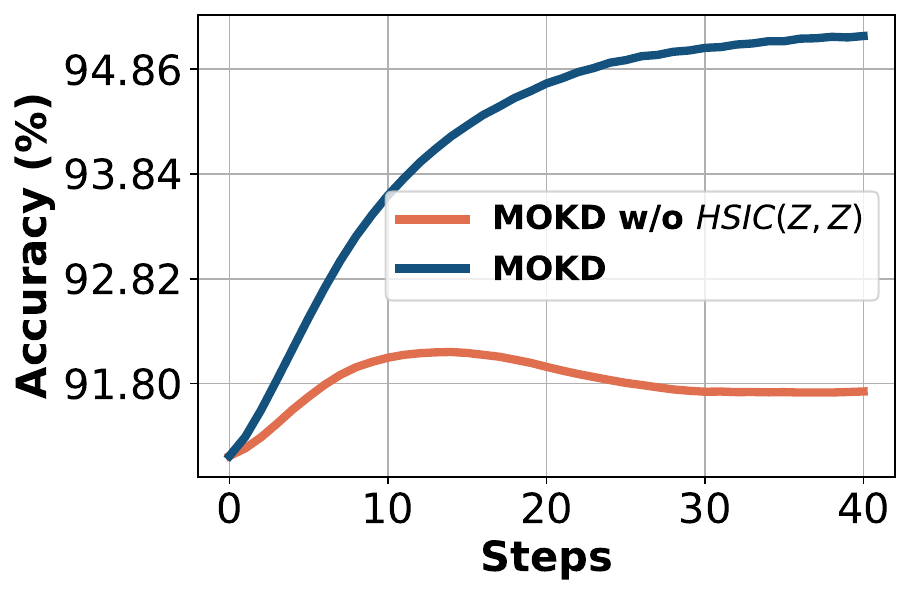}
		\end{minipage}}\vspace{-0.1cm}
    \subfigure[CIFAR 10\label{Fig:ablcurve_cifar10}]{
			\begin{minipage}[t]{0.32\linewidth}
				\centering
				\includegraphics[width=1.0\linewidth]{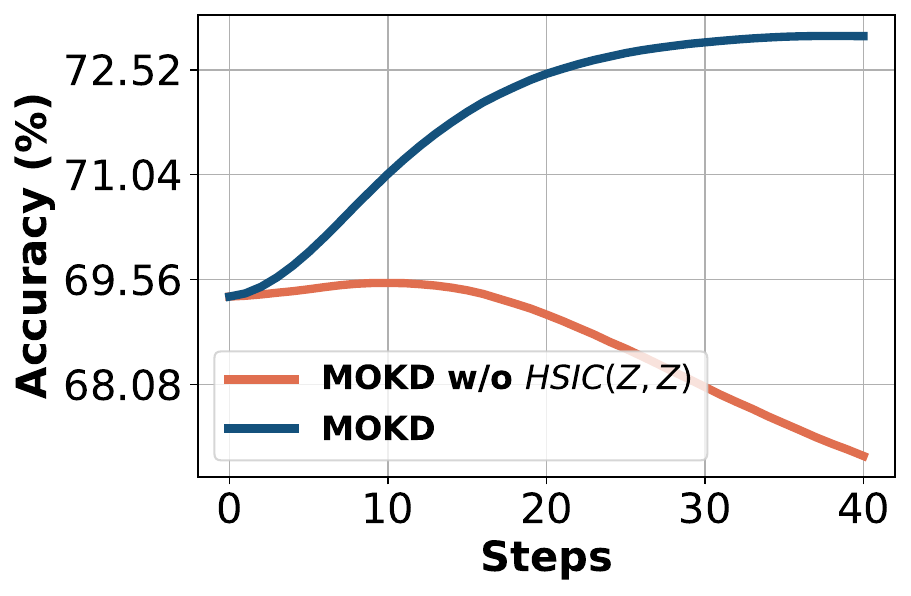}
		\end{minipage}}\vspace{-0.1cm}
  \subfigure[CIFAR 100\label{Fig:ablcurve_cifar100}]{
			\begin{minipage}[t]{0.32\linewidth}
				\centering
				\includegraphics[width=1.0\linewidth]{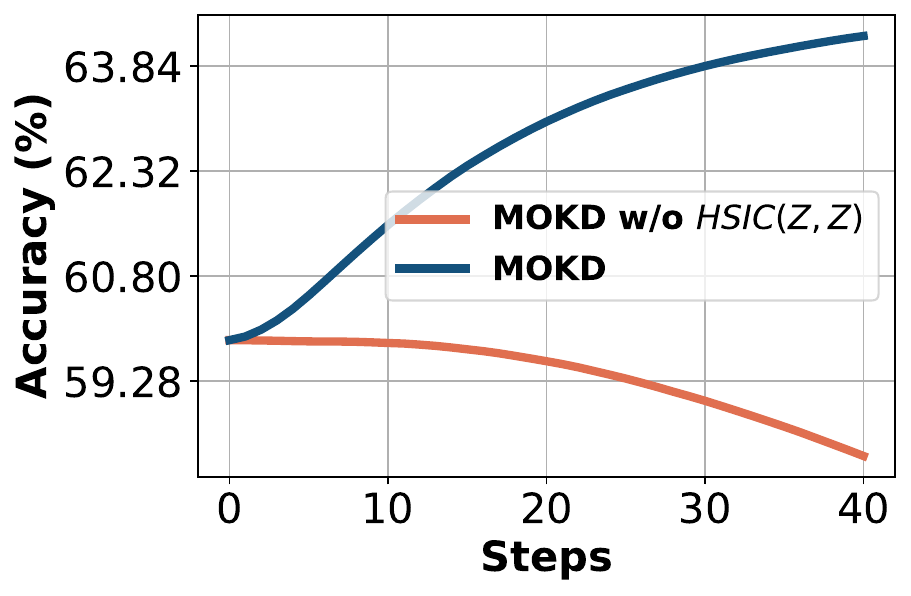}
		\end{minipage}}\vspace{-0.1cm}
    \caption{\textbf{Part test accuracy curves of datasets in Meta-Dataset.} The learning curves show that when ${\rm HSIC}(Z, Z)$ is removed, MOKD tends to overfit the training data and achieves bad generalization performance.}
    \label{Fig:ablearn_curves}
	\end{center}
	\vskip -0.2in
	\vspace{1em}
\end{figure}

\begin{figure}[t]
	\vskip 0.0in
	\begin{center}
	\centering
	\subfigure[ImageNet\label{Fig:appendix_imagenet_gamma}]{
			\begin{minipage}[t]{0.32\linewidth}
				\centering
				\includegraphics[width=1.0\linewidth]{./figs/imagenet_gamma.pdf}
		\end{minipage}}\vspace{-0.1cm}
	\subfigure[Omniglot\label{Fig:appendix_omniglot_gamma}]{
			\begin{minipage}[t]{0.32\linewidth}
				\centering
				\includegraphics[width=1.0\linewidth]{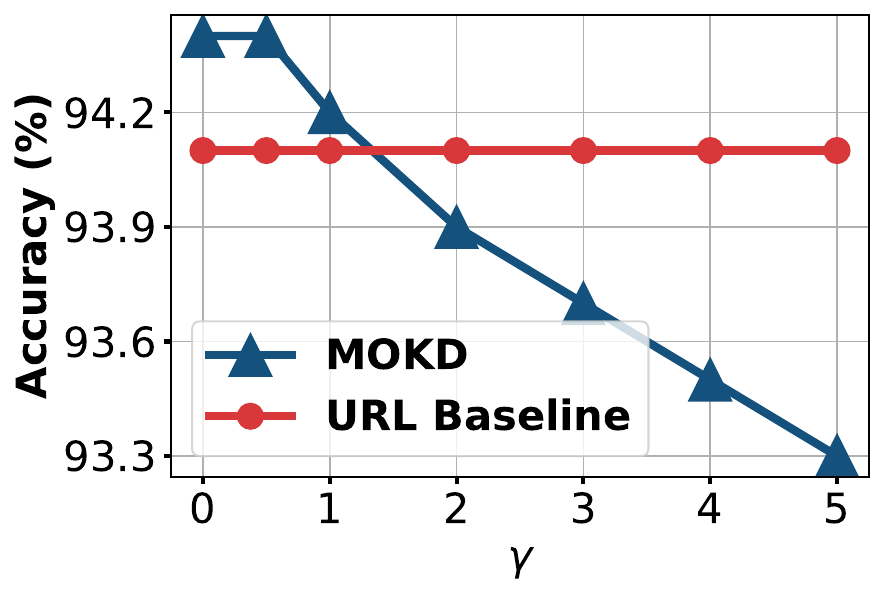}
		\end{minipage}}\vspace{-0.1cm}
	\subfigure[Aircraft\label{Fig:appendix_aircraft_gamma}]{
			\begin{minipage}[t]{0.32\linewidth}
				\centering
				\includegraphics[width=1.0\linewidth]{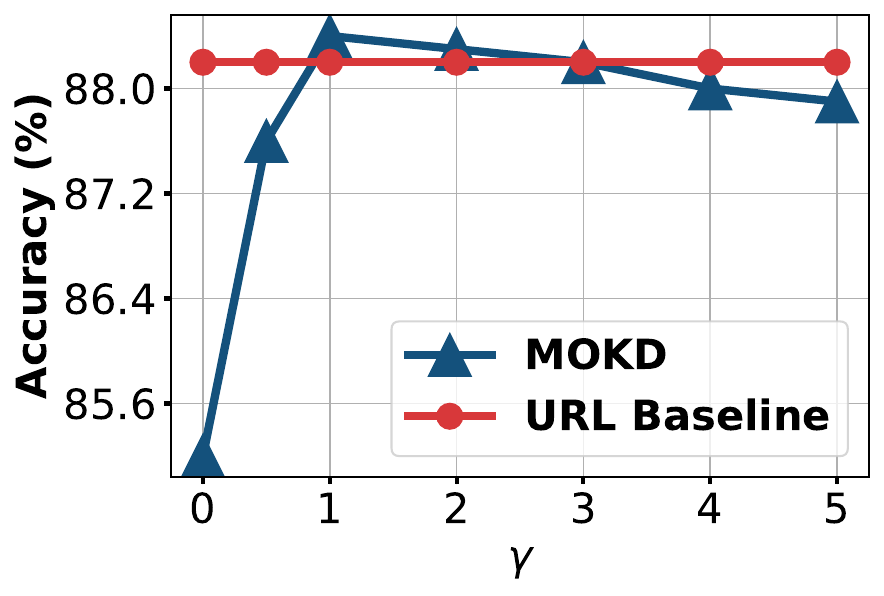}
		\end{minipage}}\vspace{-0.1cm}
	\subfigure[CU\_Birds\label{Fig:appendix_birds_gamma}]{
			\begin{minipage}[t]{0.32\linewidth}
				\centering
				\includegraphics[width=1.0\linewidth]{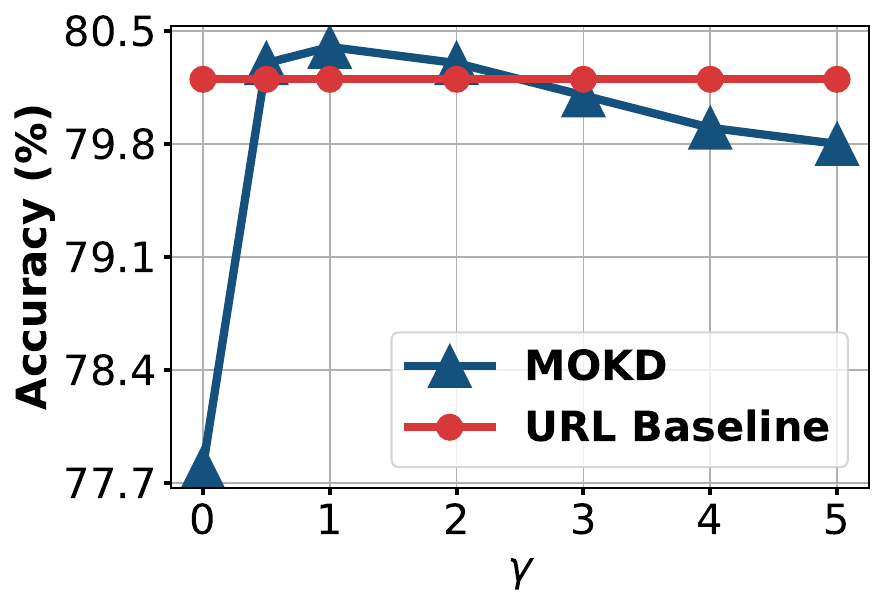}
		\end{minipage}}\vspace{-0.1cm}
	\subfigure[DTD\label{Fig:appendix_dtd_gamma}]{
			\begin{minipage}[t]{0.32\linewidth}
				\centering
				\includegraphics[width=1.0\linewidth]{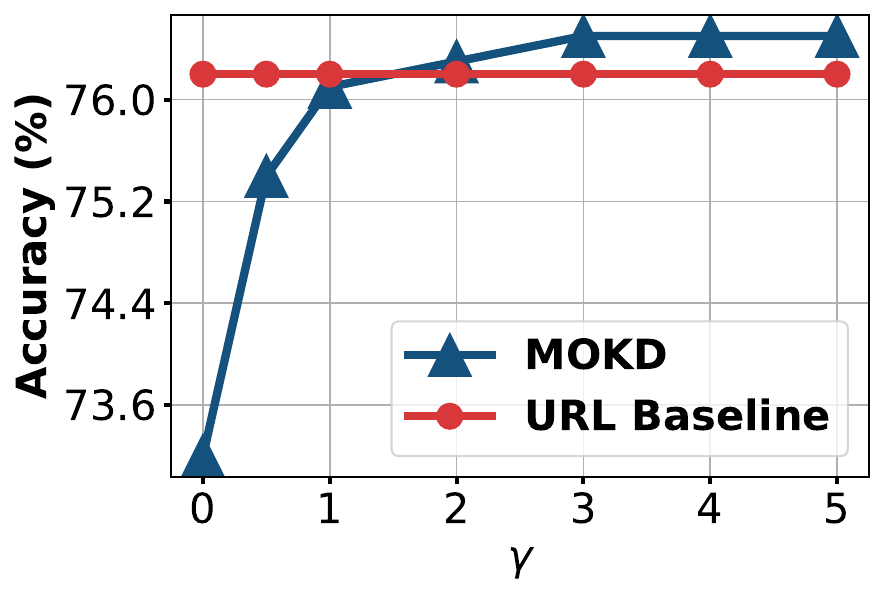}
		\end{minipage}}\vspace{-0.1cm}
	\subfigure[Quick Draw\label{Fig:appendix_quickdraw_gamma}]{
			\begin{minipage}[t]{0.32\linewidth}
				\centering
				\includegraphics[width=1.0\linewidth]{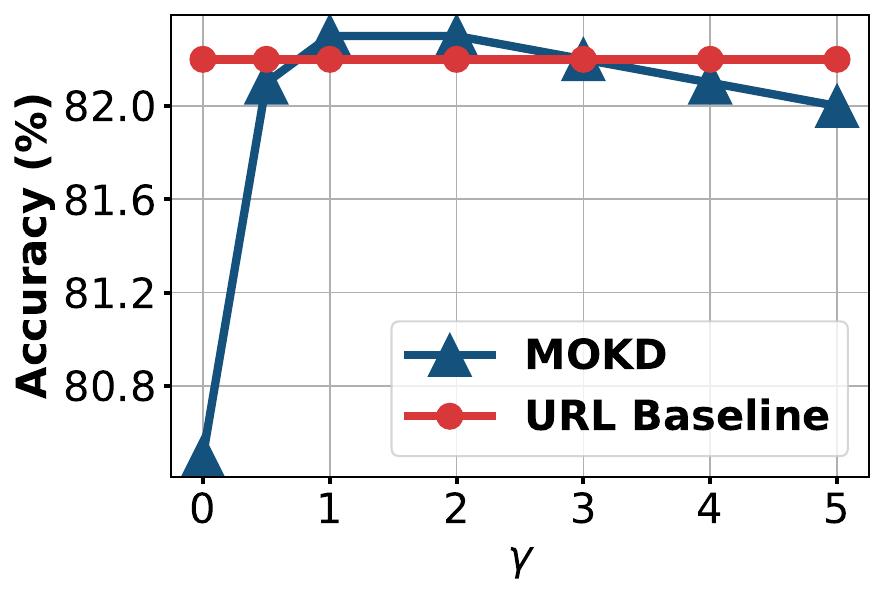}
		\end{minipage}}\vspace{-0.1cm}
	\subfigure[Fungi\label{Fig:appendix_fungi_gamma}]{
			\begin{minipage}[t]{0.32\linewidth}
				\centering
				\includegraphics[width=1.0\linewidth]{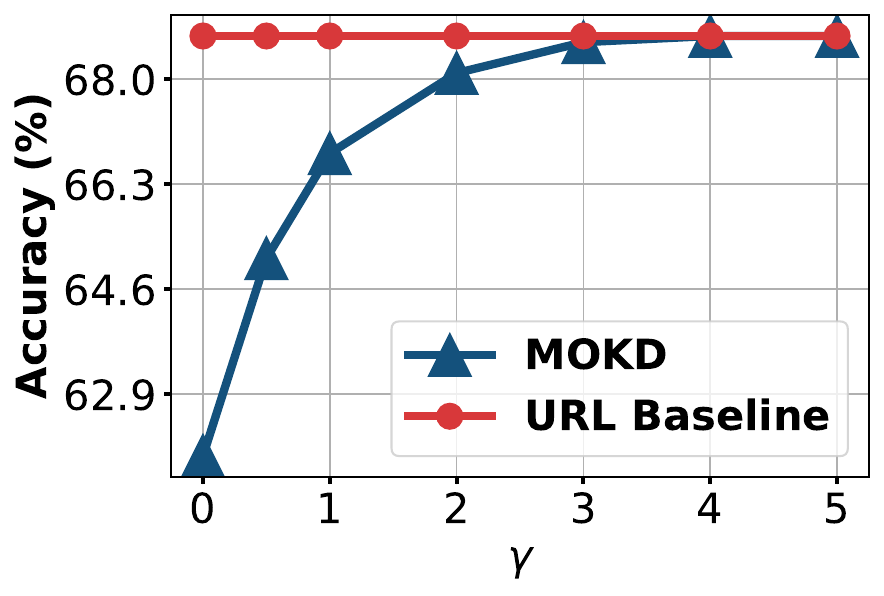}
		\end{minipage}}\vspace{-0.1cm}
	\subfigure[VGG Flower\label{Fig:appendix_gamma}]{
			\begin{minipage}[t]{0.32\linewidth}
				\centering
				\includegraphics[width=1.0\linewidth]{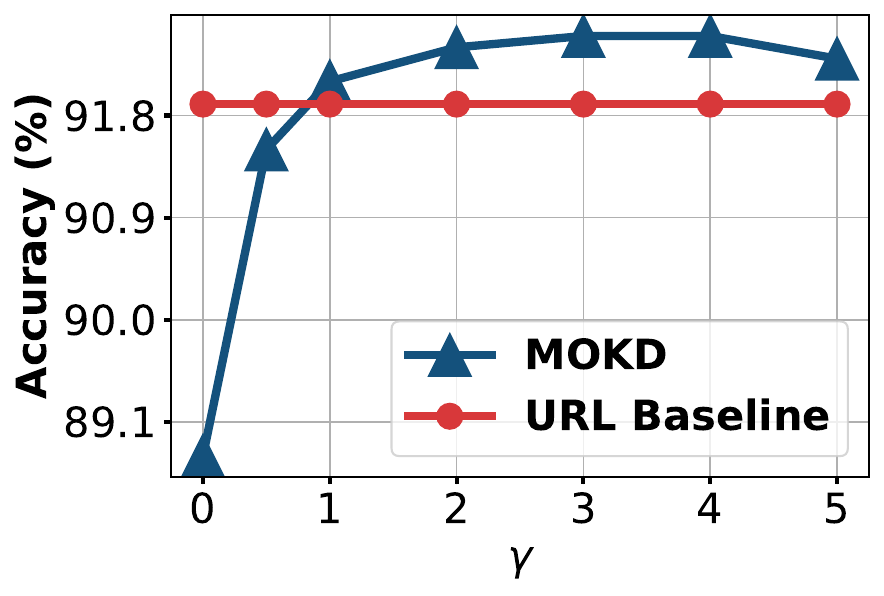}
		\end{minipage}}\vspace{-0.1cm}
	\subfigure[Traffic Sign\label{Fig:appendix_traffic_gamma}]{
			\begin{minipage}[t]{0.32\linewidth}
				\centering
				\includegraphics[width=1.0\linewidth]{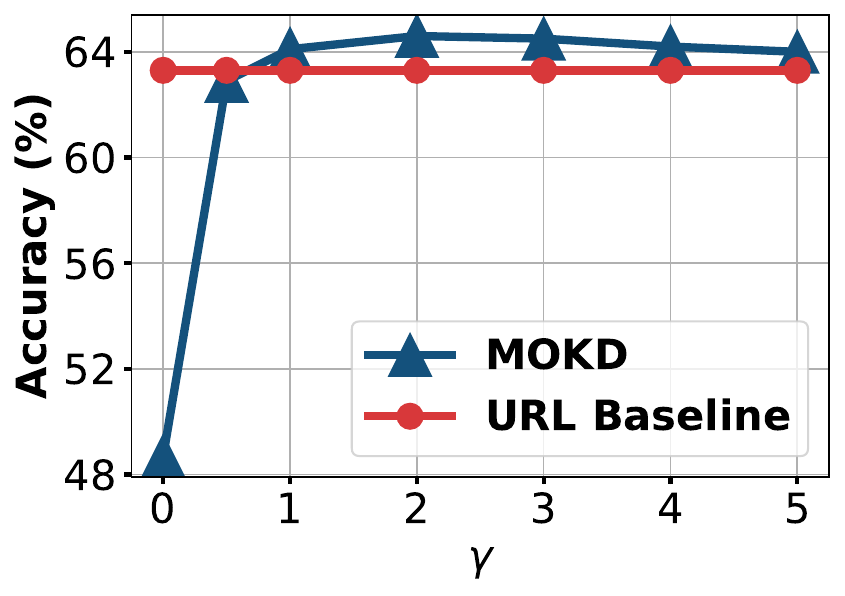}
		\end{minipage}}\vspace{-0.1cm}
	\subfigure[MSCOCO\label{Fig:appendix_mscoco_gamma}]{
			\begin{minipage}[t]{0.32\linewidth}
				\centering
				\includegraphics[width=1.0\linewidth]{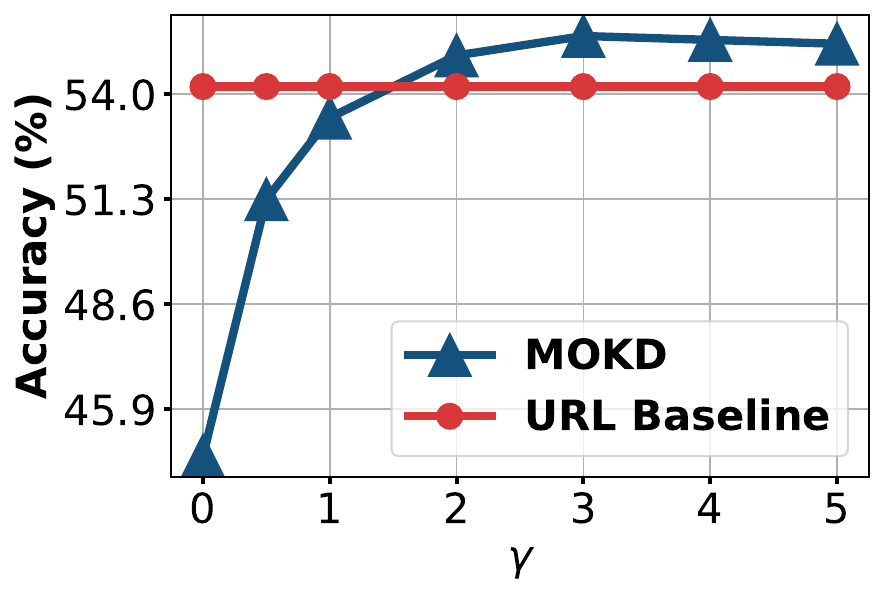}
		\end{minipage}}\vspace{-0.1cm}
	\subfigure[MNIST\label{Fig:appendix_mnist_gamma}]{
			\begin{minipage}[t]{0.32\linewidth}
				\centering
				\includegraphics[width=1.0\linewidth]{./figs/mnist_gamma.pdf}
		\end{minipage}}\vspace{-0.1cm}
	\subfigure[CIFAR 10\label{Fig:appendix_cifar10_gamma}]{
			\begin{minipage}[t]{0.32\linewidth}
				\centering
				\includegraphics[width=1.0\linewidth]{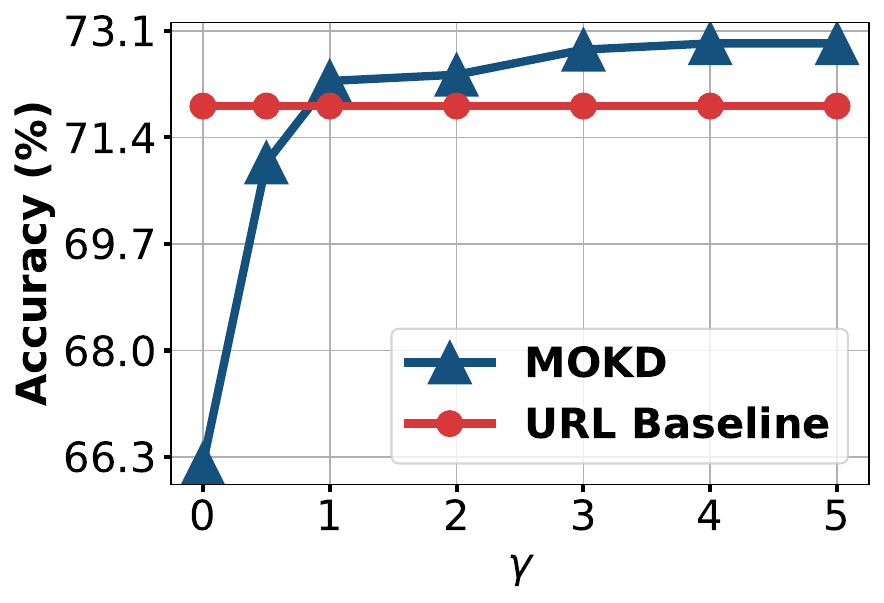}
		\end{minipage}}\vspace{-0.1cm}
	\subfigure[CIFAR 100\label{Fig:appendix_cifar100_gamma}]{
			\begin{minipage}[t]{0.32\linewidth}
				\centering
				\includegraphics[width=1.0\linewidth]{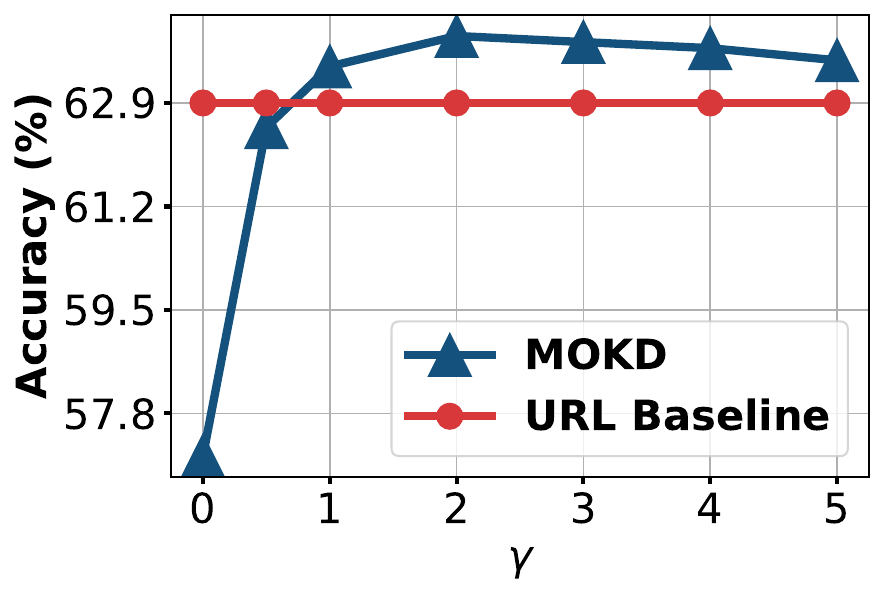}
		\end{minipage}}\vspace{-0.1cm}
    \caption{Illustration of effect of $\gamma$ on all datasets in Meta-Dataset.}
    \label{Fig:complete_effect_gamma}
	\end{center}
	\vskip -0.2in
	\vspace{1em}
\end{figure}

In our work, $\gamma$ functions as a coefficient of regularization term ${\rm HSIC}(Z, Z)$. Since ${\rm HSIC}(Z, Z)$ mainly facilitates penalizing high-variance representations and removing common information shared among samples, $\gamma$ intuitively determines the power of penalization and suppression imposed to high-variance representations and common features shared across samples. In order to figure out how datasets in Meta-Dataset react to $\gamma$, we run MOKD with different $\gamma$ values under the ``train on all datasets'' settings. The results are reported in Table~\ref{Table:analyses_on_gamma} and Fig. \ref{Fig:ablearn_curves} and \ref{Fig:complete_effect_gamma}.

According to the numerical results reported in Table~\ref{Table:analyses_on_gamma}, a general conclusion we can summarize is that different datasets prefer different values of $\gamma$. To be concrete, for simple datasets, such as Omniglot, Aircraft and MNIST, small $\gamma$ is preferred since images in these datasets are simple and the main object of each image is evident. However, for complicated datasets, such as ImageNet, Fungi, and MSCOCO, large $\gamma$ is better since images in these datasets contain abundant objects and semantic information. In some cases, the semantic information, such as the backgrounds, is useless and sometimes may have a negative effect on the performance.

In addition, a special case is that it is equivalent to performing an ablation study on ${\rm HSIC}(Z, Z)$ when $\gamma$ is set to zero. As shown in the table, the performance drops drastically on most datasets. By further plotting the learning curves (see Fig.~\ref{Fig:ablearn_curves}), we find that overfitting happens on these datasets when the ${\rm HSIC}(Z, Z)$ term is removed. Thus, such phenomenon demonstrates that ${\rm HSIC}(Z, Z)$ contributes to penalizing high-variance kernelized representations and alleviating the overfitting phenomenon.

\paragraph{Further discussion about $\gamma$.} According to the results reported in Table~\ref{Table:analyses_on_gamma}, an interesting phenomenon is that different datasets achieve their best performance with different gamma values. In our opinion, two aspects are worth noticing here.

On the one hand, as aforementioned, there exist high similarities between samples from different classes when performing classification with NCC-based loss. A reasonable cause for such a phenomenon is the trivial common features shared across samples. On the other side, according to Fig. \ref{Fig:ablearn_curves}, merely maximizing ${\rm HSIC}(Z, Y)$ results in the overfitting phenomenon.

\begin{figure}[t]
    \vspace{-0.5em}
	\vskip 0.0in
	\begin{center}
	\centering
    \subfigure[Omniglot (21 classes)\label{Fig:visual_omniglot_sim}]{
			\begin{minipage}[t]{0.485\linewidth}
				\centering
				\includegraphics[width=1.0\linewidth]{./figs/omniglot__21_heatmap.pdf}
		\end{minipage}}\vspace{-0.1cm}
    \subfigure[Aircraft (9 classes)\label{Fig:visual_aircraft_sim}]{
			\begin{minipage}[t]{0.485\linewidth}
				\centering
				\includegraphics[width=1.0\linewidth]{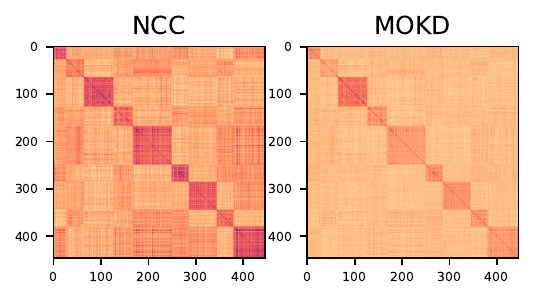}
		\end{minipage}}\vspace{-0.1cm}
    \subfigure[CU\_Birds (11 classes)\label{Fig:visual_birds_sim}]{
			\begin{minipage}[t]{0.485\linewidth}
				\centering
				\includegraphics[width=1.0\linewidth]{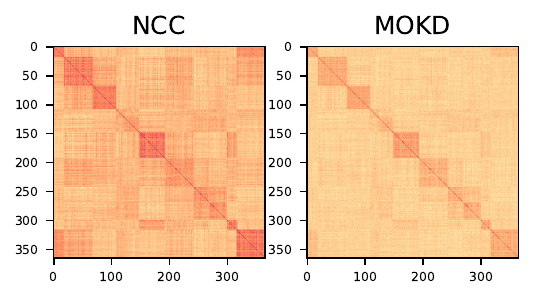}
		\end{minipage}}\vspace{-0.1cm}
    \subfigure[DTD (5 classes)\label{Fig:visual_dtd_sim}]{
			\begin{minipage}[t]{0.485\linewidth}
				\centering
				\includegraphics[width=1.0\linewidth]{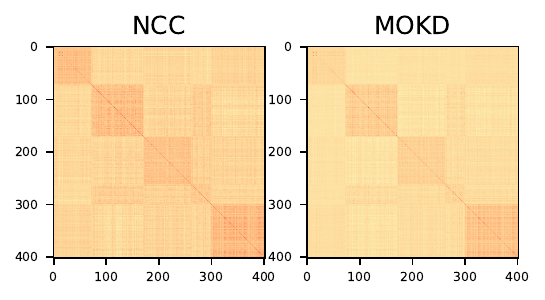}
		\end{minipage}}\vspace{-0.1cm}
    \subfigure[Quick Draw (12 classes)\label{Fig:visual_quickdraw_sim}]{
			\begin{minipage}[t]{0.485\linewidth}
				\centering
				\includegraphics[width=1.0\linewidth]{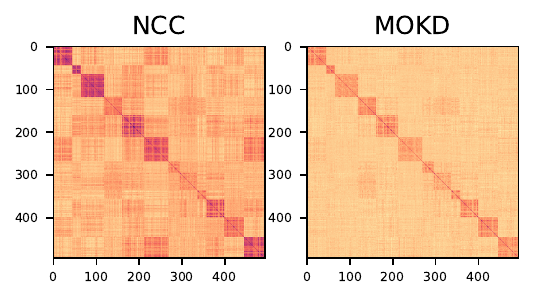}
		\end{minipage}}\vspace{-0.1cm}
    \subfigure[Fungi (33 classes)\label{Fig:visual_fungi_sim}]{
			\begin{minipage}[t]{0.485\linewidth}
				\centering
				\includegraphics[width=1.0\linewidth]{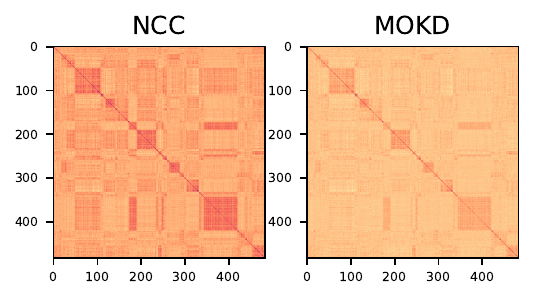}
		\end{minipage}}\vspace{-0.1cm}
    \subfigure[VGG\_Flower (16 classes)\label{Fig:visual_vgg_flower_sim}]{
			\begin{minipage}[t]{0.485\linewidth}
				\centering
				\includegraphics[width=1.0\linewidth]{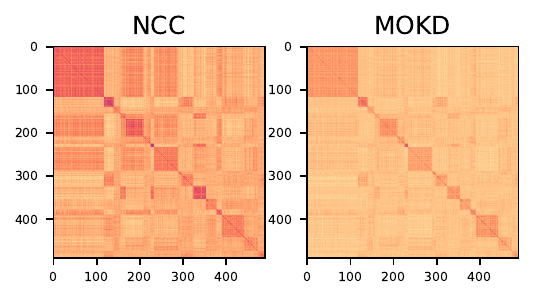}
		\end{minipage}}\vspace{-0.1cm}
    \caption{Visualization results of similarity matrices of representations respectively learned with NCC-based loss and MOKD on seen domains.}
    \label{Fig:remaining_visual_sim_seen}
	\end{center}
	\vskip -0.2in
\end{figure}

\begin{figure}[t]
    \vspace{-0.5em}
	\vskip 0.0in
	\begin{center}
	\centering
    \subfigure[Traffic Sign (12 classes)\label{Fig:visual_traffic_sign_sim}]{
			\begin{minipage}[t]{0.485\linewidth}
				\centering
				\includegraphics[width=1.0\linewidth]{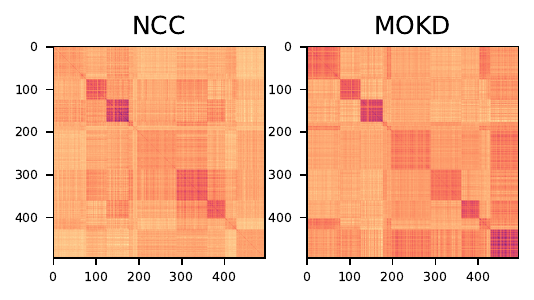}
		\end{minipage}}\vspace{-0.1cm}	
	\subfigure[MSCOCO (19 classes)\label{Fig:visual_mscoco_sim}]{
			\begin{minipage}[t]{0.485\linewidth}
				\centering
				\includegraphics[width=1.0\linewidth]{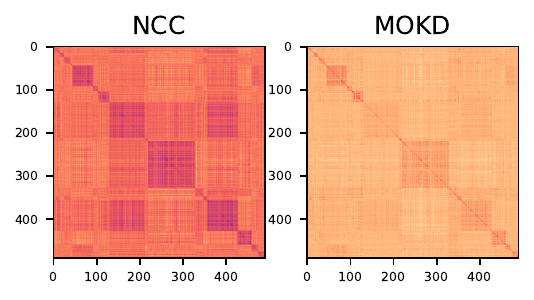}
		\end{minipage}}\vspace{-0.1cm}	
    \subfigure[MNIST (9 classes)\label{Fig:visual_mnist_sim}]{
			\begin{minipage}[t]{0.485\linewidth}
				\centering
				\includegraphics[width=1.0\linewidth]{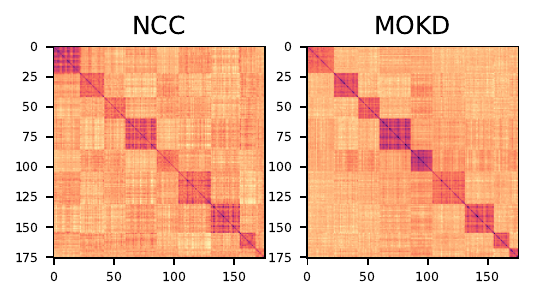}
		\end{minipage}}\vspace{-0.1cm}
	\subfigure[CIFAR 10 (6 classes)\label{Fig:visual_cifar10_sim}]{
			\begin{minipage}[t]{0.485\linewidth}
				\centering
				\includegraphics[width=1.0\linewidth]{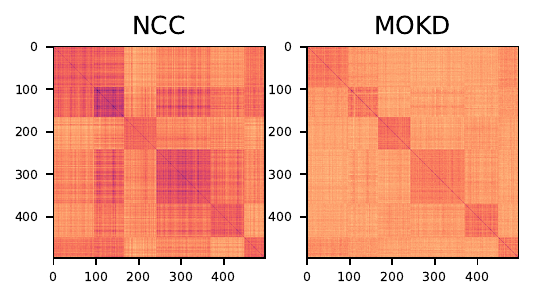}
		\end{minipage}}\vspace{-0.1cm}
    \subfigure[CIFAR 100 (24 classes)\label{Fig:visual_cifar100_sim}]{
			\begin{minipage}[t]{0.485\linewidth}
				\centering
				\includegraphics[width=1.0\linewidth]{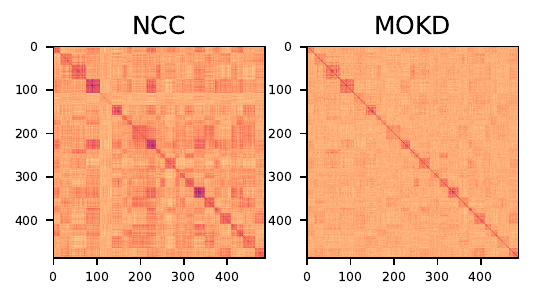}
		\end{minipage}}\vspace{-0.1cm}
    \caption{Visualization results of similarity matrices of representations repsectively learned with NCC-based loss and MOKD on unseen domains.}
    \label{Fig:remaining_visual_sim_unseen}
	\end{center}
	\vskip -0.2in
\end{figure}

For complicated datasets, such as ImageNet and MSCOCO, there are many objects and abundant semantic information in images. However, most of semantic information is useless and sometimes has a negative effect on representation learning. Thus, when tasks are sampled from these datasets, it is challenging for the model to learn definite and discriminative representations for each class. This will in turn result in uncertainties. For example, Fig. \ref{Fig:visual_vgg_flower_sim} and \ref{Fig:visual_cifar10_sim} have demonstrated this. Meanwhile, due to the scarce data in few-shot classification tasks and the strong power of ${\rm HSIC}(Z, Y)$, the model tends to learn high-variance representations and overfit the data.

Thus, a large gamma value is essential for these complicated datasets. For one thing, ${\rm HSIC}(Z, Z)$ facilitates penalizing high-variance kernelized representations for further alleviating the overfitting phenomenon. For another thing, according to its definition, ${\rm HSIC}(Z, Z)$ measures the dependence between two sets of data. Thus, minimizing ${\rm HSIC}(Z, Z)$ drives the model to learn discriminative features for each single sample so that samples are ``independent'' of each other. This further helps remove the trivial common features shared across samples and in turn, alleviates the high similarities among samples. In contrast, since simple datasets, such as Omniglot and Aircraft, own evident and definite semantic area, small $\gamma$ is enough.

Thus, in our work, we set small $\gamma$ for those simple datasets, such as Aircraft, Omniglot, and MNIST. However, for those complex datasets, such as ImageNet and MSCOCO, we set large $\gamma=3$ (inspired by SSL-HSIC~\citep{sslhsic}).

\subsection{Comparisons of Different Bandwidth Selection}\label{Appendix:bandwidth_selection}
In this paper, we adopt the grid search method to select the optimal bandwidth for the kernel to maximize its test power for dependence detection. In addition to the grid search~\citet{jitkrittum2016interpretable}, many previous works have been done to optimize the Gaussian kernel in this field. Specifically, ~\citet{el2024more} similarly proposed to select the bandwidth in a range of values $2^{\beta}$, where $\beta = \{-15, -13, ..., 4, 5\}$. \citet{sslhsic} proposed to optimize the kernel distance entropy to tune kernel parameters automatically. To figure out the differences among these strategies, we perform MOKD respectively with each of these strategies. To be clear, we denote these invariants respectively as ``MOKD + $2^{\beta}$'' and ``MOKD + entropy''. The results are reported in the following table. All results are obtained with random seed 42.

\begin{table}[H]
\vspace{-0.5em}
	\caption{\textbf{Comparisons of different bandwidth selection strategies.}}
	\label{Table:comparision_bandwidth_selection}
	\begin{center}
		\begin{tiny}
		\setlength\tabcolsep{4pt}
			\begin{threeparttable}
				\begin{tabular}{lccccccccccccc}
					\toprule
					Datasets & ImageNet & Omniglot & Aircraft & Birds & DTD & QuickDraw & Fungi & VGG\_Flower & Traffic Sign & MSCOCO & MNIST & CIFAR10 & CIFAR100 \\
					\midrule
					MOKD  & \bf{57.5$\pm$1.1} & 94.3$\pm$0.4 & 88.3$\pm$0.5 & 80.2$\pm$0.8 & \bf{76.7$\pm$0.7} & 82.5$\pm$0.6 & \bf{67.8$\pm$1.0} & \bf{92.8$\pm$0.5} & \bf{64.9$\pm$1.1} & \bf{55.6$\pm$1.0} & 95.1$\pm$0.4 & \bf{73.0$\pm$0.8} & \bf{64.5$\pm$1.0}\\
					MOKD + $2^{\beta}$ & 56.9$\pm$1.1 & \bf{94.4$\pm$0.4} & 88.3$\pm$0.5 & 80.2$\pm$0.8 & 76.2$\pm$0.7 & 82.5$\pm$0.6 & 65.2$\pm$1.0 & 92.3$\pm$0.5 & 62.7$\pm$1.2 & 54.8$\pm$1.0 & \bf{95.6$\pm$0.4} & 72.4$\pm$0.8 & 63.5$\pm$1.0\\
                    MOKD + entropy & 56.8$\pm$1.1 & 94.0$\pm$0.5 & 88.1$\pm$0.5 & 80.0$\pm$0.8 & 76.0$\pm$0.7 & 82.3$\pm$0.6 & 63.6$\pm$1.1 & 91.9$\pm$0.6 & 62.5$\pm$1.2 & 52.0$\pm$1.0 & 95.5$\pm$0.4 & 72.6$\pm$0.8 & 62.6$\pm$1.0 \\
					\bottomrule
				\end{tabular}
			\end{threeparttable}
		\end{tiny}
	\end{center}
	\vskip -0.1in
	\vspace{-1.0em}
\end{table}

As we can observe from the table, it is easy for us to observe that the grid search approaches achieve better performance than the entropy optimization method. Specifically, we can observe evident performance gaps on Fungi, VGG\_Flower, Traffic Sign, MSCOCO, and CIFAR-100 datasets. Further, we can observe that the test power maximization strategy adopted in our paper achieves better performance than ``MOKD + $2^{\beta}$. The main difference between the two strategies is the original MOKD scales the median of the kernel matrix while ``MOKD + $2^{\beta}$'' uses $2^{\beta}$ as the bandwidth. However, $2^{\beta}$ is not an empirically appropriate bandwidth for the kernel.

\end{document}